\documentclass[leqno,11pt]{article}
\usepackage{glottometrics}
\usepackage[noabbrev]{cleveref}

\glottovol{XX}
\glottoyear{20XX}
\glottodoi{Here will be added DOI number by Glottometrics.}

\runningauthors{Petrini, Casas-i-Muñoz, Cluet-i-Martinell, Wang, Bentz \& Ferrer-i-Cancho}
\runningtitle{The optimality of word lengths.}

\title{The optimality of word lengths. Theoretical foundations and an empirical study.}

\addauthor{1}{Sonia Petrini}{0000-0002-0514-6223}
\addauthor{2}{Antoni Casas-i-Muñoz}{0000-0001-5690-316X}
\addauthor{2}{Jordi Cluet-i-Martinell}{0000-0003-4188-6728}
\addauthor{2}{Mengxue Wang}{0000-0002-8262-9333}
\addauthor{3,4}{Christian Bentz}{0000-0001-6570-9326}
\addauthor{1*}{Ramon Ferrer-i-Cancho}{0000-0002-7820-923X}

\affil{1}{Quantitative, Mathematical and Computational Linguistics Research Group. Departament de Ci\`encies de la Computaci\'o, Universitat Polit\`ecnica de Catalunya (UPC), Barcelona, Catalonia, Spain.}
\affil{2}{Universitat Polit\`ecnica de Catalunya (UPC), Barcelona School of Informatics, Barcelona, Catalonia, Spain.}
\affil{3}{Department of Language Science and Technology, Saarland University, Saarbrücken, Saarland, Germany.}
\affil{4}{Chair of Multilingual Computational Linguistics, University of Passau, Passau, Bavaria, Germany.}

\correspond{*}{Corresponding author's email: rferrericancho@cs.upc.edu}

\usepackage{xcolor}
\usepackage{hyperref} % added to be able to use \autoref

\DeclareMathOperator{\expect}{\mathbb{E}}
\newcommand \expe[1] {\expect\left[\,#1\,\right]}

\DeclareMathOperator{\sgn}{sgn}

\usepackage{amsthm}

\newtheorem{property}{\bf Property}[section]
\newcommand{\repository}[0]{https://github.com/IQL-course/IQL-Research-Project-21-22}

\newcommand{\figcorrsignificance}[0]{1}

\usepackage{xeCJK} % for Chinese and Japanese script
\usepackage{float}
\usepackage{multirow}
\usepackage{subcaption}
\usepackage[toc,page,title]{appendix} % this is for complex appendices as ours; the template of the journal only offers a plain appendix.

\usepackage{gb4e}
\noautomath
\usepackage{tipa}
\RequirePackage{booktabs} 

\begin{document}
\maketitle

%==================================================
% Abstract

\begin{abstract}
% The article should contain an abstract within 200 words. The abstract states briefly the purpose of the research, methodology, key results, and major conclusions. The abstract should be a citation-free single paragraph with running sentences. Use the \texttt{\textbackslash maketitle} command before the text of your abstract as shown in this .tex file.
Zipf's law of abbreviation, namely the tendency of more frequent words to be shorter, has been viewed as a manifestation of compression, i.e. the minimization of the length of forms -- a universal principle of natural communication. Although the claim that languages are optimized has become trendy, attempts to measure the degree of optimization of languages have been rather scarce. 
Here we present two optimality scores that are dualy normalized, namely, they are normalized with respect to both the minimum and the random baseline. We analyze the theoretical and statistical advantages and disadvantages of these and other scores. Harnessing the best score, we quantify the degree of optimality of word lengths per language. This includes parallel texts in 20 languages of 9 families, written in 8 scripts, as well as spoken data for 46 languages of 12 families, two constructed languages, and one isolate. Our analyses indicate that languages are optimized to 62 or 67 percent on average (depending on the source) when word lengths are measured in characters, and to 65 percent on average when word lengths are measured in time. In general, spoken word durations are more optimized than written word lengths in characters.
Our work paves the way to measure the degree of optimality of the vocalizations or gestures of other species, and to compare them against written, spoken, or signed human languages.
\end{abstract}

%==================================================
% Keywords


 %%%%%%%%% Introduction
\section{Introduction}
\label{sec:introduction}

Language universals -- properties that apply to all languages on Earth -- are ``vanishingly few'', as exceptions can often be found \parencite{Evans2009a}. Some of the promising candidates are linguistic laws \parencite{Zipf1949a,Semple2021a}, though these have taken a rather secondary role in mainstream linguistics since Zipf's pioneering research. 
One of the most robust patterns is Zipf's law of abbreviation: more frequent words tend to be shorter \parencite{Zipf1949a,Bentz2016a}. It has been claimed that there is an even stronger law which relates the length of a word with its co-text of occurrence \parencite{Piantadosi2011a}, but further research has demonstrated that this new law is weaker and not supported across all languages tested \parencite{Meylan2021a,Levshina2022a,Koplenig2022a}. Therefore, Zipf's law of abbreviation is, at present, one of the strongest laws of communication in terms of theoretical understanding \parencite{Kanwal2017a,Ferrer2019c,Ferrer2013a}, and in terms of empirical support across languages \parencite{Bentz2016a}, linguistic levels \parencite{Torre2019a,Hernandez2019a,Koshevoy2023a} and across species \parencite{Semple2021a}. 

However, while Zipf's law of abbreviation -- and other linguistic laws -- have been investigated under the umbrella term of ``efficient communication'', there is generally a lack of precise quantification of the \textit{degree of efficiency}. How optimized are natural languages in terms of word lengths across the board? How optimized is one language compared to another? The first question relates to the universal communicative pressures shaping natural languages. The second question relates to the diversity of encoding strategies within the bounds of these universal pressures. Research in this direction is currently hampered by a lack of clear formalizations of baselines. In other words, we have to ask: optimal compared to \textit{what}? 

This question is particularly urgent given that natural languages occur in different modalities: spoken, signed, and written. Since languages are often investigated based on written documents, this raises the question to what extent different writing systems influence the measured optimality of word lengths. To illustrate this point, take the following Mandarin Chinese example and the respective parallel sentence in English from the \textit{Parallel Universal Dependencies} (PUD):

Mandarin Chinese (cmn)\footnote{This sentence is taken from the file zh\_pud-ud-test.conllu (sent\_id = n01004009) of the PUD (\url{https://universaldependencies.org/}). The tokenization and transliteration into Pinyin is here taken directly from the PUD. The glossing is provided with reference to the online dictionary at \url{https://www.mdbg.net/chinese/dictionary}. INF: infinitive verbal form; PRT: particle, here roughly translating as `in order to'; POSS: possessive particle.}
\begin{exe}
  \ex\label{example_Chinese}
  學生通過實際運用來學習科學課程的内容。
  \glll 學生 通過 實際 運用 來 學習 科學 課程 的 内容 。\\
        xuéshēng tōngguò shíjì yùnyòng lái xuéxí kēxué kèchéng de nèiróng .\\
        student through practice use.INF PRT learn.INF science course POSS content .\\
  \glt `Students learn science content by applying it.'\\
\end{exe}

Since this is a parallel text, the overall content is kept approximately constant between the Mandarin Chinese and English sentences. However, the word lengths in UTF-8 characters are vastly different. On average, we have $\frac{18}{10}=1.8$ Han characters per word, $\frac{56}{10}=5.6$ Latin characters per word in the Pinyin transliteration, and $\frac{39}{7}\sim5.6$ characters per word in the English parallel sentence. To further complicate the picture, we can also count the distinct strokes the Han characters are composed of. For instance, the first word written in Han characters 學生 \textit{xuéshēng} `student' consists of overall 21 strokes.

A similar problem arises when we compare word lengths in written characters with spoken durations. \autoref{tab:cv_examples} gives examples deriving from the Common Voice (CV)\footnote{\url{https://commonvoice.mozilla.org/en/datasets}} corpus for French and Spanish. While the French words are slightly longer on average in terms of UTF-8 characters ($5.4$ versus $5$ characters/word), they are in fact shorter on average in spoken durations averaged across speakers ($0.29$ versus $0.34$ seconds/word).
Given this state of affairs, we propose to measure the optimality of word lengths with scores which are normalized to be independent of the alphabet size (in UTF-8 characters), the number of word tokens, as well as the number of word types in a text. These can then be used for a less biased comparison of word length optimization across different languages, writing systems, and modalities.

The remainder of the article is organized as follows. In \autoref{sec:measuringOptimality}, 
we discuss the information-theoretic background of the proposed scores (\autoref{sec:background}), and introduce three optimality scores (\autoref{sec:scores}), as well as the specific baselines, namely, the \textit{minimum baseline}, and the \textit{random baseline} (\autoref{sec:baselines}). These are the building blocks of our optimality scores. 
%In \autoref{sec:theory}, we analyze the mathematical properties of $L$, $\eta$, $\Psi$ and $\Omega$ and other optimality scores, showing the technical superiority of $\Psi$ and $\Omega$. 
In \autoref{sec:material} and \autoref{sec:methodology}, we present, respectively, the materials and methods to apply these scores to real languages. 
In \autoref{sec:results}, we show that one of the scores ($\Psi$) is the best optimality score according to a wide range of criteria. $\Psi$ indicates that languages are optimized to 62\% or 67\% on average (depending on the source) when word length is measured in characters, and to 65\% when word length is measured in durations. We also find that word lengths in durations are more optimized than word lengths in characters -- when language samples are sufficiently large. Moreover, Chinese and Japanese word lengths turn out to be more optimized when measured in characters rather than in strokes (or in Latin characters after romanization). Finally, in \autoref{sec:discussion}, we discuss the repercussions of our findings on a range of research questions related to the research program on the optimality of languages more generally \parencite{Ferrer2020b}. %This program was introduced in the context of syntactic dependency distances \parencite{Ferrer2020b}. 
As an outlook, we derive further proposals for future research. 

\begin{table}[h]
\center
\begin{tabular}{lllll}
Language & Word          & No. char. & IPA & Duration\\
\hline
French   & est           & 3         & \textepsilon        & 0.14            \\
         & une           & 3         & yn\textschwa     & 0.17            \\
         & sont          & 4         & s\~{o}           & 0.23            \\
         & comme         & 5         & k\textopeno m\textschwa                & 0.22            \\
         & informations  & 12        & \~{\textepsilon}f\textopeno \textfishhookr masj\~{o}  & 0.67            \\
         \hline
         & Average       & 5.4       &     &            0.29     \\
         \hline
Spanish  & es            & 2         &  es   & 0.19            \\
         & una           & 3         &  una   & 0.21            \\
         & son           & 3         &  son  & 0.25            \\
         & como          & 4         &  komo   & 0.27            \\
         & informaciones & 13        &  informa\texttheta jones & 0.78            \\
         \hline
         & Average       & 5         &                          & 0.34 \\ 
\end{tabular}\\
\caption{Cognates in French and Spanish with number of written characters, and mean duration in seconds as measured across speakers in the Common Voice (CV) corpus.}\label{tab:cv_examples}
\end{table}

\section{Measuring the optimality of word lengths}

\label{sec:measuringOptimality}

\subsection{Information-theoretic background}\label{sec:background}
Here we aim to shift the main focus from the surface of linguistic behavior (e.g. linguistic laws), and the all-or-nothing logics of absolute universals, to the principles underlying a general theory of natural communication \parencite{Ferrer2015b,Semple2021a,Ferrer2021a}. In this setting, the principles that govern communication across species are the true universals which manifest themselves in production (potentially with exceptions), and may require specific experimental conditions to appear \parencite{Ferrer2012a,Ferrer2019a}. This shift of focus requires a split between principles on one hand, and their manifestations on the other, as well as a theoretical understanding of when and why the manifestations of a principle reach the surface of the observable world. 
Since Zipf's pioneering research, the law of abbreviation has been argued to be a manifestation of word length minimization \parencite{Zipf1949a,Ferrer2012d,Ferrer2019c}, currently known as the principle of compression \parencite{Ferrer2012d,Semple2021a}. In its simplest form, the principle of compression can be formulated as the minimization of the average length of tokens from a repertoire of $n$ types, that is defined as 
\begin{equation}
L = \sum_{i=1}^n p_i l_i,
\label{eq:mean_type_length}    
\end{equation}
where $p_i$ and $l_i$ are, respectively, the probability and the length of the $i$-th type.
In practical applications, $L$ is calculated replacing $p_i$ by the relative frequency of a type, that is 
$$p_i = f_i/ T,$$ 
where $f_i$ is the absolute frequency of a type and $T$ is the total number of tokens, i.e.
$$T = \sum_{i=1}^n f_i.$$
This leads to a definition of $L$ that is 
$$L = \frac{1}{T} \sum_{i=1}^n f_i l_i.$$

As a concrete example, take the frequency and length distributions for English and Mandarin Chinese words given in \autoref{tab:engChinese}. To calculate the average length $L$ according to Equation~\ref{eq:mean_type_length}, we have to multiply each length $l_i$ of a given word type with its probability $p_i$. The probability, in turn, is derived as the relative frequency of a given type, i.e. a given $f_i$ over the total number of tokens (sum of $f_i$'s). In fact, $L$ can be seen as the \textit{average length of word tokens}, since we take token frequencies into account via the $p_i$'s. Given these definitions, it is clear that $L$ is influenced by the type of writing system (compare Hanzi and Pinyin lengths), the number of tokens $T$ (i.e. text size), the number of types (i.e. vocabulary size), and also the alphabet size $A$ -- there will be many more different Han characters than English and Pinyin characters. Writing in a more diverse character set allows for coding of frequent words with single characters, rather than combinations of characters.

The claim that languages are optimized for efficient communication has become trendy \parencite{Liu2017a,Gibson2019a,Levshina2022b}. However, attempts to actually measure the degree of optimization in a given dimension have been rather scarce \parencite{Ferrer2018b,Coupe2019a,Ferrer2020b,Koplenig2021a}. 
Here we aim to quantify the degree of optimality of average word lengths $L$ as a window to the strength of compression in languages. 

\begin{table}[]
\centering
\begin{tabular}{lllllll}
English & $f_i$ & $l_i$ & Chinese & $f_i$ & $l^{\text{hanzi}}_i$ & $l^{\text{pinyin}}_i$ \\
\hline 
the     & 1441  & 3     & 的 \textit{de} (possessive particle)     & 1362  & 1 & 2    \\
of      & 620   & 2     & 在 \textit{zài} (preposition `in')       & 415   & 1 & 3   \\
in      & 510   & 2     & 了 \textit{le} (aspective particle)      & 380   & 1 & 2   \\
to      & 481   & 2     & 一 \textit{y\={\i}} (`one')      & 249   & 1  & 2   \\
and     & 456   & 3     & 是 \textit{shì} (copular `be')  & 215   & 1 & 3 \\ 
\end{tabular}
\caption{Five most frequent words for English and Mandarin Chinese in the PUD corpus.}\label{tab:engChinese}
\end{table}

\subsection{Optimality scores}\label{sec:scores}

The degree of optimality of syntactic dependency distances -- reflecting the manifestation of the dependency distance minimization principle -- has been investigated for two decades \parencite{Hawkins1998a,Ferrer2004b,Tily2010a,Gulordava2015,Gulordava2016a,Ferrer2020b}. 
In contrast, the degree of optimality of word lengths has been investigated only recently \parencite{Ferrer2018b,Moreno2021a,Pimentel2021a}.
This degree of optimization has been measured with the $\eta$-score \parencite{Borda2011a,Ferrer2018b,Pimentel2021a}\footnote{For \textcite{Pimentel2021a}, see Fig. 5.}:
\begin{equation}
\eta = \frac{L_{min}}{L},
\label{eq:eta}
\end{equation}
where $L$ is the average length of tokens, as defined in Equation~\ref{eq:mean_type_length} above, and $L_{min}$ is the \textit{minimum baseline}, namely, the minimum value that $L$ can achieve -- given certain assumptions. We say that a language is $x\%$ optimal with respect to $\eta$ if $\eta = x/100$ (we apply the same convention to other scores). The analyses using $\eta$ indicate that languages are $30\%$ optimal on average under \textit{non-singular} coding and $40\%$ optimal on average under \textit{unique decodability} \parencite{Ferrer2018b}. 

Here we will investigate the degree of optimality of languages with two new scores that we develop from recent research on the optimality of dependency distances \parencite{Ferrer2020b}. One is
\begin{equation}
\Psi = \frac{L_r - L}{L_r - L_{min}},
\label{eq:Psi}
\end{equation}
where $L_r$ is the random baseline for $L$.
The other is 
\begin{equation}
\Omega = \frac{\tau}{\tau_{min}},
\label{eq:Omega}
\end{equation}
where $\tau$ is the Kendall correlation coefficient between $p_i$ and $l_i$, and $\tau_{min}$ is the minimum baseline for $\tau$. For further mathematical details and properties of these scores we refer the reader to \autoref{app:optimalityScores}.\\

 %%%%%%%%% Baselines
\subsection{The baselines}

When it comes to defining baselines, researchers often turn to randomly generating data. For instance, \textcite{Miller1957} famously proposed to use ``monkey typing'' as a generative process, and then compare its output to natural language in terms of Zipfian laws. However, there are two main arguments against using random typing as a baseline:

Firstly, in stark contrast to common expectations, \textcite{Ferrer2019c} prove that random typing is actually an \textit{optimal} coding process. In fact, this is a logical consequence of how random typing is defined: when randomly drawing characters of a given alphabet with predefined probabilities, the overall probability of a string $p_i$ is a function of its length ($l_i$). 
If $q$ is the probability of producing a character (not a word delimiter) and the alphabet contains $N$ characters that are equally likely, then the probability of a type of length 1 is
$$p_i(1) = \frac{1 - q}{N}.$$
For longer types, types of length $l$ ($l>1$), we have that
$$p_i(l) = \frac{q}{N} p_i(l-1).$$
Hence, longer types are less likely because $\frac{q}{N}$ is a number between 0 and 1.
% In the case of uniform probability $q$ for all characters, this is simply \parencite{Miller1957,Ferrer2009a}
% $p_i=q^{l_i}$.
% If the probability of not producing a word delimiter is $q$ and all characters of the alphabet are equally likely, the probability of a type is 
% \textcolor{red}{$$p_i = \frac{1}{N^{l_i}} q^{l_i-1} (1 - q),$$
% where $N$ is the alphabet size. ???}  
In other words, there is a strong link between string length and string probability built into the generative process of random typing \textit{a priori}. In particular, more frequent types are shorter as expected from the compression principle.

Secondly, random typing is psychologically implausible. Humans (or animals) do not randomly churn out phonemes or graphemes when communicating. Rather, there is a host of factors -- phonotactic, morphosyntactic, semantic, pragmatic, cognitive, sociolinguistic -- which govern \textit{what} is said and \textit{how}. While it is impossible to take all of these into account, in the following we aim to define baselines which are more realistic than random typing.

\label{sec:baselines}

\subsubsection{The random baseline}

Many different pressures will act on word lengths and probabilities in the course of language change. We do not attempt to model these here. Rather, we simply assume that both the lengths of words and their probabilities are a given -- at the point in time when we measure them in corpora. Our null hypothesis is then a random one-to-one mapping of word probabilities onto word lengths, which leads to the random baseline being defined as \parencite{Petrini2022b}
\begin{equation}
L_r = \frac{1}{n} \sum_{i=1}^n l_i.
\label{eq:random_baseline}
\end{equation}
A random one-to-one mapping can be obtained in three equivalent ways (1) by shuffling the $p_i$'s, (2) by shuffling the $l_i$'s and (3) by shuffling each of them \parencite{Petrini2022b}.
This baseline has been referred to as ``shuffle coding'' in related work \parencite{Pimentel2021a}.
Note that $L_r$ actually corresponds to the mean length of \textit{word types}, whereas the mean word length $L$ corresponds to the mean length of \textit{word tokens}. If word lengths are randomly mapped onto word probabilities as explained above, the expected value of word lengths is equal to the value when all word types are equally likely, that is, the relative frequency of a word is irrelevant to its length. See \textcite{Petrini2022b} for further details on this random baseline, as well as a comparison showing that $L < L_r$ on the same languages used for the present article.  

%\textcolor{red}{To see the difference, consider a matrix with two columns, $f_i$ and $l_i$, that are used to compute $L$. The matrix in \autoref{tab:example} gives $L = \frac{235}{125} = 1.88$ and $L_r = 2$. }

\subsubsection{The minimum baseline}
\label{sec:coding_schemes}

$L_{min}$ is the minimum value that $L$ can achieve making certain assumptions. 
% If no assumptions are made,
For example, if the only assumptions are $l_i \geq 0$ and 
$$\sum_{i=1}^n p_i = 1,$$
then $L_{min} = 0$ \parencite{Ferrer2019c}. 
% In this scenario, the optimality of word lengths -- as measured for instance by $\eta$ -- becomes constant ($\eta = 0$ according to \autoref{eq:eta}). 
In plain words, not communicating at all would reduce $L_{min}$ to zero, and hence lead to maximum compression. Of course, this is not a realistic scenario if we assume that there are pressures to communicate. We might introduce another constraint, namely, that strings of length zero are not considered valid types, but rather that $l_i \geq 1$. In this case $L_{min} = 1$. That is, the minimum average length of words should be one. As a matter of fact, however, not even scripts with a rich set of characters will allow for just words of length one (remember the 1.8 Han characters per word from above).

To derive a more realistic minimum baseline, we follow standard information theory and its extensions. Here, the $p_i$'s are assumed to be given, and optimization is assumed to operate only on the $l_i$'s \parencite{Ferrer2018b,Ferrer2019c}. Arguably, this also makes sense from a linguistic point of view. How often we use a word is guided by many factors, its length will play a minor role -- if any. For example, if we have a conversation about kitchen appliances, we might have to refer to the concept of a \textit{refrigerator} a certain number of times. We cannot simply replace this word by other words -- \textit{oven}, \textit{freezer}, \textit{microwave}, etc. -- just because these are shorter. However, if we have to refer to the concept often, we might shorten its length to \textit{fridge}. So in this case, compression acts on the length of the word \textit{given} its probability.

Previous research on the optimality of word lengths has taken into account different kinds of coding schemes (e.g. non-singular coding and uniquely decodable coding) from standard information theory \parencite{Ferrer2018b, Moreno2021a}. A limitation of these approaches is that, in order to calculate $L_{min}$, one has to assume that word length is discrete,  and that all characters have the same weight. Also, one has to choose an alphabet size (for instance, one has to decide if the alphabet size will be fixed or will depend on the language).
Here we focus on computing $L_{min}$ according to the so-called \textit{rank ordering} (RO) method \parencite{Moreno2021a}, which does not suffer from the limitations above. This method has been referred to as ``Zipfian coding'' in related work \parencite{Pimentel2021a}. In particular, $L_{min}^{RO}$ is obtained when the current values of $l_i$ are reassigned to frequencies (or equivalently probabilities) so as to minimize $L$. Namely, $L$ is minimized when probabilities are sorted decreasingly and lengths are sorted increasingly \parencite{Ferrer2019c}. In this case, the $i$-th most frequent type gets the $i$-th shortest length, hence the name {\em rank ordering}. Throughout this paper and appendices, $L_{min}$ normally refers to $L_{min}^{RO}$ -- unless indicated otherwise. In parallel to $L_{min}$, $\tau_{min}$ is also defined by the rank ordering principle. See \autoref{app:baselines} and \autoref{app:technical_remmarks_on_normalization_and_coherence} for further mathematical details and discussions of this and other minimum baselines.

In a nutshell, using the rank ordering minimum baseline, we hold that the effort of communication would be minimized if indeed there was a perfect inverse match between the probabilities of words and their length, namely, if there was a perfect agreement with Zipf's law of abbreviation. The question then is how far away from this optimum a given language is. 

\subsection{The properties of the baselines and optimality scores}

\label{sec:properties_of_baselines}
In summary, we investigate optimality scores, i.e. $\eta$, $\Psi$ and $\Omega$, which are defined using the baselines $L_r$ and $L^{RO}_{min}$. 
By choosing the previous minimum baselines, the scores measure the {\em closeness to a perfect law of abbreviation}. A perfect fit to the law of abbreviation means that the lengths are arranged optimally given the probabilities of words.
Importantly, note that the minimum and the random baseline share various statistical properties with the original source:
\begin{enumerate}
    \item 
    \label{it:distributions}
    The distribution of word frequencies and the distribution of word probabilities, 
    \item
    \label{it:types_and_tokens}
    The number of tokens and the number of types, 
    \item
    The alphabet size in the case of written language (and the repertoire of phonemes and larger constructs in the case of spoken language). 
\end{enumerate}
In extension, our optimality scores assume that all these statistical characteristics are fixed; only the one-to-one mapping of word probabilities into word lengths can be changed. 
A fundamental question is whether a given comparison between two languages is supported by the mathematical properties of the optimality scores used. In the following, we discuss different scenarios of how frequency/length distributions used in communication can relate to one another, and what this implies for the usage of our optimality scores.

%For comparison, 
%the optimality scores for dependency distances discussed in \parencite{Ferrer2020b,Moreno2021a} assume that the dependency structure is fixed and that only the order of words can be changed.
%Besides, the minimum baselines of standard information theory and its extensions (optimal non-singular coding and optimal uniquely decodable encoding; \autoref{sec:coding_schemes}) do not preserve at least properties \autoref{it:distributions} and \autoref{it:types_and_tokens} above.

\subsection{General problems when comparing communication systems}
\label{sec:problems_on_communication_systems}
Borrowing the general setting of previous research into Zipf's law of abbreviation \parencite{Ferrer2019c}, we might reduce a communication system to a mapping from types to codes, which may not be discrete. The key of the mapping is the code lengths, represented as natural numbers when we measure the length of words in characters, or real numbers when we measure the duration of a vocalization or a gesture \parencite{Semple2021a}. In this general setting, suppose that we have two communication systems, A and B. 

\subsubsection{The weak recoding problem.} In this recoding problem, B is just a transformation of A 
%by 
assigning a new code, e.g. a new string, to each type of A (hence preserving the frequency of every type from A in B while changing their length). Then suppose that $s(X)$ is the optimality score of a communication system $X$. The question is, under what conditions $s(A) = s(B)$? This question cannot be answered unless we clarify what we consider to be relevant or not for $s(A) = s(B)$. Invariance under \textit{increasing linear transformation}, for instance, is a possible requirement: we will have $s(A) = s(B)$ when the score satisfies this type of invariance, i.e. the new lengths in B are an increasing linear transformation of those in A. Similarly, another condition for $s(A) = s(B)$ can be invariance under \textit{strictly increasing transformation}. A set of mathematical properties including the aforementioned ones is discussed in \autoref{app:optimalityScores} with regards to the proposed optimality scores. In this article in particular, we will deal with three instances of the weak recoding problem:
\begin{enumerate}
    \item 
    {\bf Length in characters versus duration in the same language}. In this instance, the lengths in characters in one of the communication systems, say A, have been replaced by the corresponding durations in time to produce the other, say B. This corresponds to the mapping of word lengths in characters to durations for Spanish and French examples in \autoref{tab:cv_examples}.  
    \item
    {\bf Length in Chinese/Japanese characters versus other discrete lengths (romanizations and strokes)}. Here, lengths in Chinese/Japanese characters are replaced by lengths in strokes or in romanizations, namely Latin script conversions (via Pinyin for Chinese and Romaji for Japanese). Such replacements certainly lead to an increase in $L$ (See \autoref{tab:engChinese} as well as \autoref{tab:opt_scores_pud}). However, a key question is whether the \textit{degree of optimality} will also change as a consequence of this increase in $L$. % the explanations above are relevant for Lukasz Debowki's comment on twitter.
    %We will show that the degree of optimality reduces in line with the increase in $L$. This suggests that romanizations or strokes are a poor proxy to investigate the degree of optimality of these languages with respect to languages written in other scripts.
    \item
    {\bf Removal of vowels in Latin scripts or romanizations}. The motivation for this recoding is to check if the scores are robust to varying decisions with regard to the design of a writing system. For instance, writing systems differ in how they code for vowels. Catalan -- as all Romance languages -- uses a Latin script, where both consonants and vowels are represented. In contrast, the primary writing system of Arabic is an abjad, where only consonants are represented (unless \textit{fatha} diacritics are used to indicate vowels). Removing vowels in Romance languages certainly leads to a decrease in $L$, but, again, the key question is whether their degree of optimality will also change. %Interestingly, we will show that the degree of optimality of word lengths decreases only slightly after removing vowels, despite mean word length being reduced considerably. Hence, it is important to use powerful optimality scores. % The reduction in optimality never exceeds one decimal ???.
\end{enumerate}

Suppose now that coding in A is non-singular, namely no two distinct types are assigned the same code. After replacing the strings in A by new strings, it may turn out that B ceases to be non-singular. That could happen because two types in A end up having the same code or one type is assigned the empty string. 
% Chris: new text follows.
This motivates the need to define a strong recoding problem where non-singularity is preserved.

\subsubsection{Strong recoding problem.} In this recoding problem, B is initially obtained from A as in the weak recoding problem, but additionally, all types in B that are assigned  the same string are merged into the same type, and types that are assigned the empty string are dropped. Notice that strong recoding may change the distribution of word frequencies, a feature that the weak recoding problem preserves.
% Chris: new text follows.
Notice that if A is non-singular then B will also be non-singular. 
All the instances of the weak recoding problem presented above can be investigated under the framework of the strong recoding problem as well. However, in this article, we investigate directly only the weak recoding problem for the sake of simplicity and due to pressure for space. 

\subsubsection{The free comparison problem.} Above, we have introduced problems where B stems from A. 
The free comparison problem appears when A and B are {\em a priori} two different communication systems, hence the number of types and their distributions differ. This setting raises the question of whether the optimality scores of two languages with distinct writing systems or distinct evolutionary histories are comparable. This is a big research challenge for analyses of the degree of optimality in written languages, given the disparity of writing systems of the world \parencite{Daniels1990a}, which is also reflected in the dataset in the present study 
(\autoref{tab:coll_summary_pud} and \autoref{tab:coll_summary_cv}).
Some more specific questions in this context are: 
\begin{enumerate}
    \item 
    Can we compare Romance languages against Standard Arabic, although the former code for vowels and the latter essentially do not? We will show that removing vowels in Romance languages does not have a big impact on the values of the optimality scores, suggesting that the original scores for Romance languages are comparable to those of Standard Arabic. 
    \item
    Can we compare Romance languages -- which use a script that essentially codes for phonemes -- against syllabic/logographic writing systems, namely systems that use a character for every syllable? Examples of syllabic/logographic writing systems are Chinese Hanzi, where every character stands for a syllable, or abugida writing systems.
\end{enumerate}
We here aim to design scores that abstract away from cultural differences, hence providing an insight into the actual degree of optimality of word lengths. Ultimately, an entirely ``fair'' comparison may be impossible to accomplish. Our take is that we define scores that can at least give reliable results under ideal conditions. 

%In spite of these considerations, one may argue that the relationship between the old and the new word length does not need to be linear. Crucially, only $\Omega$ is invariant under a strictly monotonic non-linear transformation of lengths. Further details about the mathematical properties of the optimality scores are given in \autoref{app:optimalityScores}. The next subsection gives additional desirable properties that will turn out to be crucial to choose the best score. 
% \input{extra/old_ugly_text}

\subsection{Desirable properties of scores}

\label{sec:other_desirable_properties}

The score fulfilling most mathematical properties (i.e. most checks in \autoref{tab:mathematical_properties_of_scores} of \autoref{app:optimalityScores}) is not necessarily the best score. Further properties are also desirable. 

Firstly, our baselines, and therefore the scores that are defined on top of them, assume that the number of tokens ($T$), the number of types ($n$), and the alphabet size ($A$) are fixed, and that only the mapping of probabilities into words can be changed (\autoref{sec:properties_of_baselines}). In this setting, a desirable property of a score is that it is not influenced by these characteristics. If a score is heavily correlated with any of these parameters across languages, it can be argued that, rather than observing differences in the degree of word length optimality of languages, we are observing differences in text length (in tokens), or in the size of the alphabet, or the vocabulary. A most critical dependence would be on the number of tokens. This would be particularly worrying when using non-parallel corpora, where higher variation in text length is expected in comparison to parallel texts. 

Secondly, an important question is whether a score converges to a stable value, given a sufficiently large language sample, and how fast this convergence is. This approaches the problem of dependency on the number of tokens within a given language. The equivalent problem across languages is tackled by the first point above.

Thirdly, another important and related question is whether we can replace more complex scores by simpler ones, regardless of all the theoretical arguments summarized in \autoref{tab:mathematical_properties_of_scores}. In the extreme case, could one of the best scores, say $\Psi$, be replaced simply by $L$ because the former is strongly correlated with the latter?

All these questions are difficult to investigate theoretically and will be tackled empirically at the beginning of \autoref{sec:results} so as to help choose a primary score for reporting results. %We will show that $\Psi$ is the most powerful score in terms of independence of these parameters (text length and size of the alphabet or size of the vocabulary) as well as in terms of convergence; $\Psi$ cannot be replaced by neither $L$ nor $\eta$ in parallel texts. 

%%%% Material
\section{Material}
\label{sec:material}

We borrow a dataset with information about word length and word frequency from recent research \parencite{Petrini2022b}, available in the repository of this article 
%\parencite{repository}. 
\parencite{petrini2026software}. 
% \footnote{In the \textit{data} folder of \url{\repository}.}. 
The dataset has been extracted from two collections of texts: 
\textit{Common Voice Forced Alignments},\footnote{\url{https://commonvoice.mozilla.org/en/datasets}, for the forced alignments see \url{https://github.com/JRMeyer/common-voice-forced-alignments.}} % (\autoref{CVFA}), 
hereafter CV, and \textit{Parallel Universal Dependencies},\footnote{https://universaldependencies.org/}
% (\autoref{PUD}), 
hereafter PUD. 
PUD comprises 20 distinct languages from 9 linguistic families and 8 scripts. 
(\autoref{tab:coll_summary_pud}).
CV comprises 46 languages from 12 linguistic families, two constructed languages, one isolate (Basque), and overall 10 scripts. 
(\autoref{tab:coll_summary_cv}).
The typological information (language family) is obtained from Glottolog 4.6.\footnote{\url{https://glottolog.org/}} %(hence Turkish is from the Turkic family, not Altaic as in the  \href{https://wals.info/}{World Atlas of Language Structures}).
The writing systems are determined according to ISO-15924 codes.
See \autoref{tab:coll_summary_pud} and \autoref{tab:coll_summary_cv} for basic statistical properties of the datasets. These are reproduced for convenience from \textcite{Petrini2022b}.\footnote{Notice that the counterpart of this table in \textcite{Petrini2022b}, Table 3, was distorted by bugs that were fixed to generate the version in this article.}

\begin{table}[H]
\centering
% latex table generated in R 4.2.1 by xtable 1.8-4 package
% Fri Sep 22 18:31:08 2023

\begin{tabular}{lllrrr}
 language & family & script & $A$ & $n$ & $T$ \\ 
    \hline
 Arabic & Afro-Asiatic & Arabic &  39 & 6596 & 18201 \\ 
  Indonesian & Austronesian & Latin &  23 & 4221 & 16305 \\ 
  Russian & Indo-European & Cyrillic &  31 & 7113 & 15588 \\ 
  Hindi & Indo-European & Devanagari &  50 & 4716 & 20796 \\ 
  Czech & Indo-European & Latin &  33 & 7069 & 15286 \\ 
  English & Indo-European & Latin &  25 & 5001 & 18021 \\ 
  French & Indo-European & Latin &  26 & 5211 & 19812 \\ 
  German & Indo-European & Latin &  28 & 6108 & 18003 \\ 
  Icelandic & Indo-European & Latin &  32 & 6035 & 16207 \\ 
  Italian & Indo-European & Latin &  24 & 5528 & 19935 \\ 
  Polish & Indo-European & Latin &  32 & 7204 & 15216 \\ 
  Portuguese & Indo-European & Latin &  37 & 5621 & 20367 \\ 
  Spanish & Indo-European & Latin &  32 & 5689 & 20602 \\ 
  Swedish & Indo-European & Latin &  25 & 5624 & 16369 \\ 
  Japanese & Japonic & Japanese & 1577 & 4852 & 24737 \\ 
  Japanese-strokes & Japonic & Japanese & 1549 & 4852 & 24737 \\ 
  Japanese-romaji & Japonic & Latin &  28 & 4849 & 24734 \\ 
  Korean & Koreanic & Hangul & 1002 & 8031 & 14475 \\ 
  Thai & Kra-Dai & Thai &  52 & 3599 & 21121 \\ 
  Chinese & Sino-Tibetan & Han (Traditional variant) & 2071 & 4970 & 17845 \\ 
  Chinese-strokes & Sino-Tibetan & Han (Traditional variant) & 2038 & 4970 & 17845 \\ 
  Chinese-pinyin & Sino-Tibetan & Latin &  54 & 4970 & 17845 \\ 
  Turkish & Turkic & Latin &  28 & 6373 & 13512 \\ 
  Finnish & Uralic & Latin &  24 & 6933 & 12691 \\ 
\end{tabular}

\caption{Summary of the main characteristics of the languages in the PUD collection. For each language, we show the linguistic family, the writing system (script name according to ISO-15924) and various numeric parameters: $A$, the observed alphabet size (number of distinct characters), $n$, the number of word types, and $T$, the number of word tokens. }
\label{tab:coll_summary_pud}
\end{table}

\begin{table}[H]
\centering
% latex table generated in R 4.1.1 by xtable 1.8-4 package
% Mon Jul 18 12:50:44 2022
\begin{tabular}{lllrrr}
Language & Family & Script & $A$ & $n$ & $T$\\ 
\hline
Arabic & Afro-Asiatic & Arabic &  31 & 6397 & 45825 \\ 
Maltese & Afro-Asiatic & Latin &  31 & 8058 & 44112 \\ 
Vietnamese & Austroasiatic & Latin &  41 & 370 & 938 \\ 
Indonesian & Austronesian & Latin &  22 & 3768 & 44210 \\ 
Esperanto & Conlang & Latin &  27 & 27759 & 406261 \\ 
Interlingua & Conlang & Latin &  20 & 5126 & 30504 \\ 
Tamil & Dravidian & Tamil &  29 & 1210 & 6439 \\ 
Persian & Indo-European & Arabic &  38 & 13115 & 1662508 \\ 
Assamese & Indo-European & Assamese &  43 & 971 & 1813 \\ 
Russian & Indo-European & Cyrillic &  32 & 31827 & 637686 \\ 
Ukrainian & Indo-European & Cyrillic &  34 & 14337 & 120760 \\ 
Panjabi & Indo-European & Devanagari &  37 &  84 &  98 \\ 
Modern Greek & Indo-European & Greek &  33 & 5813 & 37880 \\ 
Breton & Indo-European & Latin &  28 & 4228 & 38237 \\ 
Catalan & Indo-European & Latin &  39 & 79112 & 3294206 \\ 
Czech & Indo-European & Latin &  33 & 15518 & 147582 \\ 
Dutch & Indo-European & Latin &  23 & 10225 & 316498 \\ 
English & Indo-European & Latin &  28 & 173023 & 9828713 \\ 
French & Indo-European & Latin &  49 & 160243 & 3729370 \\ 
German & Indo-European & Latin &  30 & 148436 & 4230565 \\ 
Irish & Indo-European & Latin &  23 & 2251 & 22593 \\ 
Italian & Indo-European & Latin &  34 & 54996 & 811783 \\ 
Latvian & Indo-European & Latin &  27 & 7251 & 29456 \\ 
Polish & Indo-European & Latin &  32 & 25340 & 595411 \\ 
Portuguese & Indo-European & Latin &  27 & 11509 & 283048 \\ 
Romanian & Indo-European & Latin &  29 & 6423 & 33341 \\ 
Romansh & Indo-European & Latin &  26 & 9614 & 43792 \\ 
Slovenian & Indo-European & Latin &  24 & 5937 & 26304 \\ 
Spanish & Indo-European & Latin &  33 & 75010 & 1842474 \\ 
Swedish & Indo-European & Latin &  25 & 4371 & 62951 \\ 
Welsh & Indo-European & Latin &  22 & 11143 & 539621 \\ 
Western Frisian & Indo-European & Latin &  30 & 8383 & 63073 \\ 
Oriya & Indo-European & Odia &  41 & 764 & 1700 \\ 
Dhivehi & Indo-European & Thaana &  27 & 111 & 1284 \\ 
Georgian & Kartvelian & Georgian &  25 & 6505 & 12958 \\ 
Basque & Language isolate & Latin &  21 & 24748 & 458071 \\ 
Mongolian & Mongolic & Mongolian &  31 & 14608 & 70217 \\ 
Kinyarwanda & Niger-Congo & Latin &  26 & 133815 & 1939810 \\ 
Abkhazian & Northwest Caucasian & Cyrillic &  28 & 119 & 156 \\ 
Hakha Chin & Sino-Tibetan & Latin &  23 & 2499 & 17776 \\ 
Chuvash & Turkic & Cyrillic &  22 & 4311 & 13583 \\ 
Kirghiz & Turkic & Cyrillic &  30 & 10130 & 61844 \\ 
Tatar & Turkic & Cyrillic &  34 & 21823 & 144356 \\ 
Yakut & Turkic & Cyrillic &  28 & 7904 & 22577 \\ 
Turkish & Turkic & Latin &  31 & 8926 & 107686 \\ 
Estonian & Uralic & Latin &  23 & 28691 & 121549 \\ 
\end{tabular}
\caption{Summary of the main characteristics of the languages in the CV collection. For every language we show its linguistic family, the writing system (script name according to ISO-15924) and various numeric parameters: $A$, the observed alphabet size (number of distinct characters), $n$,  the number of word types, and, $T$, the number of word tokens.
'Conlang' stands for 'constructed language', that is an artificially created language. This is not a family in the proper sense as Conlang languages are not related in the common linguistic family sense.
} 
\label{tab:coll_summary_cv}
\end{table}

\newpage
\subsection{The dataset}

The dataset provides the length of a word in characters (in PUD and CV) and its duration (in CV only) in two variants, median duration and mean duration (median is used by default, but mean is also used in certain cases as a control). It also provides the length in strokes and in romanizations for Japanese and Chinese (Romaji and Pinyin, respectively) in PUD.
The traditional writing systems of Japanese and Chinese yield very short word lengths in characters, while the number of distinct characters is very large, especially compared to Western languages with mostly alphabetic writing systems \parencite{Chen2015a, Joyce2012a}.\footnote{See  
\autoref{tab:coll_summary_pud} for alphabet size and \autoref{tab:opt_scores_pud} for average word length in Chinese and Japanese.} %We wish to test whether these differences are reflected in optimality scores. 

%This poses questions on the correct methodology to capture the studied phenomenon. \textcolor{violet}{is this previous part supposed to be here?}

\subsubsection{Common Voice Forced Alignments} 
% \label{CVFA}

Notice that Abkhazian, Panjabi, and Vietnamese have a critically low number of tokens 
(less than 1000 tokens according to \autoref{tab:coll_summary_cv}).
However, we decided to include them in the analyses so as to better understand problems with sample size.

\subsubsection{Parallel Universal Dependencies} 
% \label{PUD}

Notice that three Japanese words that are \textit{hapax legomena} could not be romanized and thus the number of tokens and types varies slightly with respect to the original Japanese characters 
(\autoref{tab:coll_summary_pud}).
% (Table 4 in \textcite{Petrini2022b}).
Thus, Japanese in strokes follows the setting of the weak recoding problem only approximately.

%%%%%% Methodology

\section{Methodology}
\label{sec:methodology}

All the code used to produce the results is available in the repository of this article 
% \parencite{repository}. 
\parencite{petrini2026software}. 
% \footnote{In the \textit{code} folder of \url{\repository}.}.
The bulk of the methods are borrowed from a preceding article,
including the unsupervised method to filter words with unusual or ``foreign'' characters \parencite{Petrini2022b}.
Next we highlight methods or variants thereof which are specific to this article. 

\subsection{Tokenization}
Tokenization into word forms is provided by the respective corpora. Both the PUD and CV (forced alignments) provide their own splits of sentences in written and spoken form into tokens. There are some issues with tokenization which are relevant for our results. To illustrate this, take the same example sentence as above in Japanese.

\newpage

\noindent Japanese (jpn)\footnote{This (part of a) sentence is taken from the file ja\_pud-ud-test.conllu (sent\_id = n01004009). The tokenization is here taken directly from the PUD. The transliteration into Romaji is taken from the repository of \textcite{Petrini2022b}. The glossing is provided with reference to the online dictionary at \url{https://jisho.org/} as well as the Japanese grammar by \textcite{Akiyama2002}. PRT: case particle, here to be translated as `by'; TOP: topic marker (similar to nominative case); POSS: possessive marker (similar to genitive case); OBJ: direct object marker (similar to acusative case).} 
\begin{exe}
  \ex\label{example_Japanese}
  それ適用することで、生徒は科学の内容を学習する。
  \glll それ 適用 する こと で 、 生徒 は 科学 の 内容 を 学習 する 。\\
        sore tekiyou suru koto de , seito wa kagaku no naiyou o gakusyuu suru .\\
        it applying to.carry.out matter PRT , student TOP science POSS content OBJ learning to.carry.out\\
  \glt `Students learn science content by applying it.'\\
\end{exe}

In Japanese, particles marking topic (は \textit{wa}), possession (の \textit{no}), and direct objects (を \textit{o}) are here treated as separate words, while another grammatical analysis might treat them as affixes to the respective nouns (e.g. 生徒は \textit{seito-wa}), hence creating longer word types.

% https://ufal.mff.cuni.cz/~zeman/langtech/npfl120/slides/npfl120-03-tok.pdf
% added by Ramon 18-9-2023
In general, the UD corpus follows a syntactic definition of ``word''. According to its designers, ``the basic units of annotation are syntactic words (not phonological or orthographic words), which means that we systematically want to split off clitics, as in Spanish {\em dámelo} = {\em da me lo}, and undo contractions, as in French {\em au} = {\em à le}''.\footnote{\url{https://universaldependencies.org/u/overview/tokenization.html}.} Although it is possible to recover the unsplit versions from the PUD corpora, we here use only the split versions. After all, this is the default design choice of UD. Also, this will enable further research on the interplay between word length and syntactic dependency structures \parencite{Ferrer2021a}.\footnote{Due to a difference in pre-processing, both the split and the unsplit versions contributed word tokens in the analyses of \parencite{Petrini2022b}. Here, only the split version contributes the word tokens for analyses.}

\subsection{Weak recoding problem}

When measuring word length in Latin script compared to strokes in Japanese or Chinese, we simply preserve the original character types but recompute their length, in agreement with the definition of the weak recoding problem we introduced in \autoref{sec:problems_on_communication_systems}. We do not recompute types according to distinct strings in Latin script (that would be the strong recoding problem, which is not the target of this article). % doing it in that way: future research
It is possible and expected that distinct character types are assigned the same string in Latin script. For instance, 
書く, 格, 角, 核, 欠く, 画, 各, 佳句, 確, 斯く, 舁く , 殻, 隔, 膈 
are all written as `kaku' in Romaji. Hence, Japanese romanizations do not comply with the setting of the strong recoding problem. 

\subsection{Immediate constituents in writing systems}
\label{sec:immediate_constituents}

When measuring word length in written languages, we are mainly using \textit{immediate constituents} of written words. In Romance languages, we are measuring word length in letters of the alphabet, the immediate constituents in the written form, which are a proxy for phonemes. For syllabic writing systems (as Chinese in our dataset), the immediate constituents of words are characters that correspond to syllables. 
% The only exceptions are
In addition, for 
Chinese and Japanese, we are considering
% three 
two other possible word length units:
% characters, 
strokes and letters in Latin script romanizations. In the hierarchy from words to other units, only the original characters are immediate constituents. This means that, for each of these languages, words are unfolded into three systems, one for each unit of encoding (original characters, strokes, romanized letters/characters). For simplicity -- and to avoid that these two languages are hence over-represented in statistical analyses on the correlation between scores and certain parameters at the level of languages (\autoref{sec:other_desirable_properties}) -- we will use only their immediate constituents (namely original characters). 
% for simplicity. 
Thus, the results based on original characters are compared with those for the other languages.  

% \subsection{Statistical testing}

% When measuring the association between two variables, we use both Pearson correlation and Kendall correlation \parencite{Conover1999a}. Note that the traditional view of Pearson being a measure of linear association and Kendall correlation being a measure of monotonic non-linear association has been challenged \parencite{vandenJeuvel2022a}. 

% Given the long computational time required for Kendall $\tau$ correlation, the naive default algorithm $O(n^2)$ for computing $\tau$ was replaced by one that runs in $O(n\log n)$ time \parencite{Christensen2005a} and is 
% available in R\footnote{\url{https://www.rdocumentation.org/packages/pcaPP/versions/2.0-1/topics/cor.fk}}.

% We applied a Holm-Bonferroni correction to $p$-values\footnote{\url{https://stat.ethz.ch/R-manual/R-devel/library/stats/html/p.adjust.html}}
% in correlograms, where a correlation test was applied to all pairs of random variables from a specific set. 

%%%%% Results

\section{Results}
\label{sec:results}

In \autoref{sec:introduction}, we highlighted the importance of distinguishing between principles and their manifestations. In \autoref{sec:measuringOptimality}, 
we have addressed the problem of how to measure the degree of compression of word lengths through the relationship between the frequency of a type and its length, examining distinct scores for the optimality of word lengths. In \autoref{app:theory}, we show that $\eta$, $\Psi$, and $\Omega$ are indices of the intensity of compression because they always reach a value of
\textit{one} when the system is fully optimized according to the minimum baseline (rank ordering) in \autoref{sec:baselines}. In contrast, Pearson $r$ and Kendall $\tau$ correlations fail to give a constant value when the system is fully optimized. Therefore, $r$ and $\tau$ are suitable to investigate the manifestation of compression, i.e. the law of abbreviation \parencite{Petrini2022b}, but are a poor reflection of the actual degree of optimality with respect to the minimum baseline. 

In this results section, we first provide some intuitive understanding of the optimality scores. Secondly, we establish that $\Psi$ is the best score in terms of mathematical properties (\autoref{tab:mathematical_properties_of_scores}), and other desirable properties (\autoref{sec:other_desirable_properties}). Thirdly, we choose $\Psi$ accordingly to measure the degree of optimality of languages. Lastly, we compare $\Psi$ against the other optimality scores (excluding the correlation scores, $r$ and $\tau$, for the reasons mentioned above). 

\subsection{Understanding the optimality scores}

\begin{figure}[H]
  \centering
     \includegraphics[width = 0.9\linewidth]{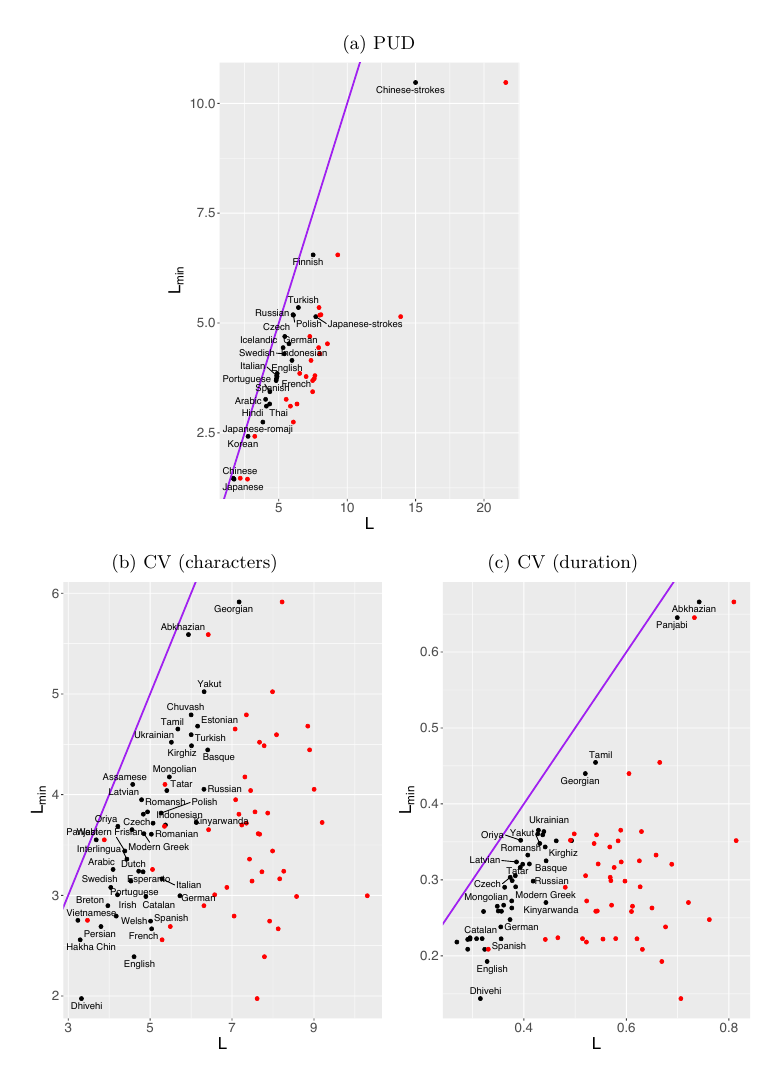}
     \caption{\label{fig:scores_ingredients_a} 
    The ingredients of the ratio $\eta$: $L_{min}$ (the minimum baseline) versus $L$ (the actual word length). The purple line is the identity function (slope of 1 and zero intercept). 
    Red points indicate the points that are expected under the null hypothesis (where $L = L_r$ is expected).
    (a) PUD collection. (b) CV collection with length measured in characters. (c) CV collection with length measured in duration. 
    }
\end{figure}

\begin{figure}[H]
    \centering
    \includegraphics[width = 0.9\linewidth]{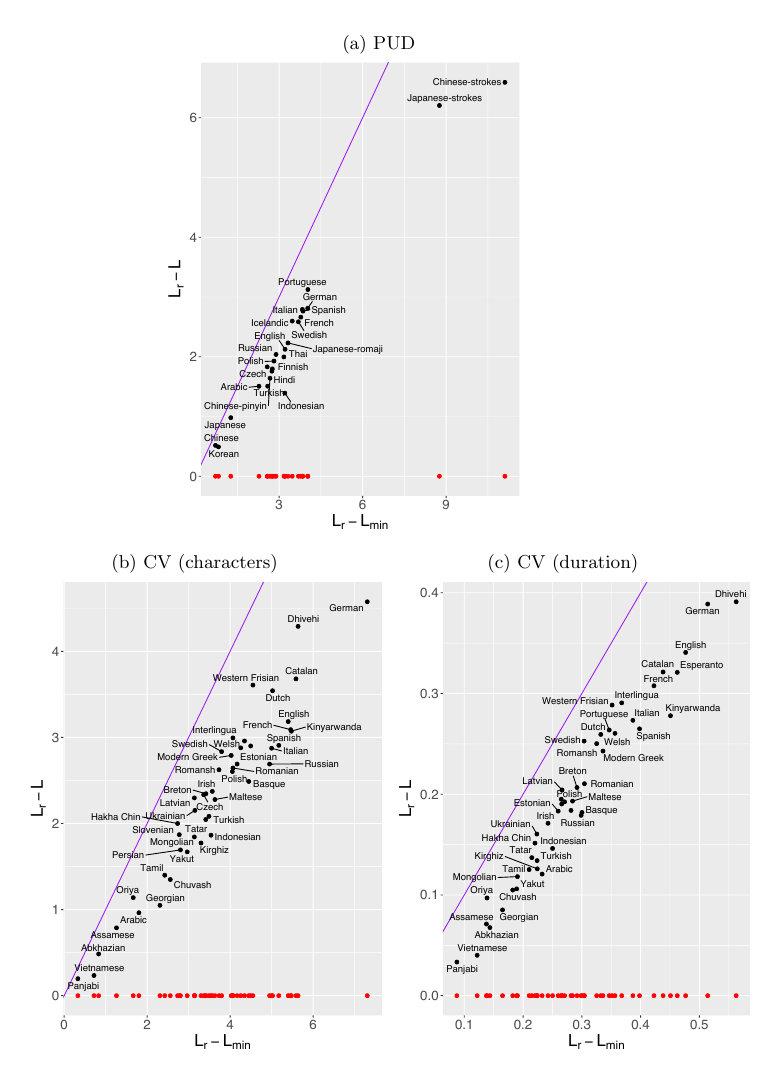}
    \caption{\label{fig:scores_ingredients_b} 
    The ingredients of the ratio $\Psi$: the gap $L_r - L$  versus the gap $L_r - L_{min}$. The purple line is the identity function (slope of 1 and zero intercept). 
    Red points indicate the points that are expected under the null hypothesis (where $L = L_r$ is expected).
    (a) PUD collection. (b) CV collection with length measured in characters. (c) CV collection with length measured in duration. 
    }
\end{figure}

\begin{figure}[H]
    \centering
    \includegraphics[width = 0.9\linewidth]{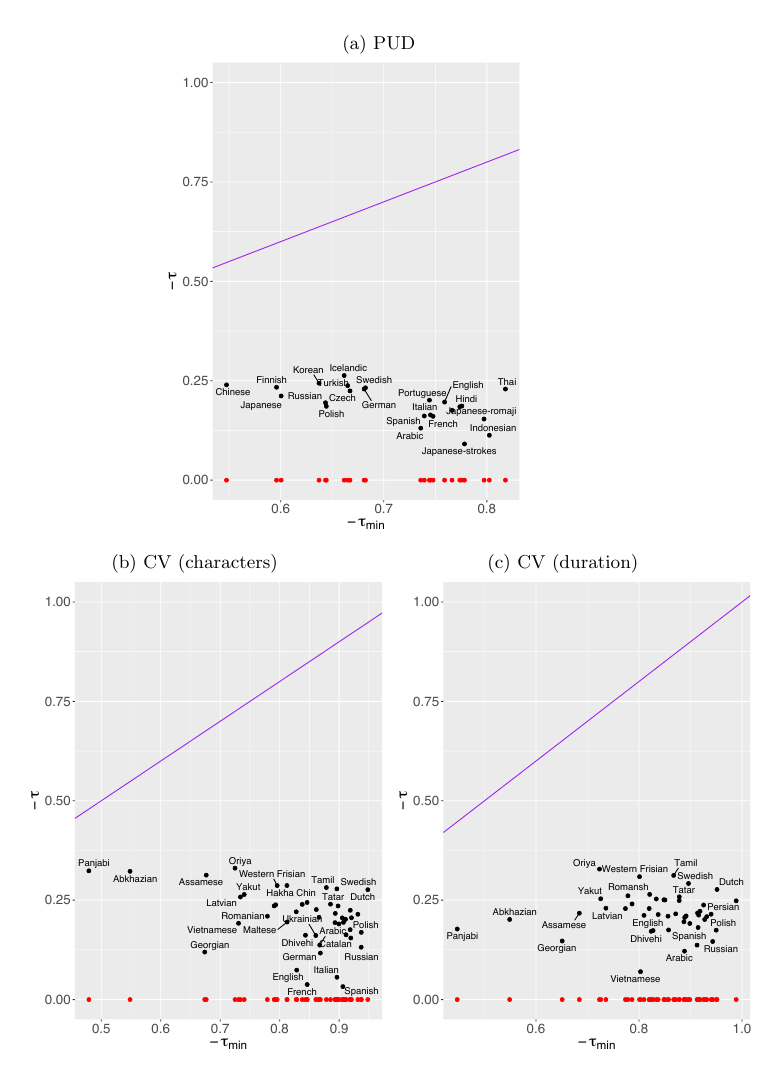}
    \caption{\label{fig:scores_ingredients_c} 
    The ingredients of the ratio $\Omega$: negative Kendall $\tau$ correlation between frequency and length, versus the negative $\tau_{min}$, the minimum baseline for $\tau$.
    Here we use the definition of $\Omega$ \autoref{eq:Omega_raw} with $\tau_r = 0$, hence the minus sign in the correlations displayed by each axis. 
    Red points indicate the points that are expected under the null hypothesis (where $\tau = \tau_r = 0$ is expected).    
    The purple line is the identity function (slope of 1 and zero intercept). (a) PUD collection. (b) CV collection with length measured in characters. (c) CV collection with length measured in duration. 
    }
\end{figure}

Here we focus on a qualitative understanding of the optimality scores $\eta$, $\Psi$ and $\Omega$ as a preparation for the results that are presented below. All these scores are ratios of the form $y/x$ (\autoref{eq:eta}, \autoref{eq:Psi}, and \autoref{eq:Omega}). That is, they give a \textit{percentage} of word length optimization with respect to some baseline(s). \autoref{fig:scores_ingredients_a},
\autoref{fig:scores_ingredients_b},
and \autoref{fig:scores_ingredients_c} show $y$ as a function of $x$ for each of them. 
A deeper understanding of the expected behavior of these scores requires taking into account the mathematical properties of the scores (\autoref{app:optimalityScores}) that are indicated in parenthesis for the mathematically oriented reader.
By definition, all the data points have to fall below the identity line ($y=x$) because the scores are designed to satisfy $y \leq x$ (the maximum value of these scores is 1). 
In case of optimal coding (minimum word length), languages would fall on the identity line (because these scores exhibit constancy under optimal coding).
\autoref{fig:scores_ingredients_a},
\autoref{fig:scores_ingredients_b},
and \autoref{fig:scores_ingredients_c} indicate that languages are not coding optimally.

The expected behavior of the scores in other conditions depends on the score. 
For example, in the absence of any word length effect (for or against compression), languages are expected to be on the abscissa axis (the line $y=0$) in case of $\Psi$ and $\Omega$ (because these scores exhibit stability under the null hypothesis). In case of $\eta$, languages are going to be distributed according to $x = L= L_r$ (because $\eta$ is not stable under the null hypothesis).

In case of some compression of word lengths, languages must be over the line $y=0$ in case of $\Psi$ and $\Omega$ -- which is what these figures show. In the case of $\eta$, a comparison of its actual value against its expected value under the null hypothesis -- that is $L_{min}/L_r$ -- is required (because $\eta$ lacks stability under the null hypothesis).   
\autoref{fig:scores_ingredients_b}
and \autoref{fig:scores_ingredients_c} reveal the presence of some degree of compression. In fact, points are in general closer to the diagonal than to the abscissa axis.

% next paragraph addresses Adele's comments on twitter.

A central score in the analyses to come is $\Psi$. The percentage $\Psi$ yields is with respect to $L_r - L_{min}$, that is the gap between the \textit{minimum possible} mean word length (obtained by assigning the $i$-th most frequent word the $i$-th shortest length in a language and recomputing mean word length) and the \textit{random} mean word length (obtained by shuffling at random word lengths or word probabilities in a language). 
In more detail, $\Psi$ is the percentage of this gap ($L_r - L_{min}$) in relation to the gap $L_r - L$, i.e. the gap between the \textit{actual} word lengths and the \textit{random} mean word length. As an example, 
in Mandarin Chinese (when the units are Han characters), $\Psi = 72.3\%$ is the percentage of the gap $x=L_r - L_{min} = 2.18 - 1.47 = 0.71$ in relation to the gap $y = L_r - L = 2.18 - 1.67 = 0.41$) (\autoref{fig:scores_ingredients_a} and \autoref{tab:opt_scores_pud}).
% \Psi = (2.18 - 1.67)/(2.18 - 1.47)

Finally, notice that similar scores, or their components, have already appeared implicitly in previous research on the optimality of word lengths \parencite{Pimentel2021a}. In Figure 4 of \parencite{Pimentel2021a}, one finds the gap $L_r - L$, the numerator of $\Psi$ (\autoref{eq:Psi}), in the $x$-axis of the left panel as well as $1/\eta$ in the $y$-axis of the right panel (but we used {\em rank ordering} to define $L_{min}$ for $\eta$ as in \autoref{eq:eta}). 

%% NEW FIGURES

% b)

% c)

\subsection{Desirable properties of the scores}

In order to identify the best score for cross-linguistic %%%%%
research on the degree of optimality, we take into account what we consider to be desirable properties for an optimality score, in addition to those that can be established by a purely mathematical analysis (\autoref{tab:mathematical_properties_of_scores}). 

First, a score should not be related to the basic language parameters -- namely observed alphabet size (number of distinct characters, $A$), observed vocabulary size (number of types, $n$), and sample size (number of tokens, $T$) -- which are assumed to be constant in the definition of the baselines. The only score consistently satisfying these properties across all collections and length definitions is $\Psi$, while both $\eta$ and $\Omega$ show some degree of significant association with $A$ and $T$ when data is more heterogeneous, i.e. in the CV collection (\autoref{fig:correlograms_params}). 

Second, the optimality score of a given language should not depend too strongly on the size of its available sample,
meaning that it should converge, and ideally it should converge fast. Again, the score which gets closest to this ideal behaviour is $\Psi$. It varies within a rather small range, and appears to approach convergence in different languages. In contrast, both $\eta$ and $\Omega$ keep their decreasing trend even for large sample sizes, yet with different speeds (\autoref{fig:convergence_pud}, \autoref{fig:convergence_cv_characters} and \autoref{fig:convergence_cv_medianDuration}).

Third, a complex score should not be replaceable by a simpler score, namely it should not be significantly and strongly associated to it. When using parallel data, none of our scores is replaceable by $L$, the simplest score. We only find that $\Omega$ could be replaced by $\eta$ (\autoref{fig:correlograms_scores}). Thus $\Psi$ is the best complex score for parallel data. %Solid conclusions for non-parallel data are beyond the scope of this article. 

See \autoref{app:desirable} for further details on the analysis of the desirable properties.

\subsection{The distribution of optimality values across languages}

\begin{figure}[H]
  \centering
     \includegraphics[width = 0.9\linewidth]{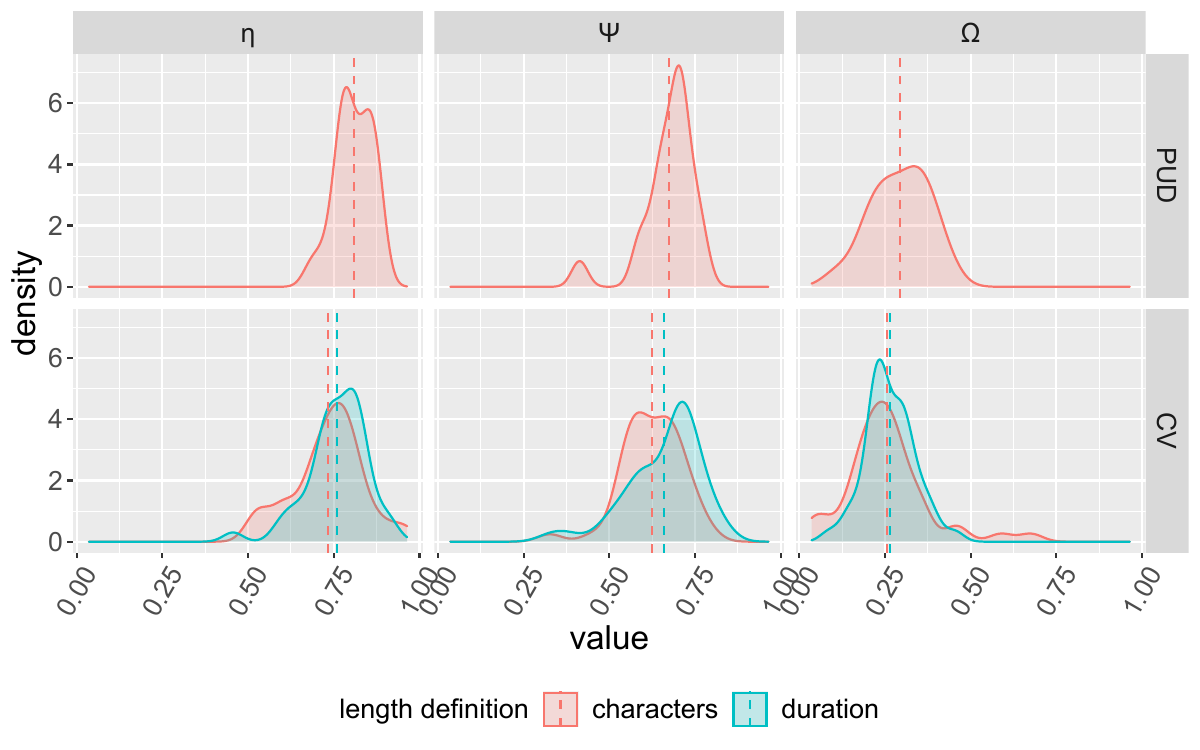}
     \caption{\label{fig:opt_scores_density} Density distribution of the scores $\eta$, $\Psi$ and $\Omega$ for length in characters or duration over languages in the PUD and CV collections. For length in characters, we only use immediate word constituents for the sake of homogeneity. Accordingly, scores in strokes or in romanizations for Chinese and Japanese in PUD are excluded. Refer to \protect \autoref{tab:opt_scores_pud},
\autoref{tab:opt_scores_cv_characters} and \autoref{tab:opt_scores_cv_meadianDuration}     
     for values on individual languages. % before it read "these values".      
     The vertical dashed lines show mean values (the values of the means are shown in \protect \autoref{tab:opt_summary}).\\
    }
\end{figure}

Here we review the distributions of optimality values for each of the scores.
For the scores on word length in characters, we excluded strokes and romanizations for Chinese and Japanese for the sake of homogeneity (i.e. we use only immediate written word constituents). In \autoref{fig:opt_scores_density}, we show the density plots for both collections, color-coded by definition of length. 
All scores show a rather narrow distribution, both in PUD and in CV, despite its heterogeneity and large size in terms of languages. % The bulk of the values is far from zero, the value expected under the null hypothesis, specially for $\eta$ and $\Psi$.
The bulk of the values is far from zero. However, this is only informative for scores whose expected value under the null hypothesis is zero, that is $\Psi$ and $\Omega$. The large observed values of $\eta$, on the other hand, are not informative in this sense. Interestingly, the values of $\Psi$ are farther from zero than those of $\Omega$.

In \autoref{tab:opt_summary}, we report the main summary statistics. In all cases, $\Psi$ and $\Omega$ % before it said "the scores"
take positive values suggesting some degree of optimization in every language. Indeed, 
% we have shown that the correlation tests support that $\Psi$ or $\Omega$ are significantly large in all languages (\autoref{sec:law_of_abbrebiation}). 
the correlation tests on the same data support that $\Psi$ or $\Omega$ are significantly large in each individual language \parencite{Petrini2022b}. 
The magnitude strongly depends on the score used. Concerning $\Psi$, our preferred score for cross-linguistic analysis, the mean values computed in the three considered scenarios (in a range from 0 to 100\%) are close to each other: 67\% in PUD with length measured in characters, 62\% in CV when considering characters, and 66\% when considering duration.  

In \autoref{tab:opt_scores_pud}, \autoref{tab:opt_scores_cv_characters} and \autoref{tab:opt_scores_cv_meadianDuration}, we detail the values of all scores and those of their ingredients for each language and collection. 

% summary statistics of Omega and eta
\begin{table}[H]
\centering
\caption{Summary statistics of optimality scores $\eta$, $\Psi$ and $\Omega$ for every collection and definition of length. 
For length in characters, we only use immediate word constituents for the sake of homogeneity. Accordingly, scores in strokes or in romanizations for Chinese and Japanese in PUD are excluded. Refer to \protect \autoref{tab:opt_scores_pud},\autoref{tab:opt_scores_cv_characters} and \autoref{tab:opt_scores_cv_meadianDuration} for values on individual languages. % before it read "these values".
} 

\label{tab:opt_summary}

\begin{tabular}{rr|rrrrrrr}
score & collection & Min. & 1st Qu. & Median & Mean & 3rd Qu. & Max. & sd\\
\hline
\multirow{3}{*}{$\eta$} 
% latex table generated in R 4.0.5 by xtable 1.8-4 package
% Wed Sep 06 18:23:40 2023
  & PUD-characters & 0.69 & 0.78 & 0.80 & 0.81 & 0.85 & 0.88 & 0.05 \\ 
   & CV-characters & 0.52 & 0.69 & 0.75 & 0.73 & 0.80 & 0.96 & 0.10 \\ 
   & CV-medianDuration & 0.46 & 0.72 & 0.76 & 0.76 & 0.81 & 0.92 & 0.09 \\ 
   \hline

\multirow{3}{*}{$\Psi$}  % latex table generated in R 4.0.5 by xtable 1.8-4 package
% Wed Sep 06 18:23:40 2023
  & PUD-characters & 0.41 & 0.65 & 0.69 & 0.67 & 0.71 & 0.78 & 0.08 \\ 
   & CV-characters & 0.33 & 0.57 & 0.63 & 0.62 & 0.68 & 0.79 & 0.09 \\ 
   & CV-medianDuration & 0.33 & 0.60 & 0.69 & 0.66 & 0.73 & 0.83 & 0.11 \\ 
   \hline

\multirow{3}{*}{$\Omega$} % latex table generated in R 4.0.5 by xtable 1.8-4 package
% Wed Sep 06 18:23:39 2023
  & PUD-characters & 0.11 & 0.23 & 0.29 & 0.29 & 0.35 & 0.44 & 0.08 \\ 
   & CV-characters & 0.04 & 0.19 & 0.24 & 0.26 & 0.30 & 0.68 & 0.12 \\ 
   & CV-medianDuration & 0.09 & 0.22 & 0.25 & 0.26 & 0.31 & 0.45 & 0.07 
\\
\end{tabular}
\end{table}

% density plots of scores

\subsection{Sorting languages by their degree of optimality}

Here we focus on PUD, as it is the only parallel corpus, and on $\Psi$, for the sake of simplicity given its mathematical and statistical properties. Sorting languages in CV might lead to misinterpretations, as its intrinsic heterogeneity hampers valid cross-linguistic comparisons (see \autoref{app:desirable} for problems and risks of drawing conclusions from CV).

\begin{figure}[H]
    \centering
    \includegraphics[width = 0.9\linewidth]{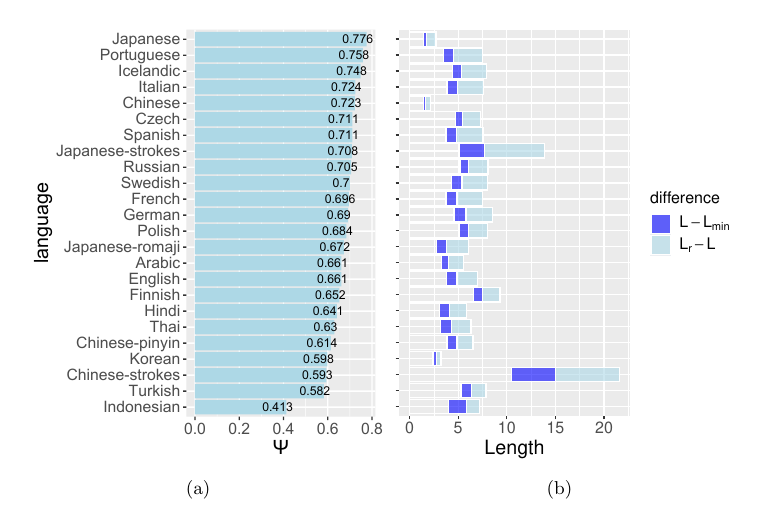}
    \caption{\label{fig:psi_pud_characters} Optimality of word lengths in PUD according to $\Psi$ with length measured in characters. (a) Absolute values of $\Psi$. (b) Visual depiction of the composition of $\Psi = (L_r - L)/(L_r - L_{min})$ using bars that consist of two segments, a dark blue one followed by a light blue one. 
    % in PUD collection with length measured in characters. 
    The bar starts at $L_{min}$, changes color at $L$, and ends at $L_r$. The dark blue segment of the bar is the difference $L-L_{min}$. This quantity is positive because $L_{min} \leq L$ by definition. The light blue segment is the difference $L_r - L$. This quantity is positive because $L < L_r$ due to compression \parencite{Petrini2022b}.
    hence the length of the whole bar is the difference $L_r - L_{min}$, and the length of the light blue segment over the whole represents $\Psi$.
    }
\end{figure}

In \autoref{fig:psi_pud_characters}(a), we show the languages in the PUD collection sorted decreasingly by $\Psi$. In \autoref{fig:psi_pud_characters}(b), we display a breakdown of the composition of $\Psi$ intended to help understand its definition as a ratio of two differences.  

Tables showing the concrete values of the scores and further details on both collections can be found in \autoref{app:opt_scores}.

% plots of PSI and its composition
% figure 5

\subsection{The weak recoding problem} 

\subsubsection{Length in characters versus duration in the same language.}

\begin{figure}[H]
    \centering
    \includegraphics[width = 0.9\linewidth]{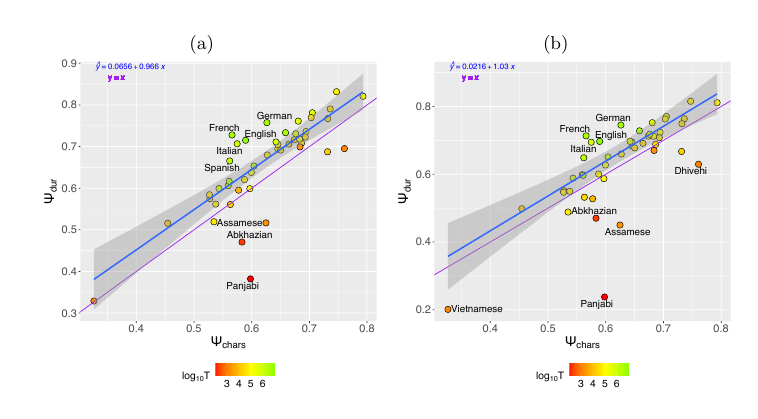}
    \caption{\label{fig:psi_timeVSspace} $\Psi$ measured in duration ($\Psi_{dur}$) versus $\Psi$ measured in characters ($\Psi_{chars}$) in the CV collection. Languages are color-coded by number of tokens in logarithmic scale to indicate orders of magnitude in sample size. We fit a robust linear model by Theil-Sen 
    %least squares 
    regression (blue line), which is then compared to the identity function, $\Psi_{dur}=\Psi_{chars}$ (purple line).     95\% confidence intervals for the regression line are shown as a gray band.
    The choice of robust linear regression is motivated by the presence of undersampled languages whose scores 
    deviate from the remainder of languages in this collection.
    For written language, only immediate word constituents are used (Japanese and Chinese are not included in CV). We report the parameters of the linear model (slope and intercept), Pearson correlation ($r$) and the standard error of the regression ($S$). 
    (a) Word length defined as median duration. $y = 0.066 + 0.966x$, $r = 0.794$ with $p$-value $4.5\times 10^{-11}$ and $S=0.227$. (b) Word length defined as mean duration. $y = 0.022 + 1.03x$, $r=0.745$ with $p$-value $3.0\times 10^{-9}$ and $S = 0.239$.
    }
\end{figure}

To understand the relation between the optimality of a language in written and in oral form, we compare the values of $\Psi$ when length is measured in duration 
% as a function of $\Psi$ 
against its values when length is measured in characters (\autoref{fig:psi_timeVSspace}). This comparison is possible only in the CV collection. In \autoref{fig:psi_timeVSspace}, we also show the outcome of fitting a linear model. Both when considering length in median and in mean duration, we find that the scores are generally preserved as supported by a significant and positive Pearson correlation over 0.74, and a linear regression that yields a slope near 1, as well as a small 
positive intercept. Interestingly, the great majority of languages show more optimization in oral than in written form (dots above the purple identity line). 
The intensity of this phenomenon seems to be related with sample size, in the sense that the languages with the smallest samples tend to be below the identity line. Recall, however, that we found no significant correlation between $\Psi$ and $T$ (\autoref{sec:other_desirable_properties}).

% omega time vs omega space

\subsubsection{Length in Chinese/Japanese characters versus other discrete lengths (Latin script characters and strokes).}

Here we analyze the impact of replacing word length in Chinese and Japanese characters, the immediate word constituents in written form, by word length in strokes or Latin script characters after romanization. 
We find that the value of $\Psi$ reduces in all cases (\autoref{fig:psi_pud_characters}). However, the value of the optimality score is still significantly large according to the corresponding correlation test, even after the replacement % (\autoref{fig:corr_significance} (a, c)).
\parencite[Figure \figcorrsignificance~(a, c)]{Petrini2022b}.
In particular, romanization affects both languages in a comparable way (scores decrease by 15\% in Chinese, and 13\% in Japanese), while measuring length in strokes has a stronger effect in the case of Chinese (18\% decrease) compared to Japanese (9\% decrease).

\subsubsection{Removal of vowels in Latin script languages or romanizations.}

\begin{figure}[H]
    \centering
    \includegraphics[width = 0.9\linewidth]{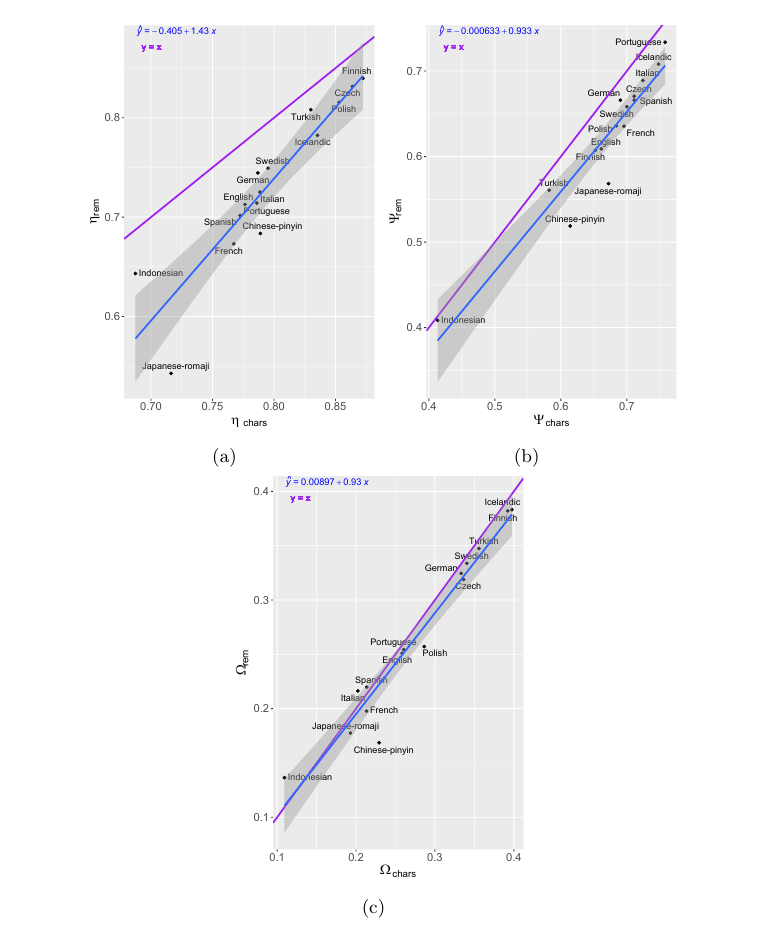}
    \caption{\label{fig:score_comparison} 
    The value of a score after removing vowels, indicated with the subindex {\em rem} (e.g. $\Psi_{rem}$), as a function of the original value with all characters (e.g. $\Psi_{chars}$) in languages using the Latin script in the PUD collection. For a given language, the new value of the score is obtained after removing vowels that may include different accents or tones. 
    The purple line is the identity function (slope of 1 and zero intercept), while the blue line is the least squares linear regression. 95\% confidence intervals for the regression line are shown as a gray band.
    For each score, we indicate the parameters of the best fit of a linear model (slope and intercept), %the coefficient of determination ($R^2$) 
    as well as the Pearson correlation coefficient ($r$) and the standard error of the regression ($S$). 
    (a) $\eta$ score. $y = - 0.405 + 1.43 x$, % $R^2=0.834$, 
    $r = 0.919$ with $p$-value $1.3 \times 10^{-6}$ and $S = 0.170$. 
    (b) $\Psi$ score. $y = - 0.001 + 0.93 x$, % $R^2 = 0.905$, 
    $r = 0.972$ with $p$-value $4.9 \times 10^{-8}$ and $S = 0.083$. 
    (c) $\Omega$ score. $y = 0.009 + 0.93 x$, % $R^2 = 0.930$, 
    $r = 0.952$ with $p$-value $1.4 \times 10^{-9}$ and $S = 0.062$. 
    }
\end{figure}

We also investigate the impact of removing vowels on the 15 languages using Latin scripts (including Chinese Pinyin and Japanese Romaji) in the PUD collection, comparing the original value of the score against its new value after this particular transformation.
\autoref{fig:score_comparison} shows the new scores as a function of the original ones, alongside the results of a linear regression. 
% For all three scores, the slope is very close to 1, but while for $\Psi$ and $\Omega$ the offset is almost 0, $\eta$ introduces quite some shifting in the scores of the re-coded data.
For both $\Psi$ and $\Omega$, the best fit of the linear model gives a slope very close to 1, and a small offset close to 0. In contrast, $\eta$ yields a slope of 1.4 and an intercept of -0.4.
Moreover, the standard error $S$ of its linear regression is more than twice the values found for $\Psi$ and $\Omega$. Undoubtedly, $\eta$ is not robust to the removal of vowels. Among all the scores, $\Omega$ has the highest Pearson correlation and the lowest standard error, suggesting it is the score that is the least sensitive to vowel removal.

% omega with and without vowels

\subsection{Impact of disabling the filter of words that contain ``foreign'' characters }

All results presented in this section have been obtained after applying a specifically developed method to filter out highly unusual characters and words -- as described % in \autoref{sec:filter}.
by \textcite{Petrini2022b}.
If the filter is disabled, we obtain some slight changes in the values, but the qualitative results remain the same.

%%%%% Discussion

\section{Discussion}
\label{sec:discussion}

The degree to which languages are optimized is receiving increasing attention \parencite{Ferrer2018b,Coupe2019a,Ferrer2020b,Koplenig2021a}.
Quite generally, linguistic research is subject to a tension between acknowledging the astonishing uniformity of languages at higher levels of abstraction, and the empirical diversity which languages exhibit in a myriad of structural features from phonology to syntax. In the domain of coding efficiency, it has been argued that distinct languages exhibit similar degrees of efficiency \parencite{Coupe2019a}. In this article, we contribute to each of these complementary views. Questions 1-2 are related to observations of homogeneity, whereas Question 3 and Question 4 shed some light on diversity. 

\subsection*{Question 1. What are the best optimality scores for cross-linguistic research?}

We measured optimality with three scores: $\eta$, $\Psi$, and $\Omega$. As shown in \autoref{tab:mathematical_properties_of_scores}, the two latter scores are endowed with better statistical and mathematical properties, and are thus more reliable than $\eta$. Concerning the comparison between the remaining two, we argue that $\Psi$ is endowed with some additional desirable properties which make it better suited for the scope of cross-linguist research.

\subsubsection*{Desirable properties.} First, an ideal score should not be sensitive to the basic language parameters that are assumed constant in the definition of the baselines, such as alphabet size, vocabulary size, and sample size.
To check whether this requirement holds for the considered scores, we computed the Kendall $\tau$ (and Pearson $r$) correlation
between them and the mentioned basic parameters (\autoref{fig:correlograms_params}). We want to stress that PUD is the only parallel corpus in the analysis, and the correlations computed in CV might suffer from confounding effects.

On the one hand, no significant linear or non-linear association could be detected for $\Psi$ -- a result consistent across collections and length definitions. On the other hand, both $\eta$ and $\Omega$ show some significant negative correlations with observed alphabet size 
and sample size in the CV collection. These two properties are, however, also strongly associated with one another. Indeed, the observed alphabet size depends on the amount of seen data (\autoref{fig:correlograms_params} (b,c,e,f)), and the correlation of $\eta$ and $\Omega$ with $A$ could potentially just be a side-effect of the relationships above, by transitivity. Nevertheless, the most critical correlation is the one observed with sample size $T$, as it implies that these two scores tend to give larger values in under-sampled languages, which we do not consider a desirable property for a score. This can also be observed when looking at the convergence speed of the scores. As shown in \autoref{fig:convergence_pud} for PUD, \autoref{fig:convergence_cv_characters} for CV when length is measured in characters, and \autoref{fig:convergence_cv_medianDuration} for CV when length is measured in duration, both $\eta$ and $\Omega$ keep their decreasing trend even for large sample sizes. In contrast, $\Psi$ generally varies within a smaller range, and importantly, it seems to reach a quite stable value in many languages, especially in CV where larger sample sizes are available. While the convergence analysis allows one to capture the evolution of the scores for different sample sizes within a language, it also gives an understanding of the behaviour of scores across languages with different sample sizes. In fact, since $\eta$ and $\Omega$ tend to decrease, they will be likely to have higher values in languages with a smaller sample, thus leading to potentially misleading conclusions in non-parallel data, or data with excessive heterogeneity in terms of sample size. Moreover, $\Psi$ and $\Omega$ have a very similar shape up to around $10^2$ tokens, after which the latter generally drops down. This highlights the importance of using parallel corpora, as well as large enough samples.

Of course, the $\Psi$ scores obtained on non-parallel data are still the reflection of optimization of a language represented by a certain doculect, rather than of the language in its entirety. However, this score is less likely to be influenced by the heterogeneity in terms of text sizes than $\Omega$ and $\eta$. For this reason, despite its relation with the Pearson correlation -- which poses a potential challenge on the ability of $\Psi$ to grasp a non linear relation between length and frequency -- we argue that this score is more suited for non-parallel data. Besides, $\Omega$ could theoretically have more power in detecting a non-linear association, but its observed relationship with the number of tokens in CV (characters) raises serious concerns about its reliability in non-parallel data. Moreover, the steep decreasing trend observed in the convergence analysis suggests that, even when dealing with texts of a comparable size, $\Omega$ will be more strictly related to text size.

Complementary arguments about the best score in the context of the recoding problems are given in Question 5.

\subsubsection*{Possible alternatives}
In our derivation of the new optimality scores for word lengths, namely $\Psi$ and $\Omega$, we followed the template to design a new score for dependency distances \parencite{Ferrer2020b}. Equivalent scores may be obtained by other means. We cannot exclude the possibility that there are simpler scores or existing scores that have the same statistical properties as our $\Psi$ score for word lengths. 

A promising candidate for future research is the $\gamma$ index, a variant of Kendall $\tau$ correlation \parencite{Goodman1963a}, whose estimator is 
$$\gamma = \frac{n_c - n_d}{n_c + n_d}.$$
The inverted sign $\gamma$, i.e. $-\gamma$, seems to have the same mathematical properties as $\Omega$ but may satisfy the desirable properties of $\Psi$. $\gamma$ is able to achieve $-1$ when the number of concordant pairs is $n_c = 0$ whereas $\tau$ only achieves that when ties are missing \parencite{Conover1999a}. 

\subsection*{Question 2. Is compression a universal principle? -- Or, are there languages showing no optimization at all?}

Despite the different definitions and properties, each of the three considered optimality scores reaches 1 when a language is fully optimized, while a value of 0 when there is no optimization at all is only expected for $\Psi$ and $\Omega$. The distribution of the values of the scores (\autoref{fig:opt_scores_density}) and, with greater detail, the summary statistics (\autoref{tab:opt_summary}), indicate that  $\Psi$ and $\Omega$ always take values that are larger than 0, with magnitudes depending on how 
% the score 
optimization is measured (the score and the units of length). This finding indicates that, globally, languages are shaped by compression (it is unlikely that all languages yield a positive value just by chance). However, to ensure that compression has shaped each of the languages, a test of the significance of the score is required.  
We will focus on $\Psi$ given our discussion above on the best score for cross-linguistic research. We have shown that testing if $\Psi$ is significantly high is equivalent to testing the law of abbreviation using a one-sided Pearson correlation test, 
% whose outcomes are shown in \autoref{fig:corr_significance} (c,d).
whose outcomes have already been shown \parencite[Figure 1 (c, d)]{Petrini2022b}. 

All languages in the sample show some significant degree of optimization in written form, independently of the units chosen to measure the latter. The test of $\Psi$ yields significance at the 99\% confidence level for all the languages in PUD. %In particular, the only language where the test of $\Psi$ does not yield significance at a 99\% confidence level is Chinese as measured in strokes in PUD. Notice that strokes are not immediate constituents of Chinese words and thus are not the most natural way of measuring word length in Chinese \parencite{Liu2017a}. 
Concerning oral form (duration), the only exception is Panjabi in the CV collection, for which we could not find a significantly small correlation. Thus its $\Psi$ score is not significantly large despite being greater than 0. This is likely related to severe under-sampling, as this particular sample contains only 98 tokens. Indeed, the only languages which are significant at the 95\% (rather than 99\%) confidence level are Abkhazian, Dhivehi, and Vietnamese, all having very small samples compared to the other languages of the CV collection. Therefore, we conclude that optimization is a universal principle, that manifests in our ensemble of languages when a large enough sample of a given language is available.

\subsection*{Question 3. What is the degree of optimization of languages?}

It has been argued that language production is not optimal but ``good enough'' \parencite{Goldberg2022a}. A necessary condition for the frequency-length association to be ``good enough'' is that the degree of optimality is statistically significant, which is the case % (\autoref{fig:corr_significance} (c,d)). 
\parencite[Figure \figcorrsignificance~(c, d)]{Petrini2022b}.
A further step is quantifying how ``good'' the mapping is, an issue addressed by inspecting the absolute value of $\Psi$ in a scale from 0 to $100\%$ as its values are significantly large. 

By looking at the summary statistics in
\autoref{tab:opt_summary} and at the probability density plots in \autoref{fig:opt_scores_density}, we can observe how the distribution of $\Psi$ is similar across the three scenarios (PUD, CV considering characters, CV considering duration). Indeed, their median, mean, and standard deviation values are comparable across collections and definitions of length. In particular, the mean value of $\Psi$ in parallel data is $67\%$, while for CV it ranges between $62\%$ (with length measured in characters) and $66\%$ (with length measured in duration). 
% Moreover, half of the languages in PUD show an optimization score larger than $70\%$, while this value slightly decreases to $69\%$ in CV with length measured in duration, and $63\%$ with length measured in characters. 
Moreover, half of the languages in PUD show an optimization score larger than $70\%$, while this value changes slightly to $69\%$ in CV when length is measured in duration, and drops to $63\%$ when length is measured in characters. 

This robustness suggests that the values of $\Psi$ that we observe are a reasonable approximation of the real degree of optimization of natural languages, at least within the scope of the families and scripts considered in this particular study. 

In sum, the degree of optimization of word lengths in languages ranges between $60\%$ and $70\%$ for the majority of the languages, both in parallel and non-parallel data. These findings are in line with the claim that languages exhibit a similar degree of coding efficiency in spite of their diversity \parencite{Coupe2019a}. Interestingly, our measurements of word length are on a scale from 0 to $100\%$ as a result of dual normalization, showing a high concentration of values within a narrow range. 

Finally, although we have excluded $\Omega$ to inspect the degree of optimization of languages for the sake of simplicity,
we would like to make a point about the possible origin of the low values (compared to $\Psi$) that $\Omega$ exhibits. We are certain that they are not due to the choice of $\tau$ as opposed to Spearman $\rho$ correlation in the definition of $\Omega$ (\autoref{eq:Omega}). 
See \autoref{app:variant_of_Omega} for further details and a further discussion.

\subsection*{Question 4. What are the most and the least optimized languages in our sample?} % Can we define a ranking of languages? -> future article

% Sonia read again
By means of $\Psi$, we are able to quantify the level of optimization of languages taking advantage of the fact that it is not correlated with the basic parameters ($A$, $n$, $T$) of each language. However, a truly ``fair'' comparison of the values of $\Psi$ in languages is hard to establish even given parallel texts. For one thing, text sizes in PUD are not large enough for $\Psi$ to reach convergence in all cases, meaning that the scores computed on our sample are likely not fully representative of the real ones (\autoref{fig:convergence_pud}). 
This still has to be kept in mind when comparing languages by optimization degree even knowing that $\Psi$ decays slowly once the text sample is sufficiently large.
Besides, large samples are available for CV but parallelism is lacking and sample sizes are very heterogeneous, making cross-linguistic comparisons problematic. For these reasons, here we focus on PUD and make emphasis on the relative order of the languages when sorted by $\Psi$ rather than on the concrete values, assuming that the ranks of languages by $\Psi$ are more stable than their absolute value of $\Psi$ when sample size is increased.

\autoref{fig:psi_pud_characters} shows the ranking of languages in PUD obtained by sorting their $\Psi$ values in decreasing order. Note that not every difference in the scores of two languages is necessarily significant. Future research should address the problem of the statistical significance of the differences between the optimality scores of languages \parencite{Ferrer2020b}.

Having said this, the languages vary in optimization between c. $41\%$ in Indonesian and c. $78\%$ in Japanese according to immediate word constituents (\autoref{fig:psi_pud_characters})\footnote{The range is based on immediate constituents of characters, this is why we are neglecting the minimum $\Psi$ achieved by Chinese characters in strokes, that is an even lower value than that of Indonesian.}. Japanese is the language showing the highest degree of optimization but only when word lengths are measured in characters (the immediate constituents of its written word forms), with $\Psi=77.6\%$. 
The writing systems used in Japanese, Chinese, and Korean clearly lead to lower values of $L_{min}$, $L$, and $L_r$ (when length is measured in characters) compared to other scripts, and are contained in a rather small interval (\autoref{fig:psi_pud_characters} (b)). However, these three languages have a very different position in the ranking, meaning that the reduced variability in word length is not {\em a priori} an advantage or a disadvantage. Together with Japanese, at the top
of the ranking we find Chinese and a cluster of Romance languages: Portuguese, Italian, Spanish, and French. Romance languages show a degree of optimality that ranges between $71-75\%$ (above English, that has $66\%$) and also show a very similar score composition (\autoref{fig:psi_pud_characters} (b)). 
% As anticipated in \autoref{sec:results}, recoding Chinese words in strokes leads to a value of $\Psi$ equal to $22.7\%$, making it the instance with the lowest optimization score. % Indonesian also shows a rather low value of $\Psi$ score ($42.8\%$) with respect to the bulk of the distribution. 

To get a better impression of the linguistic factors governing higher or lower optimality scores, we give examples of the most frequent and least frequent words of Japanese (high optimality), English (middle optimality), and Indonesian (low optimality) in \autoref{tab:examplesJapEngInd}. Given relatively constant content (parallel texts), Japanese has many high frequency particles of minimum length (length one in UTF-8 characters), and relatively few long words of length up to ten. In fact, all the longest words in this example are actually loanwords from German or English transliterated into Japanese characters. Note, though, that these loanwords do not bias the results, since they are likewise represented in all the other texts -- due to their parallel nature. In comparison, Indonesian uses particles of lower frequency\footnote{We here give raw frequency counts. The overall number of tokens in Japanese is c. 25 000, and in Indonesian c. 17 000. So the relative frequency of the most frequent particle in Japanese would be $\frac{1753}{25000}=0.07$, and for Indonesian $\frac{541}{17000}=0.03$. Hence, the observation that Indonesian has lower frequency particles also holds for \textit{relative frequencies}.} and generally higher lengths (both compared to original and romanized Japanese), and it has infrequent words of higher lengths reaching up to 20 characters. The latter is certainly related to the productive usage of prefixes and suffixes which are counted towards word length. It is possible that Indonesian and Japanese would fall closer together in the optimality range if Indonesian was written with a syllabary, and if Indonesian pre- and suffixes were interpreted as separate particles. 

% \textcolor{red}{The last sentence seems to contradict our proposal that certan scores abstract away from the characteristics of the writing system, e.g., alphabet size???}

Also, note that Japanese Kanji are borrowed from traditional Chinese Hanzi, but in Japanese, Hiragana and Katakana characters are added to the repertoire. The finding that written Japanese is more optimized than written Chinese when using the same unit (characters, romanizations and strokes) suggests that
syllabaries along with Japanese Kanji result in a higher degree of optimization of written language than the single use of Hanzi in Chinese, according to our metrics. However, this issue deserves further investigation as there is a hidden parameter that we are not controlling for: the level of granularity in the definition of a word in Chinese and Japanese (by level of granularity we mean the criteria to segment a sequence of characters into words). The current differences in optimality may reduce if the same level of granularity was used. 

\begin{table}[]
\begin{tabular}{lllllllll}
Japanese                       & $f_i$    & $l_i$  & English           & $f_i$    & $l_i$  & Indonesian  & $f_i$   & $l_i$  \\
\hline
の \textit{no} (poss. particle `of')& 1753 & 1  & the               & 1441 & 3  & yang (rel. pronoun `that')  & 541 & 4  \\
に \textit{ni} (loc. particle `at') & 1162 & 1  & of                & 620  & 2  & dan (conjunction `and') & 447 & 3  \\
は \textit{wa} (focus particle) & 1157 & 1  & in                & 510  & 2  & di (adposition `at') & 422 & 2  \\
た \textit{ta} (`did/do') & 918  & 1  & to                & 481  & 2  & nya (pronom. suffix) & 381 & 3  \\
を \textit{o} (direct object particle) & 851  & 1  & and               & 456  & 3  & untuk (adposition `for')  & 221 & 5  \\
... & ... & ... & ... & ... & ... & ... & ... & ... \\
パプアニュ一ギニア & 3    & 9  & conservationists  & 2    & 16 & mendokumentasikannya  & 1   & 20 \\
フォルクスワ一ゲン & 2    & 9  & catastrophically  & 1    & 16 & mendokumentasikan   & 1   & 17 \\
ディジボ一デンベルク & 1    & 10 & hydroelectricity  & 1    & 16 & mendemonstrasikan & 1   & 17 \\
オ一バ一マルスベルク & 1    & 10 & technologically   & 1    & 15 & mempertemukannya  & 1   & 16 \\
エンタ一テインメント & 1    & 10 & indistinguishable & 1    & 17 & menggulingkannya  & 1   & 16 \\
\end{tabular}
\caption{\label{tab:examplesJapEngInd}
Most frequent words and longest hapax legomena in Japanase, English and Indonesian texts of the PUD corpus. The Japanese hapax legomena can be interpreted as follows: パプアニュ一ギニア (`Papua New Guinea'); フォルクスワ一ゲン (`Volkswagen'); ディジボ一デンベルク (`Disibodenberg'); オ一バ一マルスベルク (`Obermarsberg'); エンタ一テインメント (`entertainment'). These interpretations are based on the Japanese-English dictionary at \url{https://jisho.org/}. The Indonesian hapax legomena can be interpreted as follows: men-dokument-asi-kan-nya: \textit{men-} is a verbal prefix for active voice, \textit{dokument} is a loanword, i.e. `document', \textit{-asi} verbal suffix especially for loanwords,  \textit{-kan} verb suffix (active/passive voice), \textit{-nya} suffix with various functions, here likely pronominal, i.e. `document this/it'. These interpretations are based on the Indonesian-English dictionary at \url{https://dictionary.cambridge.org/dictionary/indonesian-english/} as well as the Indonesian reference grammar by \textcite{Sneddon2010}.}
\end{table}

\subsection*{Question 5. What is the optimality of languages after recoding?}

We investigated the effect of recoding on optimization by assessing the degree to which ranks of scores are preserved when the respective recoding transformation is applied. In particular, we addressed the issue from three different viewpoints.

% 1
\subsubsection*{Length in characters versus duration in the same language}

By means of the CV corpus, we were able to explore the effect on optimality of measuring a word length in duration rather than in characters. 
% There is a main linear trend for the relation between the scores in the two forms (\autoref{fig:psi_timeVSspace}). Undersampled languages deviate from that trend. 
Most languages are above the identity line, indicating that they show more optimization in word durations (\autoref{fig:psi_timeVSspace}). Undersampled languages deviate from that trend. 
Especially when median duration is used (\autoref{fig:psi_timeVSspace} (a)) the slope is very close to 1 (0.966), and the intercept is small (0.066), meaning that the scores are simply shifted up by a small amount.

% In addition, the shift in scores is generally towards higher values, namely that most languages show more optimization in oral form. 
However, there is a positive relation between the number of tokens and the extent to which a language is more optimized in oral rather than written form, as measured by the ratio $\Psi_{dur}/\Psi_{chars}$.
Notice that not only the undersampled languages are below the diagonal, but also the larger the sample the higher the ratio $\Psi_{dur}/\Psi_{chars}$ (the Pearson correlation between $T$ and the ratio $\Psi_{dur}/\Psi_{chars}$ is $r = 0.389$; $p$-value 0.008). Therefore, the larger the sample, the stronger the evidence for word durations being more optimized than word lengths. The opposite finding in certain languages (lower  optimality for word durations) is likely due to undersampling.
% Thus the larger the sample, the larger the proportion between the scores for each unit of word length. Hence we cannot exclude the possibility that the 
% optimality 
% gaps between optimality of word length in character and optimality of length in duration are partially due to the heterogeneity in sample size of CV. 

% Ramon: I like the proposal. My point was about the existence of a correlation between the ratio and sample size, but not with each of the components of the ratio and sample size. Which I could not understand fully, but I still believe the correlation found with the ratio is a relevant point, which I find to be addressed in the current version.

These considerations notwithstanding, it has not escaped our attention that French, a language whose alphabetic writing system is commonly observed to deviate from actual phonetic form, exhibits one of the highest contrasts between optimality in written form and optimality in oral form, with the latter being considerably higher. However, we cannot exclude that this finding is partly driven by the larger sample size of French with respect to other Romance languages (\autoref{tab:coll_summary_cv}).

\subsubsection*{Length in Chinese/Japanese characters versus other discrete lengths (Latin script characters and strokes)}

We also recoded Chinese and Japanese using strokes and romanizations -- instead of the original characters of the main analyses. This way we investigate the effect of measuring lengths in different units. 
% what follows is to address L. Deboswki comments on twitter
It is tempting to jump to the conclusion that word lengths have to be more optimized given Chinese/Japanese characters than given romanizations or strokes. Namely, there is an obvious increase in plain word lengths for the latter. However, this line of thinking sweeps under the carpet that, after recoding, the random and minimum baselines have also changed. For this reason, the results given optimality scores built upon these baselines are hard to intuit {\em a priori.} 

Having said this, we indeed find a drop in the optimality scores for both alternative discrete lengths in both languages (\autoref{fig:psi_pud_characters}), yet the degree of optimization is still significant 
% (\autoref{fig:corr_significance} (c)). 
\parencite[Figure \figcorrsignificance~(c)]{Petrini2022b}. 
While romanization leads to a comparable effect in decreasing the scores in the two languages (the optimality of Japanese drops from $78\%$ to $67\%$ and that of Chinese drops from $72\%$ to $61\%$), the impact of using strokes as a unit leads to a larger % before "stronger" 
effect in Chinese (Japanese drops from $78\%$ to $71\%$ while Chinese drops from $72\%$ to $60\%$), as shown in \autoref{fig:psi_pud_characters}. This is likely related to Japanese Hiragana and Katakana characters being made up of fewer strokes than Chinese Hanzi characters. 
% The decreasing scores of romanization .

% Since most of the Chinese characters have one syllable, so 

% Additionally, both are using the same character sets, as we use the CJK characters, including Chinese (Chinese Hanzi and Japanese Kanji) with the exception of Japanese's other two writing systems, Hiragana and Katakana. This suggests that the use of syllabaries along with Japanese Kanji results in a more optimized writing system than only using Chinese Hanzi, according to our metrics.

% Besides, the \textcolor{marked} drop of optimality scores in Chinese measured in strokes compared with in Pinyin (romanization) reflect a fact that Chinese are mostly typed using Pinyin rather than handwriting.
Overall, these results suggest that compression operates on characters rather than on strokes. Characters (or components within characters) seem to form units within which the number of strokes is less relevant for efficiency considerations. Finally, notice that the Latin-like scripts that are commonly used to type characters on digital devices (cell phones or computers) in Chinese or Japanese, do not reach the same level of optimality as the original characters. A direct translation of Chinese/Japanese characters into a Latin alphabet plus some additional diacritics results in a reduction of the optimality of these languages in written form. 
% Chris, new text follows

Of course, we should not forget that our optimality scores ultimately reflect the link between frequency and length of words. In some writing systems this link is stronger, in others it is weaker. How exactly this relates to learning how to read and write in these systems is a separate matter. For example, in the case of writing characters, bear in mind that Chinese and Japanese speakers cannot, in general, type a single Hanzi or Kanji character by touching just one key. For that reason, in the context of writing, the comparison would be fair if they were always able to type a single character by touching just one key. All these considerations together, suggest that the pathways Western and Eastern writing systems took are not arbitrary, but (to some extent) guided by coding efficiency at the level of immediate constituents of words.

\subsubsection*{Removal of vowels in Latin script languages or romanizations}

We finally consider a scenario in which recoding implies applying a decreasing transformation to word lengths, namely removing vowels. 
Given the invariance properties of $\Psi$ and $\Omega$ (that $\eta$ lacks; \autoref{tab:mathematical_properties_of_scores}), we would expect them to yield approximately the same values for the original and the new scores computed by excluding vowels (while $\eta$ be less robust to that transformation).
% As shown in \autoref{fig:score_comparison} the regression slopes estimated for all the three scores are close to 1, suggesting that not only for $\Psi$ and $\Omega$ but even for $\eta$ the scores are kept to some degree when vowels are removed. 
Consistently, the linear regression slopes for both $\Psi$ and $\Omega$ are close to 1, while that of $\eta$ is about 1.4. Furthermore, $\eta$ has both the largest standard regression error and the largest intercept (in absolute values). 
% Thus, the new ranking obtained with $\eta$ is likely to be different from the original one due to the observed instability.

All the scores experience a decrease after the removal of vowels. Note that this is counterintuitive. Removal of vowels generally shortens words, i.e. $L$. We might jump to the conclusion that a writing system without vowels must be more optimal, but this is not the case.

Importantly, $\Psi$ and $\Omega$ show greater stability, but the latter seems to be the most capable of preserving the scores under weak recoding, having the lowest standard error, the highest Pearson correlation, and the smallest intercept. This is reasonable, given that $\Omega$ is endowed with invariance also under non-linear transformations, and there is no reason to assume that vowels would reduce word lengths in a linear way. Indeed, while on average a syllable contains one vowel, syllables can have different lengths, and thus be affected in different ways by the vowels' removal. % In conclusion, $\Psi$ and $\Omega$ are able to keep the scores values and ranking to some degree, but $\Omega$ turns out to be the most accurate and stable in doing so.  
In conclusion, $\Psi$ and $\Omega$ are both robust in the face of vowel removal, with $\Omega$ being most robust in this respect.  

An important conclusion is that we have demonstrated the existence of scores which abstract away from certain fundamental characteristics in the design of a writing system, e.g. how vowels are coded (abjads as in Standard Arabic, where only consonants are coded, or the Latin script in Romance languages, which codes for both consonants and vowels).
Crucially, the choice of the optimality score can be guided by the writing characteristics which we want to abstract from. In the larger picture, this helps with resolving the customary tension between seeing languages as uniform, or seeing each language as unique, a tension which characterizes language research. 

\subsection*{Question 6. Why do languages deviate from optimality?}

We found that word lengths in languages are not $100\%$ optimal. In previous research on the optimality of word lengths, 
% we discussed what prevents languages from reaching optimality using as minimum baselines optimal non-singular coding and optimal uniquely decodable encoding 
we have argued that perceptibility and distinguishability factors as well as phonotactic constraints are likely to prevent languages from reaching theoretical optimality \parencite{Ferrer2019c}. Along these lines, it has been demonstrated that the gap between actual average word length and the optimal one vanishes after controlling for morphological and graphotactical constraints \parencite{Pimentel2021a}. However, while these constraints might, for words in isolation, to a certain extent explain the actual level of compression, it is yet to be understood 
% \begin{enumerate}
% \item \textcolor{red}{Why even neglecting phono/graphotactic and related constraints, it is still possible to  succesfully predict the actual} relationship between word length and frequency rank \textcolor{red}{by assuming} (a) optimal non-singular coding, or (b) a combination of Zipf's rank-frequency law and optimal uniquely decodable encoding \parencite{Hernandez2019a,Torre2019a}.  
% \item Why languages with shorter syntactic dependencies between words also tend to have shorter words \parencite{Ferrer2021a}. This suggests that the online memory pressures leading to syntactic dependency distance minimization are also responsible for compression of word lengths. \textcolor{red}{Thus, compression seems to stem from cognitive constraints beyond word units.}
% \end{enumerate}
why languages with shorter syntactic dependencies between words also tend to have shorter words \parencite{Ferrer2021a}. This suggests that the online memory pressures leading to syntactic dependency distance minimization are also responsible for compression of word lengths. Thus, compression seems to stem from cognitive constraints beyond word units.

In addition to these factors acting on individuals (a speaker or a listener), the nature of the cultural process by which languages are optimized may also divert them from full optimality \parencite{Kanwal2017a}. Languages are distributed and conventional communication systems. Reaching full optimality would imply coordinated changes in a system that may not be possible in a real community of speakers. 
% From a cultural perspective, languages are conventional distributed communication systems 
% that emerge and evolve in a very wide geographical span, and in an unlimited temporal dimension. 
% As their evolution represents an online process, there is no possibility of global optimization, and despite continuously changing they have a great dependence on the past. Given these factors, the creation of a fully optimized system seems hardly feasible.

In this article, we have considered a minimum baseline that preserves the strings that are already being used by a language and simply attempts to reorder their corresponding lengths so as to minimize average word length. Once a communication system has been established, such reordering is a serious challenge in a conventional system like language, as this implies changing the meaning of the strings. Languages can only increase their optimality gradually.
Two speakers can change locally their conventions to increase their optimality but the challenge is to propagate those changes over the entire population. 
% What seems to allow for this phenomenon to happen is locality. In fact, while it is difficult to apply these changes on a global scale - given the conventionality of language and the heterogeneity of its scope - optimization can and actually is performed on a local basis, in which a new convention can be spread and adopted easily.
Among the gradual changes, some look easier than others for a cognitive system. 
% Another related question is, once a sub-optimal system has emerged, why it is difficult to turn it optimal? 
Optimizing a system implies changing the lengths of the strings that are being assigned to each type. The lower the resemblance between the new string and the old string, the higher the cognitive effort. Thus language users will have difficulties to interpret a string with its new meaning or to change their mental mapping of meanings to strings. We actually see every-day examples of word length reduction of the cognitively easy kind: acronyms and abbreviations for frequently used expressions are used to communicate more efficiently (e.g. 'org.' to say 'organization', 'asap' to say 'as soon as possible'). That is the case of certain areas of specialization (e.g. 'CPU' to say 'central processing unit' in Computer Science, 'ACL' to say 'anterior cruciate ligament' in Anatomy), communication on social media (e.g. 'idk' to say 'I don't know', 'brb' to say 'be right back'), or when taking personal notes.  

However, there is also preliminary evidence for the opposite trend: word lengths seem to have been increasing over historical time in Chinese and Arabic \parencite{Chen2015a,Milicka2018a}. This is relevant to our study for three reasons. First, it indicates a potential diachronic pattern of word lengths at odds with the law of abbreviation, which is a synchronic observation. Second, it suggests that the contribution to the cost of communication stemming from word lengths has been increasing over time in these languages. Third, it highlights the necessity for scores which are comparable over time. In fact, we have defined above word length scores which are less affected by the scarcity of texts samples. These are hence amenable to diachronic studies where textual material from earlier time periods is often sparse. 

In sum, all the arguments above raise the question if full optimality could be reached just by gradual, local, distributed and cognitively easy changes. We thus have to ponder if reaching full optimality is actually necessary for a language. The actual degree of optimality of word lengths in languages (more than $50\%$ on scale from 0 to $100\%$ according to $\Psi$) may just be ``good enough'' in language production \parencite{Goldberg2022a}. 

% What seems to allow for this phenomenon to happen is locality. In fact, while it is difficult to apply these changes on a global scale - given the conventionality of language and the heterogeneity of its scope - optimization can and actually is performed on a local basis, in which a new convention can be spread and adopted easily.

% Nevertheless, merging the new shorter strings to effectively optimize the language globally is not only difficult due to collisions ('ACL' would be interpreted as 'access control list' by a computer security expert), but it would also likely not lead to greater global optimization, as every setting requires different terminology, which implies that what is frequently used in a field is not in another. [...]

\subsection*{Future research}

In this article, we have introduced two new scores that require further theoretical and empirical research. Although we have focused on $\Psi$ for the sake of simplicity, further attention to $\Omega$ and its variants is necessary. 
% Chris, new text follows
We have made a contribution to the problem of the best score for cross-linguistic research, but the problem should be the subject of further research.
% The questions of the best score for cross-linguistic research is an open problem. 

For simplicity, we have investigated only the weak recoding problem. The next step is initiating research on the strong recoding problem, which may provide further arguments for the choice of the best score.  

We have approached the degree of optimality of word lengths only synchronically. However, we have established some foundations for research on the evolution of communication. In particular, we hope that $\Psi$ contributes to revise the finding that word lengths have been increasing over centuries in Chinese and Arabic based on $L$, namely simple average word lengths \parencite{Chen2015a,Milicka2018a}, and extending the analysis to more languages. At present, these findings suggest that languages may have been evolving against the principle of compression. 

Finally, our investigation of the degree of optimality of word durations has paved the way for exploring the degree of optimality of the durations of vocalizations or gestures of other species \parencite{Semple2021a}.

\section*{Data accessibility}
Data and code for this research work are stored in GitHub:\\ \url{\repository}.\\ 
For a permanent version of the software used in the paper see:\\ 
\url{https://doi.org/10.5281/zenodo.18890298}.

\section*{Authors' contributions}

CRediT (Contributor Roles Taxonomy) contributions for each author are as follows.

SP: Conceptualization, Formal Analysis, Investigation, Software, Supervision, Validation, Visualization, Writing-original draft, Writing-review \& editing;
ACM: Data curation, Resources, Software;
JCM: Writing-review \& editing;
MW: Writing-review \& editing, Software, Visualization;
CB: Conceptualization, Writing-review \& editing;
RFC: Conceptualization, Formal analysis, Funding acquisition, Methodology, Project Administration, Supervision, Writing-original draft, Writing-review \& editing.

% \section*{Conflict of interest declaration}

% We declare we have no competing interests.

\section*{Acknowledgements}

This article is one of the products of the research project of the 1st edition of the master degree course ``Introduction to Quantitative Linguistics'' at Universitat Politècnica de Catalunya. We are specially grateful to two students of that course: L. Alemany-Puig for helpful discussions and computational support, and M. Michaux for comments on early versions of the manuscript. We also thank J. Moreno-Fernández for contributing to early developments of this research program \parencite{Moreno2021a}. We thank M. Farrús and A. Hernández-Fernández for advice on voice datasets. We also thank S. Komori for advice on Japanese, Y. M. Oh for advice on Korean and S. Semple for helping us to improve English.
Finally, we thank an anonymous reviewer, A. Goldberg and L. Debowski for valuable feedback. 
% SP is funded by the grant 'Thesis abroad 2021/2022' from the University of Milan. 
RFC was supported by a recognition 2021SGR-Cat (01266 LQMC) from AGAUR (Generalitat de Catalunya). 

RFC and CB were funded by the grant PID2024-155946NB-I00 funded by Ministerio de Ciencia, Innovación y Universidades (MICIU), Agencia Estatal de Investigación (AEI/10.13039/501100011033) and the European Social Fund Plus (ESF+).

CB was funded by the \textit{Deutsche Forschungsgemeinschaft} (FOR 2237:  Words, Bones, Genes, Tools - Tracking Linguistic, Cultural and Biological Trajectories of the Human Past), the \textit{Schweizerischer Nationalfonds zur Förderung der Wissenschaftlichen Forschung} (Non-randomness in Morphological Diversity: A Computational Approach Based on Multilingual Corpora, 176305), and the European Union (ERC, EVINE, 101117111). Views and opinions expressed are however those of the author(s) only and do not necessarily reflect those of the European Union or the European Research Council Executive Agency. Neither the European Union nor the granting authority can be held responsible for them.  
The sponsors had no role in study design; collection, analysis, and interpretation of data; writing of the paper; and decision to submit it for publication.

\section*{Abbreviations}

\noindent {\bf CV} Common Voice Forced Alignements

\noindent {\bf NS} Non-singular coding

\noindent {\bf PUD} Parallel Universal Dependencies

\noindent {\bf RO} Rank Ordering

\noindent {\bf UD} Uniquely decodability

\printbibliography

%%%%%%%%%%% Appendix

%==================================================
% Appendix

\appendix % this is not from the original template of the journal

\begin{appendices} % this is not from the original template of the journal

% https://tex.stackexchange.com/questions/559218/use-appendix-letter-in-figure-and-table-captions
\counterwithin{figure}{section}
\counterwithin{table}{section}
\renewcommand\thefigure{\thesection\arabic{figure}}
\renewcommand\thetable{\thesection\arabic{table}}

%%%%%%%%%%% Theory

\section{The optimality scores}
\label{app:optimalityScores}

\subsection{The design of two new optimality scores}

By adapting the new score for the optimality of dependency distances \parencite{Ferrer2020b} to word lengths, we obtain \autoref{eq:Psi}.
Importantly, it can be shown that $\Psi$ is proportional to the Pearson correlation $r$ between frequency and length (\autoref{app:Pearson_correlation}), i.e.
$$\Psi = -a r,$$
where
$$a = \frac{(n-1)s_p s_l}{L_r - L_{min}}$$
is a constant determined by the distribution of the $l_i$'s and the distribution of the $p_i$'s; $s_p$ and $s_l$ are the standard deviations of word probability and word length, respectively.
Traditionally, $r$ is seen as a measure of linear association between two variables (\textcite{vandenJeuvel2022a} and references therein). Therefore, a potential limitation of $\Psi$ is that it may be unable to capture the actual non-linear association between frequency and length \parencite{Sigurd2004a,Strauss2007a}. % add more references ???

This takes us to a second attempt to define an optimality score for word lengths, that is based on Kendall $\tau$ correlation.
Applying the same template that we used for the design of $\Psi$, we get the score
\begin{equation}
\Omega = \frac{\tau_r - \tau}{\tau_r - \tau_{min}},
\label{eq:Omega_raw}
\end{equation}
where $\tau$ is the actual Kendall $\tau$ correlation between frequency and length, $\tau_r$ is the expected value of $\tau$ under our null hypothesis \parencite{Petrini2022b} and $\tau_{min}$ is 
% the minimum value of $\tau$ that can be achieved by one of these shufflings.
defined by rank ordering as for $L_{min}$ (\autoref{sec:coding_schemes}).
Since $\tau_r = \expe{\tau} = 0$ \parencite{vandenJeuvel2022a}, % look for a better reference ???
$\Omega$ becomes simply
$$\Omega = \frac{\tau}{\tau_{min}}.$$

Since $L=L_{min}$ implies $n_c = 0$ \parencite{Ferrer2019c}, the definition of Kendall $\tau$ (\autoref{eq:Kendall_tau_correlation}) yields
\begin{eqnarray*} 
% \Omega & = & \frac{\frac{n_c - n_d}{{n \choose 2}}}{\frac{- n_{d,min}}{{n \choose 2}}}.\\
%       & = & \frac{n_d - n_c}{n_{d,min}},
\Omega & = & \frac{c(n_c - n_d)}{c (0 - n_{d,min})}\\
      & = & \frac{n_d - n_c}{n_{d,min}},
\end{eqnarray*}
where $n_{d,min}$ is the number of discordant pairs when $L = L_{min}$.

Having said that, notice that one cannot expect $\Psi$ to be unable to capture non-linear associations. It has been shown that there are non-linear associations that are better captured by Pearson correlation, against common belief \parencite{vandenJeuvel2022a}. Therefore, $\Psi$ and $\Omega$ should be seen, for the time being, as distinct approaches to deal with monotonic associations. 

\subsection{The statistical significance of an optimality score}

According to the mathematical analysis in \autoref{app:Pearson_correlation}, testing if
$\eta$ or $\Psi$ are significantly large (or $L$ is significantly small), as expected by the principle of compression, is equivalent to testing if Pearson $r$ is significantly small by means of a left-sided correlation test. In contrast,
testing if $\Omega$ is significantly large is equivalent to testing if $\tau$ is significantly small. 
Therefore, testing if $\Omega$ is significantly large is also equivalent to testing the law of abbreviation by means of a left-sided Kendall $\tau$ correlation test. 
Similarly, testing if $\Psi$ (or $\eta$) is significantly large is also equivalent to testing the law of abbreviation by means of a left-sided Pearson $r$ correlation test. In light of these equivalences, previous research on the law of abbreviation by means of Pearson $r$ or Kendall $\tau$ \parencite{Petrini2022b,Bentz2016a,Semple2010a,Ferrer2009g}
has implicitly tested if $\eta$, $\Psi$ or $\Omega$ are significantly large.

% DO WE NEED THAT?
% To show that word lengths optimized significantly (namely, $L$ is significantly small or $\eta$ or $\Psi$ are significantly large), we used a left-sided Pearson correlation test.

\subsection{Overview of the theoretical properties of the scores}

Here we examine the theoretical properties of distinct scores, including correlation coefficients between word frequency and word length (Pearson $r$ and Kendall $\tau$). We begin with a quick summary. Firstly, $L$ is the only score that is not normalized. However, while $\eta$ is singly normalized (it is normalized only with respect to the minimum baseline, $L_{min}$), $\Psi$ and $\Omega$ are dually normalized because they are normalized with respect to the minimum baseline ($L_{min}$ or $\tau_{min}$) and the random baseline ($L_r$ or $\tau_r=0$). $\eta$, $\Psi$ and $\Omega$ exhibit constancy under optimal coding (namely $\Omega = \Psi = \eta = 1$ when $L = L_{min}$). 
$\Psi$ and $\Omega$, as the correlation scores, exhibit stability under a random mapping of probabilities into length -- namely their expected value in that random mapping is zero -- while that of $\eta$ is moving (depending on certain parameters of the communication system). Hence $\eta$ lacks this property.
Constancy under optimality and stability under the null hypothesis together make a critical difference. The scores that lack one of them or both, i.e. $L$, $\eta$ and the correlation scores ($r$ and $\tau$) are less informative about the actual distance to optimality than $\Psi$ and $\Omega$ when comparing distinct systems, e.g. distinct languages \parencite{Ferrer2018b} or the same language at different evolutionary stages \parencite{Chen2015a}. 
\autoref{tab:mathematical_properties_of_scores} summarizes the desirable mathematical properties of these scores. 

After this quick overview, we refer the reader to \autoref{app:constany_under_optimal_coding_theory} for further details on constancy under optimal coding and expand the other mathematical properties giving further details (in parenthesis, we indicate the sections of \autoref{app:theory} where full detail is given)
% From a mathematical standpoint, we evaluate the scores considering the following desirable properties: 
\begin{itemize}
    \item 
    {\em Stability under the null hypothesis} (\autoref{app:stability_under_null_hypothesis_theory}). A score exhibits stability under the null hypothesis if its expectation takes a constant value under the null hypothesis \parencite{Ferrer2020b}. The optimality scores $\Psi$ and $\Omega$ as well as the correlation coefficients $r$ and $\tau$, are stable under the null hypothesis, namely their expectation is zero when the mapping of word probabilities into word lengths is random \parencite{Petrini2022b}. In contrast, $\eta$ lacks that property.  
    \item
    {\em Invariance under linear transformation} (\autoref{app:invariance_under_linear_transformation_theory}).
    A score exhibits invariance under linear transformation when its value does not change if a linear transformation is applied to the target variable \parencite{Ferrer2020b}. The Pearson correlation between two variables  is invariant under linear transformation only if (a) the linear transformation that is applied to only one of the variables is increasing or (b) the transformations that are applied to each of the variables are both increasing or both decreasing (\autoref{app:Pearson_correlation_linear_transformation}); the same applies to Kendall $\tau$ correlation. For simplicity, here we focus on increasing linear transformations of word length. \\
$\Psi$ and $\Omega$ are invariant (do not change) if length is transformed linearly, namely, each length $l$ is replaced by a linear function of it, i.e. $g(l) = al + b$, where $a$ and $b$ are some constants, with $a>0$. In contrast, $\eta$, is not invariant under a linear transformation but invariant under a proportional change of scale, which is obtained when $b=0$.
    \item
    {\em Invariance under monotonic transformation} (\autoref{app:invariance_under_non_linear_monotonic_transformation_theory}).
A score exhibits invariance under strictly increasing transformation if its value does not change if a transformation of this sort is applied to the target variable.
The Kendall $\tau$ correlation between two variables is invariant under strictly non-linear monotonic transformation only if (a) the transformation that is applied to only one of the variables is increasing or (b) the transformations that are applied to each of the variables are both increasing or both decreasing (\autoref{app:Kendall_correlation_strictly_monotonic_transformation}). In contrast, that invariance is only warranted for Pearson correlation when the transformations are linear. \\
For simplicity, here we focus on increasing non-linear transformations of word length.  
$\Omega$ is invariant (does not change) if each length $l$ is replaced by a strictly increasing function of it, $h(l)$. Notice that $h$ can be non-linear. 
In contrast, $\Psi$ and $\eta$ are not invariant under that transformation.
\end{itemize}

\begin{table}
\caption{\label{tab:mathematical_properties_of_scores} Summary of the mathematical properties of each of the word length scores: mean word length ($L$), the Pearson and the Kendall correlation between frequency and length ($r$ and $\tau$, respectively) and the optimality scores $\eta$, $\Psi$ and $\Omega$. Concerning invariance, we assume for simplicity that the transformation is applied to word length only.  }
\begin{tabular}{lrrrrrr}
Properties & $L$ & $\eta$ & $r$ & $\Psi$ & $\tau$ & $\Omega$ \\
\hline
Normalization & none & single & single & dual & single & dual \\
Constancy under optimal coding            & & $\checkmark$ &              & $\checkmark$ & & $\checkmark$ \\
Stability under the null hypothesis             & &              & $\checkmark$ & $\checkmark$ & $\checkmark$ & $\checkmark$ \\
Invariance under translation                  & &              & $\checkmark$ & $\checkmark$ & $\checkmark$ & $\checkmark$ \\
Invariance under proportional change of scale & & $\checkmark$ & $\checkmark$ & $\checkmark$ & $\checkmark$ & $\checkmark$ \\
Invariance under increasing linear transformation        & &              & $\checkmark$ & $\checkmark$ & $\checkmark$ & $\checkmark$ \\
Invariance under strictly increasing non-linear transformation        & &              &     & & $\checkmark$ & $\checkmark$ \\
\end{tabular}
\end{table}

\section{The minimum baselines}
\label{app:baselines}
To calculate $\eta$, $L_{min}$ has been computed making two kinds of assumptions borrowed from standard information theory and its extensions \parencite{Ferrer2018b,Ferrer2019c}:
\begin{itemize}
    \item
    Non-singular coding (NS), namely two word types should not be assigned the same string. We use $L_{min}^{NS}$ to refer to the specific value of $L_{min}$ in this setting.  
    \item 
    Unique decodability (UD), namely the assigned strings, in addition to being non-singular, should have a unique segmentation when concatenated forming a sequence of strings without blanks or other sorts of separators. 
    We use $L_{min}^{UD}$ to refer to the specific value of $L_{min}$ in this setting.
\end{itemize}
Each of these assumptions is known as coding scheme in the language of information theory \parencite{Cover2006a}. See \textcite{Ferrer2019c} for further details on these two ways of defining the minimum baseline.

Previous research on the optimality of word lengths has used $L_{min}^{NS}$ and $L_{min}^{UD}$ \parencite{Ferrer2018b, Moreno2021a}. 
Here we focus on a third way of computing $L_{min}$, that we call rank ordering (RO) \parencite{Moreno2021a} and that is called ``Zipfian coding'' in related work \parencite{Pimentel2021a}. We use $L_{min}^{RO}$ to refer to this specific value of $L_{min}$.  
This method follows from the generalized theory of compression \parencite{Ferrer2019c} and has already been used in previous research on the optimality of word lengths \parencite{Moreno2021a,Pimentel2021a}. 

$L_{min}^{RO}$ is obtained when the current values of $l_i$ are reassigned to frequencies (or equivalently probabilities) so as to minimize $L$. 
$L$ is minimized when probabilities are sorted decreasingly and lengths are sorted increasingly \parencite{Ferrer2019c}. Then the $i$-th most frequent type gets the $i$-th shortest length, hence the name {\em rank ordering}. 

\begin{table}[ht]
  \caption{\label{tab:example} Matrix indicating the frequency and length of three types. The mean token length is $L = \frac{235}{125}=1.88$ and the mean type length is $L_r = \frac{6}{3} = 2$. } 
  
  \begin{center} 
  \begin{tabular}{lll}
  $i$ & $f_i$ & $l_i$ \\ 
  \hline 
  1   &  100  & 2  \\         
  2   &  20   & 1  \\
  3   &  5    & 3 \\
  \end{tabular}
  \end{center}
\end{table}

 To see this, consider the matrix with two columns, $f_i$ and $l_i$, that are used to compute $L$ (\autoref{tab:example}).  
 Suppose that we shuffle the column of $l_i$ in \autoref{tab:example}. We use $l_i'$ to refer to the new value of $l_i$ after the shuffling and $L'$ to refer to the new value of $L$, that is \begin{equation*}
 L' = \frac{1}{T} \sum_{i = 1}^n f_i l_i'.
 \end{equation*}
\autoref{tab:all_mappings} shows all the possible shufflings (permutations) of the $l_i$ column that can be produced and the corresponding values of $L'$ for the matrix in \autoref{tab:example}.
Then $L_r$ is the average value of $L'$ over all permutations of column $l_i$ and
 $L_{min}^{RO}$ is the value of $L'$ of the  permutation  that minimizes the value of $L'$. According to \autoref{tab:all_mappings}, $L_{min}^{RO} = \frac{155}{125}=1.24$ for the matrix in \autoref{tab:example}. 
 By symmetry, $L_{min}^{RO}$ is also the value of $L'$ of the permutation of column $f_i$ that minimizes the value of $L'$, where now
 \begin{equation*}
 L' = \frac{1}{T} \sum_{i = 1}^n f_i' l_i
 \end{equation*}
 and $f_i'$ is the new value of $f_i$ after the shuffling.
 
\begin{table}
  \caption{\label{tab:all_mappings} All the $3! = 6$ permutations of the column $l_i$ in \protect \autoref{tab:example} that can be produced. Each permutation is indicated with letters from A to F. A is the one minimizing $L$.
  $L'$, the mean length of tokens in a given permutation, is shown at the bottom for each permutation. } 
  
  \begin{center} 
  \begin{tabular}{llllllll}
      &       & A     & B     & C     & D     & E     & F    \\
  $i$ & $f_i$ & $l_i'$ & $l_i'$ & $l_i'$ & $l_i'$ & $l_i'$ & $l_i'$\\
  \hline 
  1   &  100  & 1     & 1     & 2     & 2     & 3     & 3    \\         
  2   &  20   & 2     & 3     & 1     & 3     & 1     & 2    \\
  3   &  5    & 3     & 2     & 3     & 1     & 2     & 1    \\
  \hline  
  \\
      &  $L'$  & $\frac{155}{125}=1.24$   & $\frac{170}{125}=1.36$   & $\frac{235}{125}=1.88$   & $\frac{265}{125}=2.12$   & $\frac{330}{125}=2.64$   & $\frac{345}{125}=2.76$  \\ 
  \end{tabular}
  \end{center}
\end{table}

The rationale of rank ordering is that it is sensitive to imperceptibility and distinguishability factors as well as to the phonotactic constraints shaping the length of words in a language \parencite[Section 2.2]{Ferrer2019c}. In contrast, optimal non-singular coding and optimal uniquely decodable encoding completely neglect constraints that are language specific or specific to humans (e.g. their cognition, their anatomy, etc.), and abstract away from the cultural and biological evolution processes that lead to more efficient languages \parencite{Kanwal2017a}. Rank ordering is conservative  with respect to optimality because it assumes that a language cannot produce word lengths that are better than the ones that are being observed. In contrast, non-singular coding and uniquely decodable coding point towards the theoretical limits of optimal coding in languages.

The Kendall $\tau$ correlation can be defined as 
% \parencite{Conover1999a}
% \parencite{Kendall1938a,Kendall1945a}
\parencite{Kendall1970a}
\begin{equation}
% \tau = \frac{n_c - n_d}{{n \choose 2}},
\tau = c(n_c - n_d),
\label{eq:Kendall_tau_correlation}
\end{equation}
where $n_c$ is the number of concordant pairs, $n_d$ is the number of discordant pairs and $c$ is a normalization factor.
The previous definition of $ \tau$ (\autoref{eq:Kendall_tau_correlation}) is the template for the two correlation scores that were proposed originally by Kendall, now known as $\tau$-a and $\tau$-b 
% \parencite{Kendall1938a,Kendall1945a}. 
\parencite{Kendall1970a}. 
$\tau$-a is the simplest and is obtained when 
\begin{equation*}
c = \frac{1}{{\binom{n}{2}}}.
\end{equation*}
$\tau$-b is obtained by another $c$ the corrects for ties. It is worth mentioning that $\tau$-b is the one it implemented in the base R statistical programming language.

$\tau_{min}^{RO}$ is the smallest value that $\tau$ can achieve by shuffling the column of the $p_i$'s or the column of the $l_i$'s in the matrix. $\tau_{min}^{RO}$ coincides with the ordering of the $l_i$'s in the matrix such that $L$ is minimized, namely, $L = L_{min}^{RO}$ \parencite{Ferrer2019c}. 
$\tau_{min}^{RO} = -1$ if there are no ties both in the $p_i$'s and the $l_i$'s; otherwise (the common situation), $ \tau_{min}^{RO}> -1 $ (\autoref{app:theory}, Property \ref{prop:bounds}).  

% $L_{min}^{RO}$ and $\tau_{min}^{RO}$ are
$\tau_{min}^{RO}$ is unaffected if the lengths are replaced by new ``lengths'' that result from applying a monotonically increasing function of $l$ as the following property states. That property will be used to analyze the mathematical properties of distinct scores in 
% \autoref{sec:theory}.
\autoref{app:theory}.

\begin{property}
\label{minimum_baseline}

Let $L_{min}^{RO}$ be the minimum value of $L$ under rank ordering, defined as
\begin{equation*}
L_{min}^{RO} = \sum_{i=1}^n p_i l_{i,min}. 
\end{equation*}
Let $h(l)$ be a strictly monotonically increasing function of $l$. 
We use a prime mark, $'$, to indicate the new value of a function after applying $h(l)$ to the $l_i$'s. Accordingly, 
\begin{equation*}
L' = \sum_{i=1}^n p_i h(l_i)     
\end{equation*}
and 
$$\tau' = \tau(p,h(l)).$$
Then the minimum value that $L'$ can achieve is 
\begin{equation}
{L'}_{min}^{RO} = \sum_{i=1}^n p_i h(l_{i,min})    
\label{eq:new_minimum_length}
\end{equation}
while the minimum of $\tau$ remains unaltered, i.e.  
$${\tau'}_{min}^{RO} = \tau_{min}^{RO}.$$

\end{property}
\begin{proof}
Recall that $L_{min}^{RO}$ is computed by sorting the lengths increasingly and the probabilities decreasingly \parencite{Ferrer2019c} in the matrix that is used to calculate $L$. Applying the same method to calculate ${L'}_{min}^{RO}$, it turns out that, if the strictly monotonic transformation is increasing, the sorting by the new lengths that are obtained after replacing each original length $l$ by $h(l)$ will preserve the position of the lengths in the original sorting, giving \autoref{eq:new_minimum_length}.
For the same reason, $\tau_{min}' = \tau_{min}$, because the ordering of the $l_i$'s that minimizes $L$ (yielding $L_{min}^{RO}$) coincides with the ordering of the $h(l_i)$'s that minimizes ${L'}_{min}^{RO}$ since $h(l)$ is strictly increasing.
\end{proof}

In this article, we will investigate the degree of optimality of languages  focusing on rank ordering as a lower bound of the true degree of optimality of word lengths in languages. The choice of rank ordering is for the sake of simplicity and to ensure that the optimality scores are properly normalized as explained in \autoref{app:theory}. 
% Given that focus, hereafter, we use $L_{min}$ and $\tau_{min}$ to refer to $L_{min}^{RO}$ and $\tau_{min}^{RO}$, respectively.

\section{Theory}
\label{app:theory}

\subsection{Technical remarks on normalization and coherence}

\label{app:technical_remmarks_on_normalization_and_coherence}

$\eta$ was designed to be normalized, namely, $0 \leq \eta \leq 1$ \parencite{Ferrer2019c}. However, 
a theoretical challenge in the calculation of $\eta$ with the two versions of $L_{min}$ from information theory (optimal non-singular coding and optimal uniquely decodable encoding in \autoref{sec:coding_schemes} and \autoref{app:baselines}),  namely $L_{min}^{NS}$ and $L_{min}^{NS}$, 
is to ensure that the assumptions of each coding scheme hold. If a language satisfies them, then
$\eta \leq 1$ because $L \leq L_{min}$. If not, then $\eta > 1$ may be possible. The same problem may concern $\Psi$ or $\Omega$ depending on the choice of the minimum baseline. This is another reason why we choose a minimum baseline based on rank ordering, i.e. $L_{min}^{RO}$,  so that the random baseline and the minimum baseline are perfectly aligned: $L_{min}^{RO}$ is the minimum average token length over all permutations in \autoref{tab:all_mappings}; $L_r$ is the average token length over all these permutations \parencite{Petrini2022b}. Thus, our choice of the baselines warrants that $\eta, \Psi, \Omega \leq 1$ for any communication system. Hereafter, we use $L_{min}$ and $\tau_{min}$ to refer to $L_{min}^{RO}$ and $\tau_{min}^{RO}$, respectively.

\subsection{Constancy under optimal coding}

\label{app:constany_under_optimal_coding_theory}

A score exhibits constancy under optimality if it takes a constant value in case the system under consideration is organized optimally \parencite{Ferrer2020b}.  
$\eta$, $\Psi$ and $\Omega$ exhibit constancy under optimality, namely they take a value of 1 when word lengths are minimum and actually their value never exceeds one. In case of optimal coding, $\tau \leq 0$ \parencite{Ferrer2019c}. Although $r$ and $\tau$ are bounded below by -1, they do not necessary reach -1 when lengths are minimum \parencite{Ferrer2019c}. Obviously, $L$ lacks any of these properties since $L \geq L_{min}$ and $L_{min}$ does not need to be one. In the language of mathematics,
\begin{property}
\label{prop:bounds}
\noindent
\begin{enumerate}
    \item 
    $\Psi, \Omega, \eta \leq 1$, with equality if and only if $L_{min} = L$.
    \item
    $-1 \leq r, \tau \leq 1$, but $L_{min} = L$ does not imply $r = - 1$ or $\tau = - 1$.
\end{enumerate}
 
\end{property}
\begin{proof}
\noindent
\begin{enumerate}
    \item 
Since $L_{min} \leq L$ by definition, 
\begin{eqnarray*}
\Omega = \frac{\tau}{\tau_{min}} \leq 1\\
\eta = \frac{L_{min}}{L} \leq 1.
\end{eqnarray*}
For the same reason, 
$$ - L \leq - L_{min}$$
and adding $L_r$ to both sides of the equation, one obtains
$$L_r - L \leq L_r - L_{min}.$$
Dividing by $L_r - L_{min}$ on both sides of the inequality, one gets 
$$\frac{L_r - L}{L_r - L_{min}} \leq \frac{L_r - L_{min}}{L_r - L_{min}}$$ because $L_r \geq  L_{min}$ and hence
$\Psi \leq 1$. 

Finally, notice that $\tau_{min} \leq \tau$ by definition and $\tau_{min}$ is negative \parencite{Ferrer2019c}. Then, dividing by $\tau_{min}$ on both sides of $\tau \geq \tau_{min}$, we obtain
\begin{eqnarray*}
\Omega = \frac{\tau}{\tau_{min}} \leq \frac{\tau_{min}}{\tau_{min}} = 1. 
\end{eqnarray*}

\item
It is well-known that $-1 \leq r, \tau \leq 1$ \parencite[Section 5.4]{Conover1999a}. When $L_{min} = L$, $r = - 1$ is only possible if the relationship between $p$ and $l$ is perfectly linear in the optimal shuffling of value $l_i$'s in the matrix; $\tau = - 1$ can only be achieved if there are not ties neither within the $p_i$'s nor withing the $l_i$'s \parencite{Ferrer2019c}. Notice that, Kendall proposed two rank correlation scores, i.e. $\tau$-a and $\tau$-b 
% \parencite{Kendall1938a,Kendall1945a}. 
\parencite{Kendall1970a}. 
$\tau$-b, which is the one that base R implements, incorporates a correction for ties but such correction does not ensure that $\tau$-b yields a constant value ($\tau\mbox{-b} = -1$) when there are no concordant pairs. % See test.R 
\end{enumerate}
\end{proof}

\subsection{Stability under the null hypothesis}

\label{app:stability_under_null_hypothesis_theory}

% \textcolor{red}{$\Psi$, $\Omega$, $r$ and $\tau$ are stable under the null hypothesis but $\eta$ is not.} In the language of mathematics,

\begin{property}
\label{stability_property}
$$\expect[\Psi] = \expect[\Omega] = \expect[r] = \expect[\tau] = 0$$ while 
\begin{equation}
\expect[\eta] \geq \frac{L_{min}}{L_r}.
\label{eq:expectation_eta}
\end{equation}
\end{property}

\begin{proof}
It is well-known that $\expect[r] = \expect[\tau] = 0$ (see \textcite{vandenJeuvel2022a} and references therein). 
On the one hand, 
   \begin{eqnarray*}
   \expect[\Omega] & = & \expect\left[\frac{\tau}{\tau_{min}}\right] \\
                   & = & \frac{1}{\tau_{min}} \expect[\tau] \\
                   & = & 0                     
   \end{eqnarray*} 
and    
   \begin{eqnarray*}
   \expect[\Psi] & = & \expect\left[\frac{L_r - L}{L_r - L_{min}}\right] \\
                   & = & \frac{1}{L_r - L_{min}} \left(L_r - \expect[L]\right) \\
                   & = & \frac{1}{L_r - L_{min}} \left(L_r - L_r\right) \\
                   & = & 0.                      
   \end{eqnarray*} 
On the other hand,
\begin{eqnarray*}
\expect[\eta] & = & \expect\left[\frac{L_{min}}{L}\right] \\
              & = & L_{min}\expect\left[\frac{1}{L}\right]. \\
\end{eqnarray*}
$\frac{1}{L}$ is a convex function of $L$ because $L$ is positive by definition.  
Jensen's inequality gives $$\expect\left[\frac{1}{L}\right] \geq \frac{1}{\expect[L]}.$$ 
Finally
\begin{eqnarray*}
\expect[\eta] & \geq & \frac{L_{min}}{L_r}. 
\end{eqnarray*}
\end{proof}
Obviously, $L$ lacks stability under the null hypothesis since $\expect[L] = L_r$ and that is the mean length of types.  

Property \ref{stability_property} only indicates a lower bound for $\expect[\eta]$. Then the true $\expect[\eta]$ might be stable under the null hypothesis in the sense that $\expect[\eta] = k$, where $k$ is some constant. The fact that $L_{min}, L_r > 0$ and  \autoref{eq:expectation_eta} imply $k > 0$. In contrast, $\expect[\Psi] = \expect[\Omega] = k$ with $k = 0$.
However, in 
% \autoref{app:analysis},  
\autoref{app:stability_under_null_hypothesis_test}, 
we demonstrate that $\eta$ is not stable under the null hypothesis with real data, indeed $\expect[\eta] \approx \frac{L_{min}}{L_r}$, and confirm that $\Psi$ and $\Omega$ are  stable under the null hypothesis while $L$ is not. 
% , and investigate a stronger stability condition. 

% \subsection{The relationship between $\Psi$ and Pearson correlation}
\subsection{The relationship between \texorpdfstring{$L$}{L}, \texorpdfstring{$\eta$}{eta}, \texorpdfstring{$\Psi$}{Psi} and Pearson correlation}

\label{app:Pearson_correlation}

First, we show the relationship between $\Psi$ and covariance or Pearson correlation through their estimators.

Given a sample of $n$ points, $\left\{(x_1, y_1),...,(x_i, y_i),...,(x_n,y_n) \right\}$,
the sample covariance is defined as 
$$s_{xy} = \frac{1}{n-1}\left(\sum_{i=1}^n x_i y_i - n \bar{x}\bar{y}\right),$$
where $\bar{x}$ is the sample mean of $x$ and $\bar{y}$ is the sample mean for $y$, i.e. 
\begin{eqnarray*}
\bar{x}= \frac{1}{n} \sum_{i=1}^n x_i \\
\bar{y}= \frac{1}{n} \sum_{i=1}^n y_i.
\end{eqnarray*}

Now we replace the random variables $x$ and $y$ by $p$ (the probability of a type) and $l$ (the length/duration of a type). Accordingly, 
the sample of $n$ points becomes $\left\{(p_1, l_1),...,(p_i, l_i),...,(p_n,l_n)\right\}$, one point per type. 
Then the covariance between $p$ and $l$ is 
$$ s_{pl} = \frac{1}{n-1}\left(\sum_{i=1}^n p_i l_i - n \bar{p}\bar{l}\right).$$
Recalling the definition of $L$ (\autoref{eq:mean_type_length}) and noting that $\bar{p} = \frac{1}{n}$ and $\bar{l} = L_r$ (\autoref{eq:random_baseline}), we finally obtain
\begin{equation}
s_{pl} = \frac{1}{n-1}(L - L_r).
\label{eq:covariance}
\end{equation}
Then $\Psi$ turns out to be proportional to the sample covariance, i.e. 
\begin{equation}  
\Psi = \frac{L_r - L}{L_r - L_{min}} = - \frac{n-1}{L_r - L_{min}} s_{pl}
\label{eq:Psi_and_covariance}
\end{equation}
but with an opposite sign.

The sample Pearson correlation is  
\begin{equation}
r = \frac{s_{xy}}{s_x s_y},
\label{eq:Pearson_correlation}
\end{equation}
where $s_x$ and $s_y$ are the sample standard deviation of $x$ and $y$, i.e.
\begin{eqnarray*}
s_x = \sqrt{\frac{1}{n-1} \sum_{i=1}^n (x_i - \bar{x})^2} \\
s_y = \sqrt{\frac{1}{n-1} \sum_{i=1}^n (y_i - \bar{y})^2}.
\end{eqnarray*}
Combining \autoref{eq:covariance} and \autoref{eq:Pearson_correlation}, we find that 
$$r = \frac{L - L_r}{(n-1)s_p s_l}.$$
Combining \autoref{eq:Psi_and_covariance} and \autoref{eq:Pearson_correlation}, we find
that $\Psi$ is negatively proportional to the sample Pearson correlation, i.e.
$$\Psi = -\frac{(n-1)s_p s_l}{L_r - L_{min}} r.$$

Similarly, it is easy to see that $\eta$ and $L$ are linear functions of $r$ or $s_{pl}$. For instance, 
$$\eta =\frac{1}{L_{min}} \left[(n-1)s_p s_l r + L_r \right].$$ 
Other linear relationships can be derived similarly.

\subsection{Invariance of Pearson correlation under linear transformation}

\label{app:Pearson_correlation_linear_transformation}

% https://stats.stackexchange.com/questions/177940/invariance-of-correlation-to-linear-transformation-textcorraxb-cyd
% accurate (but too strict) argument on invariance under linear transformation for Pearson correlation     https://math.stackexchange.com/questions/4399610/pearson-correlation-as-a-measure-for-non-linear-dependence

The Pearson correlation between two random variables $X$ and $Y$ is the covariance normalized by the product of the standard deviations, i.e.
$$r(X,Y) = \frac{COV(X,Y)}{\sigma(X) \sigma(Y)}.$$
It is often stated that Pearson correlation $r(X,Y)$ between two random variables $X$ and $Y$ (or its estimator) is invariant under linear transformation, i.e. $r(X, Y)$ does not change if is $X$ is replaced by $f(X) = aX + b$ and $y$ is replaced by $g(Y) = cY + d$
% \parencite{Wikipedia, other sources}, 
\parencite[Question 3.11]{Gujarati2009a}.
However, the statement is not accurate enough as the following property clarifies.

\begin{property}
$r(X,Y)$ is invariant under linear transformation. 
\begin{itemize}
\item
In a strict or strong sense, i.e. 
$$r(aX + b, cY + d) = r(X,Y),$$ 
if and only if $\sgn(a) = \sgn(c)$
and 
$$r(aX + b, cY + d) = -r(X,Y),$$ 
if and only if $\sgn(a) \neq \sgn(c)$
\item 
In a weak sense, i.e. 
$$|r(aX + b, cY + d)| = |r(X,Y)|,$$ 
in any circumstance (namely independently of $a$, $b$, $c$ and $d$). 
\end{itemize}
\end{property}
\begin{proof}
Knowing that \parencite[Sections 4.3 and 4.6]{DeGroot2002a}
\begin{eqnarray*}
COV(aX + b, cY + d) = a c COV(X,Y) \\
\sigma(aX+b) = |a|\sigma(X)\\
\sigma(cY+d) = |c|\sigma(Y),\\
\end{eqnarray*}
it follows that
\begin{equation*}
r(aX + b, cY + d) = \frac{a c}{|a||c|} r(X, Y).
\end{equation*}
Hence the statements of this property.
\end{proof}

% \subsection{Invariance of the Kendall $\tau$ correlation under strictly monotonic transformation}
\subsection{Invariance of Kendall \texorpdfstring{$\tau$}{tau} correlation under strictly monotonic transformation}

\label{app:Kendall_correlation_strictly_monotonic_transformation}

The Kendall $\tau$ correlation between to random variables $X$ and $Y$ can be defined on the probability that two pairs of values $(X_1, Y_1)$ and $(X_2, Y_2)$ are concordant, i.e.  
$$Pr\{(X_1 - X_2)(Y_1 - Y_2) > 0\},$$
and on the probability that they are discordant, i.e.
$$Pr\{(X_1 - X_2)(Y_1 - Y_2) < 0\},$$
as \parencite{vandenJeuvel2022a}
\begin{eqnarray*}
\tau & = & \tau(X,Y) \\ 
     & = & Pr\{(X_1 - X_2)(Y_1 - Y_2) > 0\} - Pr\{(X_1 - X_2)(Y_1 - Y_2) < 0\}.
\end{eqnarray*}
This definition corresponds to Kendall $\tau$-a \parencite{Kendall1970a}.

Kendall $\tau$ correlation $r(X,Y)$ is invariant under non-linear transformation but under certain conditions. If $X$ is replaced by applying some function $h_x$ and $Y$ is replaced by some function $h_y$, then $\tau$ is invariant when  both $h_x(X)$ and $h_y(Y)$ are strictly monotonically increasing or both $h_x(X)$ and $h_y(Y)$ are strictly monotonically decreasing (\textcite{vandenJeuvel2022a} and references therein). If only one variable is replaced, say $X$, then $h(X)$ must be strictly monotonically increasing. The argument is detailed in the following property and its proof.

\begin{property}
$\tau(X,Y)$ is invariant strictly monotonic linear transformation 
\begin{itemize}
\item
In a strict or strong sense, i.e. 
$$\tau(h_x(X), h_y(Y)) = \tau(X,Y)$$ 
if the trend of $h_x(X)$ and that of $h_y(Y)$ is the same (both increasing or both decreasing)
and 
$$\tau(h_x(X), h_y(Y)) = - \tau(X,Y)$$ 
if the trends differ (one is increasing and the other is decreasing) and
\item 
In a weak sense, i.e. 
$$|\tau(h_x(X), h_y(Y))| = |\tau(X,Y)|,$$ 
in any circumstance (namely independently of the trend of $h_x(X)$ and $h_y(Y)$). 
\end{itemize}
\end{property}

\begin{proof}
Notice that 
\begin{itemize}
    \item 
    $\tau$ is symmetric, namely, $\tau(X,Y) = \tau(Y,X)$.
    \item
    When $h_x(X)$ and $h_y(Y)$ are strictly monotonic (increasing or decreasing; the direction of $h_x(X)$ and $h_y(Y)$ does not need to be the same), 
$$ Pr\{(h_x(X_1) - h_x(X_2))(h_y(Y_1) - h_y(Y_2)) = 0\} = Pr\{(X_1 - X_2)(Y_1 - Y_2) = 0\}.$$
\end{itemize}

If we apply a strictly monotonically increasing ($i$) function $h_i$ to one of the variables, say $X$, 
we get that $\tau(h_i(X),Y) = \tau(X,Y)$. To see it, notice that  
$$Pr\{(h_i(X_1) - h_i(X_2))(Y_1  - Y_2) > 0\} = Pr\{(X_1 - X_2)(Y_1 - Y_2) > 0\}$$
and then
\begin{eqnarray*}
Pr\{(h_i(X_1) - h_i(X_2))(Y_1 - Y_2)) < 0\} & = & 1 - Pr\{(h_i(X_1) - h_i(X_2))(Y_1 - Y_2) > 0\} \\ 
  & & - Pr\{(h_i(X_1) - h_i(X_2))(Y_1 - Y_2) = 0\} \\
  & = & 1 - Pr\{(X_1 - X_2)(Y_1 - Y_2) > 0\} - Pr\{(X_1 - X_2)(Y_1 - Y_2) = 0\} \\
  & = & Pr\{(X_1 - X_2)(Y_1 - Y_2) < 0\}.
\end{eqnarray*}
By symmetry, we also have that $\tau(X, h_i(Y)) = \tau(X, Y)$. 

If we apply a strictly monotonically decreasing ($d$) function $h_d$ to one of the variables, say $X$, 
we get that $\tau(h_d(X),Y) = -\tau(X,Y)$. 
To see it, 
recall that 
$$Pr\{(h_d(X_1) - h_d(X_2))(Y_1  - Y_2) = 0\} = Pr\{(X_1 - X_2)(Y_1 - Y_2) = 0\}$$
and notice that  
$$Pr\{(h_d(X_1) - h_d(X_2))(Y_1  - Y_2) > 0\} = Pr\{(X_1 - X_2)(Y_1 - Y_2) < 0\}$$
and 
$$Pr\{(h_d(X_1) - h_d(X_2))(Y_1  - Y_2) < 0\} = Pr\{(X_1 - X_2)(Y_1 - Y_2) > 0\}.$$
% \textcolor{red}{Should we add a more detailed justification for the two last equations???}
Hence $\tau(h_d(X),Y) = -\tau(X,Y)$.
By symmetry, we also have that $\tau(X, h_d(Y)) = - \tau(X, Y)$.

Now suppose that we apply a strictly monotonic function $h_x$ to $X$ and 
a strictly monotonic function $h_y$ to $Y$. 
\begin{itemize}
\item 
    If both functions are increasing, by iterating on the arguments above, we get 
$$\tau(h_x(X), h_y(Y)) = \tau(X, h_y(Y)) = \tau(X, Y).$$ 
\item
    If both functions are decreasing, by iterating on the arguments above, we get 
$$\tau(h_x(X), h_y(Y)) = - \tau(X, h_y(Y)) = \tau(X, Y).$$ 
\item
    If one function is decreasing and the other is increasing, say $h_x$ is increasing and $h_y$ is decreasing, 
    $$\tau(h_x(X), h_y(Y)) = \tau(X, h_y(Y)) = - \tau(X, Y).$$
    If $h_x$ is decreasing and $h_y$ is increasing, $$\tau(h_x(X), h_y(Y)) = - \tau(X, Y)$$ again by the symmetry of $\tau$.
\end{itemize}
Hence the statements of this property.
\end{proof}

\subsection{Invariance under linear transformation}

\label{app:invariance_under_linear_transformation_theory}

% In the language of mathematics,

\begin{property}
\label{invariance_property}
We use a prime, $'$, to indicate the new value of a score after applying a linear transformation $g(l) = al +b$ to $l$ with $a > 0$.
Let $\eta''$ be the value of $\eta$ after applying a proportional change of scale to $l$, namely $g(l)$ with $b=0$. Then
\begin{eqnarray*}
\Psi = \Psi' & \Omega = \Omega' \\
r = r' & \tau = \tau'\\
\eta = \eta'' &
\end{eqnarray*}
while $L = L'$ and $\eta = \eta'$ do not generally hold. 
\end{property}

\begin{proof}
$r$ and $\tau$ are invariant under that linear transformation (\autoref{app:Pearson_correlation_linear_transformation} and \autoref{app:Kendall_correlation_strictly_monotonic_transformation}), hence $r = r'$ and $\tau = \tau'$ . 

Knowing that 
$$L = \sum_{i} p_i l_i,$$
it is easy to show that, after applying a linear transformation to word lengths ($g(l)$ can be increasing or decreasing), one obtains 
\begin{eqnarray*}
L' = aL + b \\
L_r' = a L_r + b.
\end{eqnarray*}
The condition $a > 0$ and Property \ref{minimum_baseline} give
\begin{equation}
L_{min}' = aL_{min} + b.
% \label{eq:new_minimum_length}
\end{equation}
Hence
\begin{eqnarray*}
\Psi'  & = & \frac{L_r' - L'}{L_r' - L_{min}'} \\
         & = & \frac{aL_r+b - aL-b}{aL_r+b - aL_{min}-b} \\
         & = & \frac{a(L_r - L)}{a(L_r - L_{min})} \\
         & = & \Psi.
\end{eqnarray*}
% Besides,
% \begin{eqnarray*}
%   |\tau(p, l)| = |\tau(p, g(l))| \\
%   |\tau_{min}(p, l)| = |\tau_{min}(p, g(l))|\\
%   \sgn(\tau(p, l)\tau(p, % g(l)))=\sgn(\tau_{min}(p, l)\tau_{min}(p, g(l)))
% \end{eqnarray*}
In contrast,
\begin{eqnarray*}
\eta'  & = & \frac{L_{min}}{L'} \\
       & = & \frac{aL_{min} + b}{aL +b}.
\end{eqnarray*}
When $b = 0$, $$\eta' = \eta'' = \eta.$$

Concerning $\Omega$, recall that $\tau$ is invariant under strictly increasing transformation (\autoref{app:Kendall_correlation_strictly_monotonic_transformation}). When $a>0$, the linear transformation is strictly increasing, hence the invariance of $\tau$ under that transformation gives 
$$\tau' = \tau(p,g(l)) = \tau(p,l) = \tau.$$
Besides, $a > 0$ and Property \ref{minimum_baseline} give $\tau_{min}' = \tau_{min}$.
\end{proof}

\subsection{Invariance under non-linear monotonic transformation}

\label{app:invariance_under_non_linear_monotonic_transformation_theory}

\begin{property}
\label{non_linear_invariance_property}
We use a prime, $'$, to indicate the new value of a score after applying a strictly increasing non-linear transformation $h(l)$ to $l$.
Then 
\begin{eqnarray*}
\tau = \tau' \\ 
\Omega = \Omega' 
\end{eqnarray*}
while 
\begin{eqnarray*}
L = L' \\
r = r' \\
\Psi = \Psi' \\
\eta = \eta' 
\end{eqnarray*}
do not generally hold. 
\end{property}
\begin{proof}
Since $h(l)$ is strictly increasing, then $\tau$ is invariant (\autoref{app:Pearson_correlation_linear_transformation} and \autoref{app:Kendall_correlation_strictly_monotonic_transformation}), i.e.   
$$\tau(p,l) = \tau(p,h(l)).$$
As $h(l)$ is monotonically increasing, Property \ref{minimum_baseline} gives
$$\tau_{min}'=\tau_{min}(p,h(l)) = \tau_{min}(p,l).$$
Finally, combining the results above
\begin{eqnarray*}
\Omega' & = & \frac{\tau(p,h(l))}{\tau_{min}(p,h(l))} \\
  & = & \frac{\tau(p,l)}{\tau_{min}(p,l)} \\
  & = & \Omega.
\end{eqnarray*}
$L' = L$ is trivially unwarranted.
To see that $r' = r$ is not warranted, suppose that $r(p, l) = 1$, namely the association between both variables is perfectly linear and increasing. If we apply a strictly increasing non-linear transformation $h(l)$, then $r(p,h(l)) < 1$ because the association between $p$, $h(l)$ is not linear.    
Similarly, $\Psi' = \Psi$ is unwarranted. A counterexample suffices. Consider configuration $C$ in \autoref{tab:example}. $L_r = 2$, $L=1.88$ and $L_{min}=1.14$ give $\Psi \approx 0.158$. Now suppose that we apply $h(l) = 2^l$ to C. Then $L_r' = \frac{14}{13}$, $L' = \frac{32}{25}$, and $L_{min}' = \frac{64}{75}$ give $\Psi' = 0.\overline{8}$. Hence $\Psi \neq \Psi'$. 
The finding is not surprising given that  $\Psi$ is a linear function of the Pearson correlation coefficient (\autoref{app:Pearson_correlation}). 
$\eta$ cannot be invariant under that transformation because it is not even invariant if the transformation is strictly increasing but linear with non-zero slope (Property \ref{non_linear_invariance_property}).
\end{proof}

\section{Analysis}
\label{app:analysis}

Here we present complementary analyses, tables and plots.

% optimality scores tables
\subsection{Optimality scores}
\label{app:opt_scores}

In \autoref{tab:opt_scores_pud}, \autoref{tab:opt_scores_cv_characters} and \autoref{tab:opt_scores_cv_meadianDuration}, we show the values of the optimality scores and their ingredients for PUD and for CV when length is measured in characters and also in duration, respectively.

%% PUD
\begin{table}[H]
\centering
\caption{Word lengths and their optimality in PUD. Word length is measured in number of characters. Han script is used in its traditional variant.}  
\label{tab:opt_scores_pud}
\centerline{
{\scriptsize
\begin{tabular}{lllrrrrr|rrr}
language & family & script & $L_{min}$ & $L$ & $L_r$ & $\tau$ & $\tau_{min}$ & $\eta$ & $\Psi$ & $\Omega$ \\ 
  \hline
% latex table generated in R 4.2.1 by xtable 1.8-4 package
% Mon Sep 25 18:10:44 2023
 Arabic & Afro-Asiatic & Arabic & 3.40 & 4.16 & 5.56 & -0.13 & -0.72 & 0.82 & 0.65 & 0.18 \\ 
  Indonesian & Austronesian & Latin & 4.03 & 5.89 & 7.18 & -0.09 & -0.80 & 0.68 & 0.41 & 0.11 \\ 
  Russian & Indo-European & Cyrillic & 5.18 & 6.04 & 8.08 & -0.19 & -0.64 & 0.86 & 0.71 & 0.30 \\ 
  Hindi & Indo-European & Devanagari & 3.11 & 4.09 & 5.85 & -0.19 & -0.78 & 0.76 & 0.64 & 0.24 \\ 
  Czech & Indo-European & Latin & 4.70 & 5.44 & 7.27 & -0.22 & -0.67 & 0.86 & 0.71 & 0.34 \\ 
  English & Indo-European & Latin & 3.77 & 4.86 & 6.99 & -0.20 & -0.76 & 0.78 & 0.66 & 0.26 \\ 
  French & Indo-European & Latin & 3.71 & 4.85 & 7.55 & -0.17 & -0.76 & 0.77 & 0.70 & 0.22 \\ 
  German & Indo-European & Latin & 4.55 & 5.79 & 8.55 & -0.23 & -0.68 & 0.79 & 0.69 & 0.33 \\ 
  Icelandic & Indo-European & Latin & 4.44 & 5.32 & 7.91 & -0.26 & -0.66 & 0.84 & 0.75 & 0.40 \\ 
  Italian & Indo-European & Latin & 3.89 & 4.89 & 7.72 & -0.16 & -0.76 & 0.80 & 0.74 & 0.22 \\ 
  Polish & Indo-European & Latin & 5.16 & 6.05 & 7.98 & -0.19 & -0.64 & 0.85 & 0.68 & 0.29 \\ 
  Portuguese & Indo-European & Latin & 3.68 & 4.68 & 7.50 & -0.17 & -0.74 & 0.79 & 0.74 & 0.24 \\ 
  Spanish & Indo-European & Latin & 3.75 & 4.85 & 7.57 & -0.16 & -0.74 & 0.77 & 0.71 & 0.21 \\ 
  Swedish & Indo-European & Latin & 4.30 & 5.41 & 7.99 & -0.23 & -0.68 & 0.80 & 0.70 & 0.34 \\ 
  Japanese & Japonic & Japanese & 1.46 & 1.75 & 2.76 & -0.22 & -0.60 & 0.84 & 0.78 & 0.36 \\ 
  Japanese-strokes & Japonic & Japanese & 5.15 & 7.70 & 13.91 & -0.09 & -0.78 & 0.67 & 0.71 & 0.12 \\ 
  Japanese-romaji & Japonic & Latin & 2.68 & 3.84 & 6.02 & -0.14 & -0.80 & 0.70 & 0.65 & 0.18 \\ 
  Korean & Koreanic & Hangul & 2.42 & 2.75 & 3.24 & -0.24 & -0.64 & 0.88 & 0.60 & 0.38 \\ 
  Thai & Kra-Dai & Thai & 3.16 & 4.33 & 6.33 & -0.23 & -0.82 & 0.73 & 0.63 & 0.28 \\ 
  Chinese & Sino-Tibetan & Han (Traditional variant) & 1.53 & 1.73 & 2.36 & -0.26 & -0.55 & 0.88 & 0.76 & 0.48 \\ 
  Chinese-strokes & Sino-Tibetan & Han (Traditional variant) & 10.47 & 15.00 & 21.59 & -0.18 & -0.77 & 0.70 & 0.59 & 0.24 \\ 
  Chinese-pinyin & Sino-Tibetan & Latin & 3.78 & 4.90 & 6.49 & -0.16 & -0.76 & 0.77 & 0.59 & 0.21 \\ 
  Turkish & Turkic & Latin & 5.26 & 6.35 & 7.87 & -0.24 & -0.67 & 0.83 & 0.58 & 0.36 \\ 
  Finnish & Uralic & Latin & 6.55 & 7.51 & 9.31 & -0.23 & -0.60 & 0.87 & 0.65 & 0.39 

\\
\end{tabular}
}
}
\end{table}

%% CV characters
\begin{table}[H]
\centering
\caption{Word lengths and their optimality in CV. Word length is measured in number of characters. 'Conlang' stands for 'constructed language', that is an artificially created language. This is not a family in the proper sense, and Conlang languages are not related in the common family sense.}
\label{tab:opt_scores_cv_characters}
{\scriptsize
\begin{tabular}{lllrrrrr|rrr}
language & family & script & $L_{min}$ & $L$ & $L_r$ & $\tau$ & $\tau_{min}$ & $\eta$ & $\Psi$ & $\Omega$ \\ 
  \hline
% latex table generated in R 4.1.1 by xtable 1.8-4 package
% Fri Jul 15 12:23:10 2022
 Arabic & Afro-Asiatic & Arabic & 3.26 & 4.10 & 5.06 & -0.14 & -0.87 & 0.80 & 0.53 & 0.16 \\ 
  Maltese & Afro-Asiatic & Latin & 3.72 & 5.07 & 7.35 & -0.20 & -0.81 & 0.73 & 0.63 & 0.24 \\ 
  Vietnamese & Austroasiatic & Latin & 2.75 & 3.24 & 3.47 & -0.19 & -0.73 & 0.85 & 0.33 & 0.26 \\ 
  Indonesian & Austronesian & Latin & 3.70 & 5.37 & 7.24 & -0.20 & -0.91 & 0.69 & 0.53 & 0.22 \\ 
  Esperanto & Conlang & Latin & 3.23 & 4.83 & 7.73 & -0.18 & -0.92 & 0.67 & 0.65 & 0.19 \\ 
  Interlingua & Conlang & Latin & 3.36 & 4.43 & 7.43 & -0.24 & -0.79 & 0.76 & 0.74 & 0.30 \\ 
  Tamil & Dravidian & Tamil & 4.65 & 5.68 & 7.08 & -0.28 & -0.88 & 0.82 & 0.58 & 0.32 \\ 
  Persian & Indo-European & Arabic & 2.69 & 3.80 & 5.49 & -0.21 & -0.92 & 0.71 & 0.60 & 0.22 \\ 
  Assamese & Indo-European & Assamese & 4.10 & 4.57 & 5.36 & -0.31 & -0.68 & 0.90 & 0.62 & 0.46 \\ 
  Russian & Indo-European & Cyrillic & 4.05 & 6.31 & 9.00 & -0.13 & -0.94 & 0.64 & 0.54 & 0.14 \\ 
  Ukrainian & Indo-European & Cyrillic & 4.52 & 5.52 & 7.67 & -0.16 & -0.86 & 0.82 & 0.68 & 0.19 \\ 
  Panjabi & Indo-European & Devanagari & 3.55 & 3.68 & 3.88 & -0.32 & -0.48 & 0.96 & 0.60 & 0.68 \\ 
  Modern Greek & Indo-European & Greek & 3.61 & 4.85 & 7.64 & -0.24 & -0.85 & 0.75 & 0.69 & 0.29 \\ 
  Breton & Indo-European & Latin & 2.90 & 3.97 & 6.31 & -0.24 & -0.90 & 0.73 & 0.69 & 0.26 \\ 
  Catalan & Indo-European & Latin & 2.99 & 4.90 & 8.58 & -0.15 & -0.92 & 0.61 & 0.66 & 0.17 \\ 
  Czech & Indo-European & Latin & 3.81 & 4.83 & 7.17 & -0.22 & -0.92 & 0.79 & 0.69 & 0.24 \\ 
  Dutch & Indo-European & Latin & 3.24 & 4.72 & 8.26 & -0.28 & -0.95 & 0.69 & 0.71 & 0.29 \\ 
  English & Indo-European & Latin & 2.39 & 4.61 & 7.79 & -0.07 & -0.83 & 0.52 & 0.59 & 0.09 \\ 
  French & Indo-European & Latin & 2.67 & 5.04 & 8.13 & -0.04 & -0.85 & 0.53 & 0.57 & 0.04 \\ 
  German & Indo-European & Latin & 3.00 & 5.73 & 10.30 & -0.12 & -0.87 & 0.52 & 0.63 & 0.13 \\ 
  Irish & Indo-European & Latin & 3.01 & 4.20 & 6.58 & -0.21 & -0.93 & 0.71 & 0.66 & 0.23 \\ 
  Italian & Indo-European & Latin & 3.16 & 5.29 & 8.16 & -0.06 & -0.90 & 0.60 & 0.57 & 0.06 \\ 
  Latvian & Indo-European & Latin & 3.95 & 4.79 & 7.09 & -0.26 & -0.73 & 0.82 & 0.73 & 0.35 \\ 
  Polish & Indo-European & Latin & 3.82 & 5.27 & 7.87 & -0.17 & -0.94 & 0.72 & 0.64 & 0.18 \\ 
  Portuguese & Indo-European & Latin & 3.14 & 4.53 & 7.49 & -0.19 & -0.91 & 0.69 & 0.68 & 0.21 \\ 
  Romanian & Indo-European & Latin & 3.61 & 5.03 & 7.67 & -0.21 & -0.78 & 0.72 & 0.65 & 0.27 \\ 
  Romansh & Indo-European & Latin & 3.83 & 4.94 & 7.56 & -0.24 & -0.79 & 0.78 & 0.70 & 0.30 \\ 
  Slovenian & Indo-European & Latin & 3.65 & 4.56 & 6.43 & -0.21 & -0.87 & 0.80 & 0.67 & 0.24 \\ 
  Spanish & Indo-European & Latin & 2.74 & 5.01 & 7.92 & -0.03 & -0.91 & 0.55 & 0.56 & 0.04 \\ 
  Swedish & Indo-European & Latin & 3.08 & 4.04 & 6.87 & -0.28 & -0.90 & 0.76 & 0.75 & 0.31 \\ 
  Welsh & Indo-European & Latin & 2.79 & 4.17 & 7.05 & -0.21 & -0.91 & 0.67 & 0.68 & 0.23 \\ 
  Western Frisian & Indo-European & Latin & 3.44 & 4.38 & 7.99 & -0.29 & -0.80 & 0.79 & 0.79 & 0.36 \\ 
  Oriya & Indo-European & Odia & 3.68 & 4.21 & 5.35 & -0.33 & -0.73 & 0.87 & 0.68 & 0.46 \\ 
  Dhivehi & Indo-European & Thaana & 1.97 & 3.32 & 7.61 & -0.16 & -0.84 & 0.59 & 0.76 & 0.19 \\ 
  Georgian & Kartvelian & Georgian & 5.91 & 7.17 & 8.22 & -0.12 & -0.67 & 0.82 & 0.45 & 0.18 \\ 
  Basque & Language isolate & Latin & 4.44 & 6.41 & 8.89 & -0.16 & -0.91 & 0.69 & 0.56 & 0.18 \\ 
  Mongolian & Mongolic & Mongolian & 4.18 & 5.47 & 7.31 & -0.23 & -0.86 & 0.76 & 0.59 & 0.26 \\ 
  Kinyarwanda & Niger-Congo & Latin & 3.72 & 6.13 & 9.20 & -0.19 & -0.90 & 0.61 & 0.56 & 0.21 \\ 
  Abkhazian & Northwest Caucasian & Cyrillic & 5.59 & 5.94 & 6.42 & -0.32 & -0.55 & 0.94 & 0.58 & 0.59 \\ 
  Hakha Chin & Sino-Tibetan & Latin & 2.56 & 3.29 & 5.29 & -0.29 & -0.81 & 0.78 & 0.73 & 0.35 \\ 
  Chuvash & Turkic & Cyrillic & 4.79 & 6.00 & 7.35 & -0.22 & -0.83 & 0.80 & 0.53 & 0.27 \\ 
  Kirghiz & Turkic & Cyrillic & 4.49 & 6.01 & 7.78 & -0.19 & -0.89 & 0.75 & 0.54 & 0.22 \\ 
  Tatar & Turkic & Cyrillic & 4.04 & 5.41 & 7.45 & -0.24 & -0.89 & 0.75 & 0.60 & 0.27 \\ 
  Yakut & Turkic & Cyrillic & 5.02 & 6.32 & 7.99 & -0.26 & -0.74 & 0.79 & 0.56 & 0.36 \\ 
  Turkish & Turkic & Latin & 4.60 & 6.00 & 8.09 & -0.22 & -0.89 & 0.77 & 0.60 & 0.24 \\ 
  Estonian & Uralic & Latin & 4.68 & 6.16 & 8.85 & -0.24 & -0.84 & 0.76 & 0.65 & 0.29 

\\
\end{tabular}
}
\end{table}

%% CV duration
\begin{table}[H]
\centering
\caption{Word lengths and their optimality in CV. Word length is measured in duration.
The format is the same as in \autoref{tab:opt_scores_cv_characters}.
%'Conlang' stands for 'constructed language', that is an artificially created language. This is not a family in the proper sense, and Conlang languages are not related in the common family sense.
} 
\label{tab:opt_scores_cv_meadianDuration}
{\scriptsize
\begin{tabular}{lllrrrrr|rrr}
language & family & script & $L_{min}$ & $L$ & $L_r$ & $\tau$ & $\tau_{min}$ & $\eta$ & $\Psi$ & $\Omega$ \\ 
  \hline
% latex table generated in R 4.1.1 by xtable 1.8-4 package
% Fri Jul 15 12:23:10 2022
 Arabic & Afro-Asiatic & Arabic & 0.35 & 0.46 & 0.58 & -0.12 & -0.89 & 0.76 & 0.52 & 0.14 \\ 
  Maltese & Afro-Asiatic & Latin & 0.26 & 0.35 & 0.54 & -0.21 & -0.81 & 0.74 & 0.68 & 0.26 \\ 
  Vietnamese & Austroasiatic & Latin & 0.21 & 0.29 & 0.33 & -0.07 & -0.80 & 0.72 & 0.33 & 0.09 \\ 
  Indonesian & Austronesian & Latin & 0.27 & 0.38 & 0.52 & -0.22 & -0.91 & 0.72 & 0.58 & 0.24 \\ 
  Esperanto & Conlang & Latin & 0.35 & 0.49 & 0.81 & -0.18 & -0.91 & 0.71 & 0.69 & 0.20 \\ 
  Interlingua & Conlang & Latin & 0.32 & 0.40 & 0.69 & -0.24 & -0.79 & 0.81 & 0.79 & 0.31 \\ 
  Tamil & Dravidian & Tamil & 0.45 & 0.54 & 0.66 & -0.31 & -0.87 & 0.84 & 0.60 & 0.36 \\ 
  Persian & Indo-European & Arabic & 0.26 & 0.36 & 0.54 & -0.25 & -0.99 & 0.73 & 0.65 & 0.25 \\ 
  Assamese & Indo-European & Assamese & 0.36 & 0.43 & 0.50 & -0.22 & -0.68 & 0.84 & 0.52 & 0.32 \\ 
  Russian & Indo-European & Cyrillic & 0.30 & 0.42 & 0.60 & -0.15 & -0.94 & 0.71 & 0.60 & 0.15 \\ 
  Ukrainian & Indo-European & Cyrillic & 0.37 & 0.43 & 0.59 & -0.18 & -0.86 & 0.85 & 0.72 & 0.20 \\ 
  Panjabi & Indo-European & Devanagari & 0.65 & 0.70 & 0.73 & -0.18 & -0.45 & 0.92 & 0.38 & 0.40 \\ 
  Modern Greek & Indo-European & Greek & 0.29 & 0.38 & 0.63 & -0.21 & -0.84 & 0.76 & 0.72 & 0.26 \\ 
  Breton & Indo-European & Latin & 0.22 & 0.31 & 0.51 & -0.25 & -0.88 & 0.72 & 0.71 & 0.28 \\ 
  Catalan & Indo-European & Latin & 0.24 & 0.35 & 0.68 & -0.21 & -0.94 & 0.67 & 0.73 & 0.23 \\ 
  Czech & Indo-European & Latin & 0.30 & 0.37 & 0.57 & -0.21 & -0.92 & 0.81 & 0.74 & 0.23 \\ 
  Dutch & Indo-European & Latin & 0.22 & 0.29 & 0.55 & -0.28 & -0.95 & 0.75 & 0.78 & 0.29 \\ 
  English & Indo-European & Latin & 0.19 & 0.33 & 0.67 & -0.17 & -0.83 & 0.59 & 0.72 & 0.21 \\ 
  French & Indo-European & Latin & 0.21 & 0.32 & 0.63 & -0.21 & -0.86 & 0.64 & 0.73 & 0.24 \\ 
  German & Indo-European & Latin & 0.25 & 0.37 & 0.76 & -0.22 & -0.87 & 0.67 & 0.76 & 0.25 \\ 
  Irish & Indo-European & Latin & 0.22 & 0.30 & 0.47 & -0.24 & -0.93 & 0.76 & 0.71 & 0.26 \\ 
  Italian & Indo-European & Latin & 0.26 & 0.38 & 0.65 & -0.19 & -0.90 & 0.70 & 0.71 & 0.21 \\ 
  Latvian & Indo-European & Latin & 0.32 & 0.39 & 0.59 & -0.23 & -0.74 & 0.84 & 0.77 & 0.31 \\ 
  Polish & Indo-European & Latin & 0.30 & 0.38 & 0.57 & -0.17 & -0.95 & 0.79 & 0.71 & 0.18 \\ 
  Portuguese & Indo-European & Latin & 0.27 & 0.35 & 0.61 & -0.22 & -0.92 & 0.76 & 0.76 & 0.24 \\ 
  Romanian & Indo-European & Latin & 0.27 & 0.36 & 0.57 & -0.23 & -0.77 & 0.74 & 0.69 & 0.30 \\ 
  Romansh & Indo-European & Latin & 0.33 & 0.41 & 0.66 & -0.26 & -0.78 & 0.82 & 0.77 & 0.34 \\ 
  Slovenian & Indo-European & Latin & 0.36 & 0.44 & 0.63 & -0.25 & -0.85 & 0.83 & 0.72 & 0.30 \\ 
  Spanish & Indo-European & Latin & 0.22 & 0.36 & 0.62 & -0.14 & -0.91 & 0.63 & 0.67 & 0.15 \\ 
  Swedish & Indo-European & Latin & 0.22 & 0.27 & 0.52 & -0.29 & -0.90 & 0.81 & 0.83 & 0.33 \\ 
  Welsh & Indo-European & Latin & 0.22 & 0.32 & 0.58 & -0.20 & -0.93 & 0.70 & 0.73 & 0.22 \\ 
  Western Frisian & Indo-European & Latin & 0.26 & 0.32 & 0.61 & -0.31 & -0.80 & 0.80 & 0.82 & 0.39 \\ 
  Oriya & Indo-European & Odia & 0.35 & 0.39 & 0.49 & -0.33 & -0.72 & 0.89 & 0.70 & 0.45 \\ 
  Dhivehi & Indo-European & Thaana & 0.14 & 0.32 & 0.71 & -0.17 & -0.82 & 0.46 & 0.70 & 0.21 \\ 
  Georgian & Kartvelian & Georgian & 0.44 & 0.52 & 0.61 & -0.15 & -0.65 & 0.85 & 0.52 & 0.23 \\ 
  Basque & Language isolate & Latin & 0.33 & 0.44 & 0.63 & -0.21 & -0.93 & 0.73 & 0.61 & 0.22 \\ 
  Mongolian & Mongolic & Mongolian & 0.29 & 0.36 & 0.48 & -0.25 & -0.85 & 0.80 & 0.62 & 0.29 \\ 
  Kinyarwanda & Niger-Congo & Latin & 0.27 & 0.44 & 0.72 & -0.21 & -0.89 & 0.61 & 0.62 & 0.24 \\ 
  Abkhazian & Northwest Caucasian & Cyrillic & 0.67 & 0.74 & 0.81 & -0.20 & -0.55 & 0.90 & 0.47 & 0.37 \\ 
  Hakha Chin & Sino-Tibetan & Latin & 0.22 & 0.29 & 0.44 & -0.25 & -0.83 & 0.76 & 0.69 & 0.30 \\ 
  Chuvash & Turkic & Cyrillic & 0.36 & 0.44 & 0.54 & -0.26 & -0.82 & 0.82 & 0.57 & 0.32 \\ 
  Kirghiz & Turkic & Cyrillic & 0.34 & 0.44 & 0.57 & -0.20 & -0.89 & 0.78 & 0.56 & 0.22 \\ 
  Tatar & Turkic & Cyrillic & 0.31 & 0.38 & 0.52 & -0.26 & -0.88 & 0.80 & 0.64 & 0.29 \\ 
  Yakut & Turkic & Cyrillic & 0.35 & 0.43 & 0.54 & -0.25 & -0.73 & 0.81 & 0.56 & 0.35 \\ 
  Turkish & Turkic & Latin & 0.32 & 0.41 & 0.54 & -0.21 & -0.89 & 0.78 & 0.60 & 0.23 \\ 
  Estonian & Uralic & Latin & 0.32 & 0.39 & 0.58 & -0.23 & -0.82 & 0.81 & 0.71 & 0.28 

\\
\end{tabular}
}
\end{table}

% \subsection{Substituting Kendall $\tau$ correlation with Spearman $\rho$ correlation in the definition of $\Omega$}
\subsection{Substituting Kendall \texorpdfstring{$\tau$}{tau} correlation with Spearman \texorpdfstring{$\rho$}{rho} correlation in the definition of \texorpdfstring{$\Omega$}{Omega}}

\label{app:variant_of_Omega} 

% We use Spearman's $\rho$ correlation instead of Kendall's $\tau$ in the definition of $\Omega$ so as to understand if a different rank-correlation coefficient would lead to a change in the scale of variation of values of $\Omega$, turning them closer to those of $\Psi$. For every language in PUD, \autoref{tab:omega_tau_omega_ro_pud} shows the value of $\Omega$ based on Kendall $\tau$ correlation, indicated as $\Omega_{\tau}$, the value of $\Omega$ based on Spearman $\rho$ correlation, indicated as $\Omega_{\rho}$, and the values of the correlations involved in their calculation. When using $\rho$, the resulting values of $\Omega$ are very similar to the original ones, with slight increments not larger than 0.03.

We use Spearman's $\rho$ correlation instead of Kendall's $\tau$ in the definition of $\Omega$ so as to understand if a different rank-correlation coefficient would lead to a change in the scale of variation of values of $\Omega$, turning them closer to those of $\Psi$.
In \autoref{tab:omega_tau_omega_ro_pud}, we show $\Omega_\tau$, the original definition of $\Omega$, $\Omega_\rho$, the new definition of $\Omega$ that results from replacing $\tau$ by $\rho$, and all the correlations necessary to compute them. The values of $\Omega_\rho$ increase with respect to those of $\Omega_\tau$, but by no more than 10\%, suggesting that the rather low values of $\Omega$ compared to those of $\Psi$, are not due to the initial choice of $\tau$ to define $\Omega$ (\autoref{eq:Omega}).  
We suspect that the low values of $\Omega_\rho$ and $\Omega_\tau$ could originate from some similarity between $\tau$ and $\rho$ (e.g. both are rank correlation scores and then both may yield low values of $\Omega$) or the fact that the template of the definition of $\Omega_\rho$ and that of $\Omega_\tau$ is the same ($\Omega_\rho$ and $\Omega_\tau$ differ only in the choice of the correlation coefficient). However, that issue should be the subject of further research.

% pud
\begin{table}[H]
\centering
\caption{Kendall $\tau$ versus Spearman $\rho$ in the frequency-length correlation and also in $\Omega$ in the PUD collection. $\tau$ and $\rho$ are the original correlations and $\tau_{min}$ and $\rho_{min}$ are the minimum correlations according to the rank ordering minimum baseline. 
$\Omega_{\tau}$ is the original value of $\Omega$ whereas 
$\Omega_{\rho}$ is the value of $\Omega$ that is obtained when Kendall $\tau$ is replaced by Spearman $\rho$ in the definition of $\Omega$. 
} 
\label{tab:omega_tau_omega_ro_pud}
{\scriptsize
\begin{tabular}{lllrrrr|rr}
language & family & script & $\tau$ & $\rho$ & $\tau_{min}$ & $\rho_{min}$ & $\Omega_{\tau}$ & $\Omega_{\rho}$ \\ 
  \hline
% latex table generated in R 4.0.5 by xtable 1.8-4 package
% Wed Sep 06 18:29:37 2023
 Arabic & Afro-Asiatic & Arabic & -0.13 & -0.16 & -0.74 & -0.82 & 0.18 & 0.19 \\ 
  Indonesian & Austronesian & Latin & -0.09 & -0.11 & -0.81 & -0.89 & 0.11 & 0.12 \\ 
  Russian & Indo-European & Cyrillic & -0.19 & -0.23 & -0.64 & -0.73 & 0.30 & 0.32 \\ 
  Hindi & Indo-European & Devanagari & -0.19 & -0.23 & -0.78 & -0.86 & 0.24 & 0.27 \\ 
  Czech & Indo-European & Latin & -0.22 & -0.27 & -0.67 & -0.75 & 0.34 & 0.36 \\ 
  English & Indo-European & Latin & -0.20 & -0.24 & -0.76 & -0.85 & 0.26 & 0.28 \\ 
  French & Indo-European & Latin & -0.16 & -0.20 & -0.75 & -0.83 & 0.21 & 0.23 \\ 
  German & Indo-European & Latin & -0.23 & -0.28 & -0.68 & -0.77 & 0.33 & 0.36 \\ 
  Icelandic & Indo-European & Latin & -0.26 & -0.32 & -0.66 & -0.75 & 0.40 & 0.42 \\ 
  Italian & Indo-European & Latin & -0.15 & -0.18 & -0.74 & -0.83 & 0.20 & 0.22 \\ 
  Polish & Indo-European & Latin & -0.18 & -0.22 & -0.65 & -0.73 & 0.29 & 0.30 \\ 
  Portuguese & Indo-European & Latin & -0.19 & -0.24 & -0.74 & -0.83 & 0.26 & 0.28 \\ 
  Spanish & Indo-European & Latin & -0.16 & -0.19 & -0.74 & -0.83 & 0.21 & 0.23 \\ 
  Swedish & Indo-European & Latin & -0.23 & -0.28 & -0.68 & -0.77 & 0.34 & 0.36 \\ 
  Japanese & Japonic & Japanese & -0.21 & -0.25 & -0.60 & -0.67 & 0.35 & 0.37 \\ 
  Japanese-strokes & Japonic & Japanese & -0.09 & -0.12 & -0.78 & -0.87 & 0.12 & 0.13 \\ 
  Japanese-romaji & Japonic & Latin & -0.15 & -0.19 & -0.80 & -0.87 & 0.19 & 0.21 \\ 
  Korean & Koreanic & Hangul & -0.24 & -0.27 & -0.64 & -0.71 & 0.38 & 0.39 \\ 
  Thai & Kra-Dai & Thai & -0.23 & -0.29 & -0.82 & -0.90 & 0.28 & 0.32 \\ 
  Chinese & Sino-Tibetan & Han (Traditional variant) & -0.24 & -0.27 & -0.55 & -0.59 & 0.44 & 0.45 \\ 
  Chinese-strokes & Sino-Tibetan & Han (Traditional variant) & -0.18 & -0.24 & -0.77 & -0.87 & 0.24 & 0.27 \\ 
  Chinese-pinyin & Sino-Tibetan & Latin & -0.18 & -0.21 & -0.77 & -0.85 & 0.23 & 0.25 \\ 
  Turkish & Turkic & Latin & -0.24 & -0.29 & -0.67 & -0.76 & 0.36 & 0.38 \\ 
  Finnish & Uralic & Latin & -0.23 & -0.28 & -0.60 & -0.69 & 0.39 & 0.41 
 
\\
\end{tabular}
}
\end{table}

\subsection{Stability under the null hypothesis}

\label{app:stability_under_null_hypothesis_test}

% We estimate $\expect[\eta]$, $\expect[\Psi]$ and $\expect[\Omega]$ with a Monte Carlo procedure over a maximum of $10^6$ randomizations to check if  $\eta$ is not stable under the null hypothesis of a random mapping of word frequencies into lengths, while $\Psi$ and $\Omega$ both yield values whose distributions is centered around 0 as expected from theory (\autoref{tab:mathematical_properties_of_scores})

We estimate $\expect[\eta]$, $\expect[\Psi]$, and $\expect[\Omega]$ under the null hypothesis of a random mapping of word frequencies into lengths by means of a Monte Carlo procedure over a maximum of $10^6$ randomizations. The goal is to confirm that $\Psi$ and $\Omega$ both yield values tending to zero (\autoref{tab:mathematical_properties_of_scores}) while checking if 
$\eta$ is eventually stable under that null hypothesis.

% summary statistics of Omega and eta
\begin{table}[H]
\centering
\caption{Summary statistics of estimates of $\expect[\eta]$, $\expect[\Psi]$ and $\expect[\Omega]$ under the null hypothesis, for languages in every collection and each definition of length in CV. In PUD, scores in strokes for Chinese and Japanese are excluded for the sake of homogeneity. 'sd' stands for standard deviation.}
\label{tab:opt_summary_null_kendall}

\begin{tabular}{rr|rrrrrrr}
score & collection & Min. & 1st Qu. & Median & Mean & 3rd Qu. & Max. & sd \\
\hline
\multirow{3}{*}{$\expect[\eta]$} 
% latex table generated in R 4.0.5 by xtable 1.8-4 package
% Thu Sep 14 15:07:29 2023
  & PUD-characters & 0.472 & 0.526 & 0.552 & 0.580 & 0.647 & 0.747 & 0.079 \\ 
   & CV-characters & 0.268 & 0.423 & 0.495 & 0.518 & 0.575 & 0.915 & 0.145 \\ 
   & CV-duration & 0.213 & 0.426 & 0.504 & 0.518 & 0.603 & 0.882 & 0.139 \\ 
   \hline

\multirow{3}{*}{$\expect[\Psi]$}  % latex table generated in R 4.0.5 by xtable 1.8-4 package
% Thu Sep 14 15:07:29 2023
  & PUD-characters & -0.000 & -0.000 & 0.000 & -0.000 & 0.000 & 0.000 & 0.000 \\ 
   & CV-characters & -0.000 & -0.000 & 0.000 & 0.000 & 0.000 & 0.000 & 0.000 \\ 
   & CV-duration & -0.000 & -0.000 & -0.000 & 0.000 & 0.000 & 0.001 & 0.000 \\ 
   \hline

\multirow{3}{*}{$\expect[\Omega]$} % latex table generated in R 4.0.5 by xtable 1.8-4 package
% Thu Sep 14 15:07:29 2023
  & PUD-characters & -0.000 & -0.000 & -0.000 & -0.000 & 0.000 & 0.000 & 0.000 \\ 
   & CV-characters & -0.000 & -0.000 & 0.000 & -0.000 & 0.000 & 0.000 & 0.000 \\ 
   & CV-duration & -0.000 & -0.000 & 0.000 & 0.000 & 0.000 & 0.000 & 0.000

\\
\end{tabular} 
\end{table}

\begin{figure}[H]
    \centering
    \includegraphics[width = 0.9\linewidth]{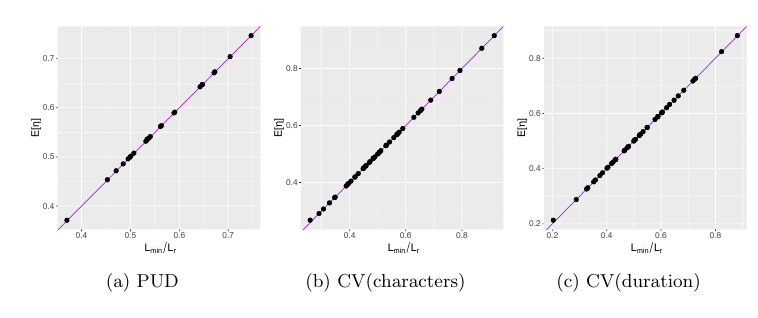}
    \caption{\label{fig:E[eta]_LminLr} The Monte  Carlo estimate of $\expect[\eta]$ against its theoretical lower bound, namely $\frac{L_{min}}{L_r}$. The purple line indicates the identity function, $\expect[\eta] = \frac{L_{min}}{L_r}$. (a) PUD collection with length measured in characters. (b) CV collection with length measured in characters. (c) CV collection with length measured in duration. 
    } 
\end{figure}

The summary in \autoref{tab:opt_summary_null_kendall} confirms the predictions in \autoref{app:stability_under_null_hypothesis_theory}, namely that $\expect[\Psi]$ and $\expect[\Omega]$
are close to 0, while $\expect[\eta]$ takes values spread across its whole domain, ranging from a minimum of 0.21 in CV (duration) to a maximum of 0.91 in CV (characters). However, the values taken by $\expect[\eta]$ are not arbitrary: the lower bound derived in 
\autoref{app:stability_under_null_hypothesis_theory}
is actually a very accurate estimate of the expectation itself (\autoref{fig:E[eta]_LminLr}).

% eta and Lmin/L

\section{Desirable properties of the scores}
\label{app:desirable}

Here we investigate empirically the desirable properties of scores presented in \autoref{sec:other_desirable_properties}.

\subsection{Dependence of the scores on basic parameters}

First, we explore the association between the optimality scores ($\eta, \Psi,\Omega $) and basic language parameters: observed alphabet size $A$ (number of distinct characters), observed vocabulary size $n$ (number of types) and text length $T$ (number of tokens). The alphabet and vocabulary size observed in the analyzed text samples are lower bounds of the true values in the language due to undersampling. However, the real parameters are likely to be a strictly monotonic function of the observed ones. For this reason, we measure the association with Kendall $\tau$ correlation, that is known to be invariant under strictly monotonically increasing transformations (\autoref{app:theory}). Given the recent challenges to the common view about the limits of Pearson $r$ correlation \parencite{vandenJeuvel2022a}, we also use $r$ to verify the robustness of the results.

\begin{figure}[H]
    \centering
    \includegraphics[width = 0.9\linewidth]{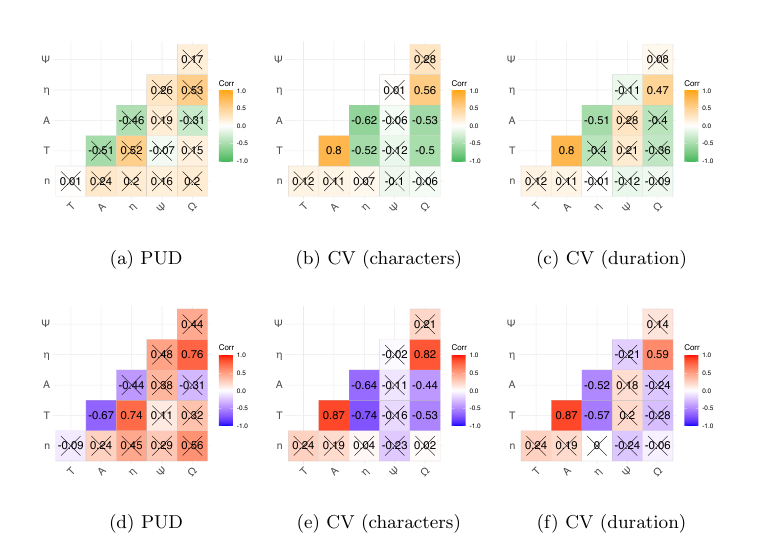}
    \caption{\label{fig:correlograms_params} Correlations between  optimality scores ($\eta$, $\Psi$ and $\Omega$) and basic language parameters ($T$, $n$ and $A$). Recall that $T$ is the number of tokens, $n$ is the number of types, $A$ is the alphabet size. Correlation tests are two-sided and $p$-values are corrected using Holm-Bonferroni correction. A crossed value indicates a non significant correlation coefficient at a 95\% confidence level. For PUD, only immediate word constituents are used to measure word length in Chinese and Japanese. (a) Kendall $\tau$ correlation in PUD collection with word length measured in characters. (b) Kendall $\tau$ correlation in the CV collection with word length measured in characters. (c) Kendall $\tau$ correlation for CV collection with word length measured in duration. (d) Same as (a) using Pearson correlation. (e) Same as (b) using Pearson correlation. (f) Same as (c) using Pearson correlation.   
    }
\end{figure}

In the parallel corpus PUD, where there is less diversity in terms of basic parameters, the only significant association found for a score is that between $\eta$ and $T$, when Pearson correlation is used
% no significant association is found for any score, both with Kendall and with Pearson correlation 
(\autoref{fig:correlograms_params} (a, d)). Concerning the CV collection, $\eta$ and $\Omega$ show sensitivity to the alphabet and sample size of a language, especially when length is measured in characters (\autoref{fig:correlograms_params} (b, e)). However, the association of the scores with alphabet size might be a reflection of the strong correlation existing between the amount of tokens and the subset of the real alphabet that can be observed in them.
Interestingly, when length is measured in duration, the association with $\eta$ remains while no significant association is detected between $\Omega$ and basic parameters, consistently for the two correlation coefficients (\autoref{fig:correlograms_params} (c, f)).
Thus, $\Psi$ is the only score among the analysed ones that does not seem to be related to $A$, $n$, or $T$, even in a highly heterogeneous collection.

\subsection{Convergence speed}

\label{app:convergence}

We investigate the dependence of $\eta$, $\Psi$, and $\Omega$ on text size $t$ (in number of tokens) within each language. We wish to know whether they converge or not and their convergence speed.

\subsubsection*{Methods}

For a given $t$, we sample $t$ tokens uniformly at random from the whole text. We do not use a prefix of length $t$ of the text as in related research on convergence of entropy estimators \parencite{Bentz2017a} because both PUD and CV contain a series of disconnected groups of sentences. In PUD each group of sentences is taken from a distinct text source and groups consist of a few sentences, often just one sentence. By proceeding in this way, we are easing the convergence of the estimators of the scores that we use, that neglect word order in their definition. We explore $t$ by increasing powers of 2. For each $t$, we perform $10^2$ experiments and compute the average over the available values. Indeed, for small sample sizes, computations could result in divisions by 0, or the impossibility to compute $\tau$ (for $\Omega$) given the low amount of distinct types, thus not yielding a value for that particular experiment. 

\subsubsection*{Results}

\begin{figure}[H]
    \centering
    \includegraphics[width = 0.8\linewidth]{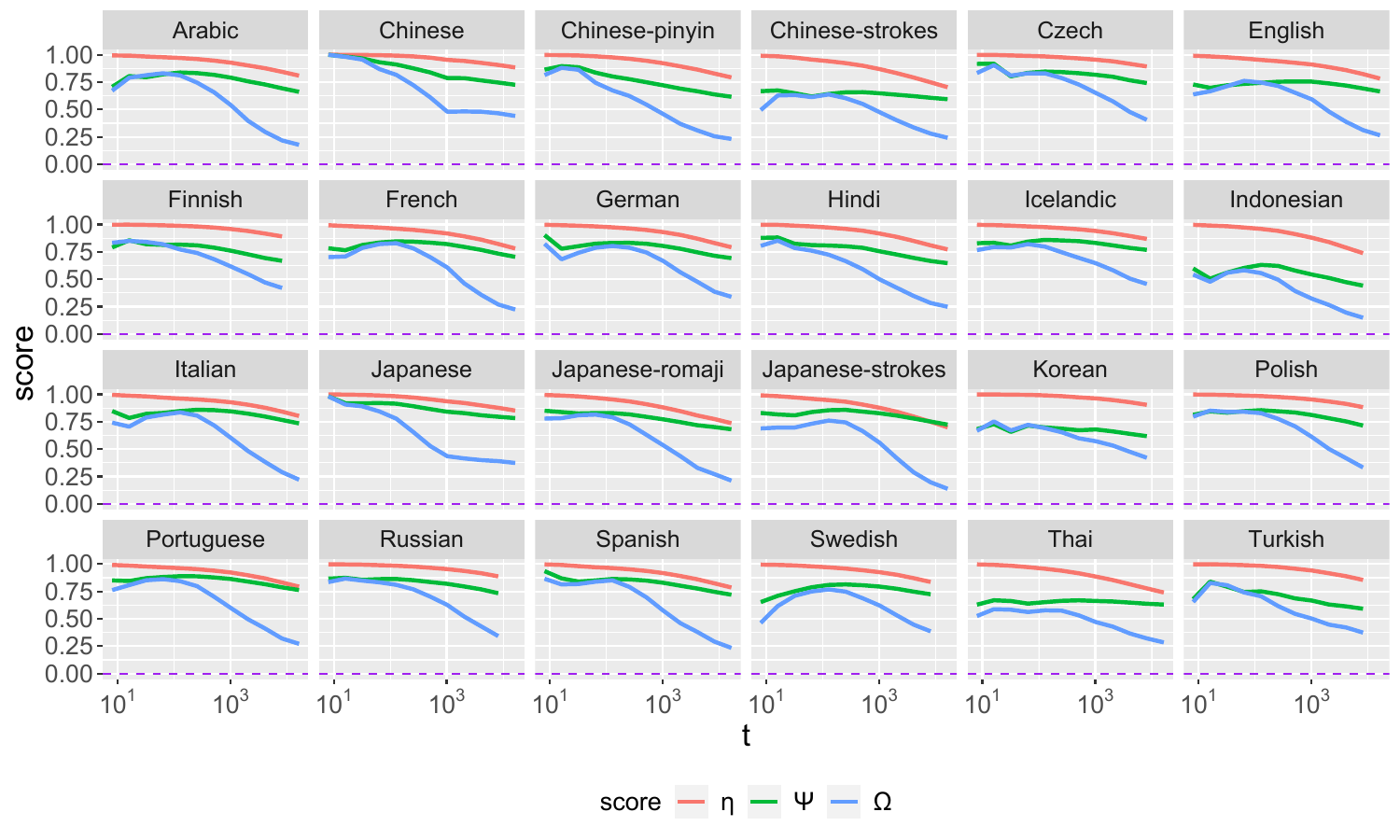}
    \caption{\label{fig:convergence_pud} Convergence of the scores in the PUD collection. Values of the optimality scores ($\eta$, $\Psi$, and $\Omega$) for increasing number of tokens $t$. For each value of $t$, we show the average value of the score over $10^{2}$ random experiments (only those for which a value could be computed). The dashed line marks 0.}
\end{figure}

\begin{figure}[H]
    \centering
    \includegraphics[width = 0.9\linewidth]{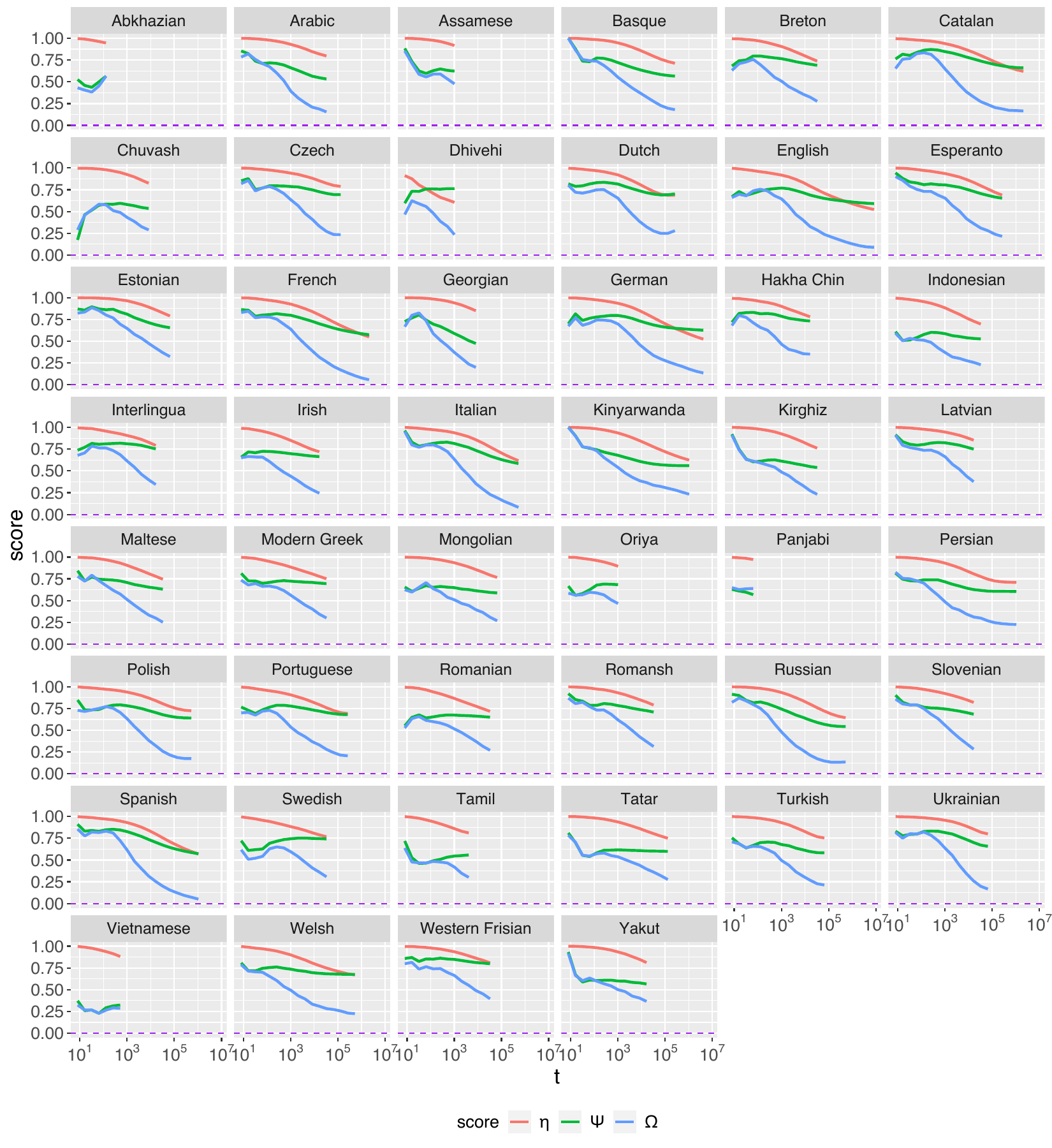}
    \caption{\label{fig:convergence_cv_characters} Convergence of the scores in the CV collection with word length measured in characters. The format is the same as in \autoref{fig:convergence_pud}.
    % Values of the optimality scores ($\eta$, $\Psi$, and $\Omega$) for increasing number of tokens $t$. For each value of $t$, we show the average value of the score over $10^{2}$ random experiments (only those for which a value could be computed). The dashed line marks 0.
    }
\end{figure}

\begin{figure}[H]
    \centering
    \includegraphics[width = 0.9\linewidth]{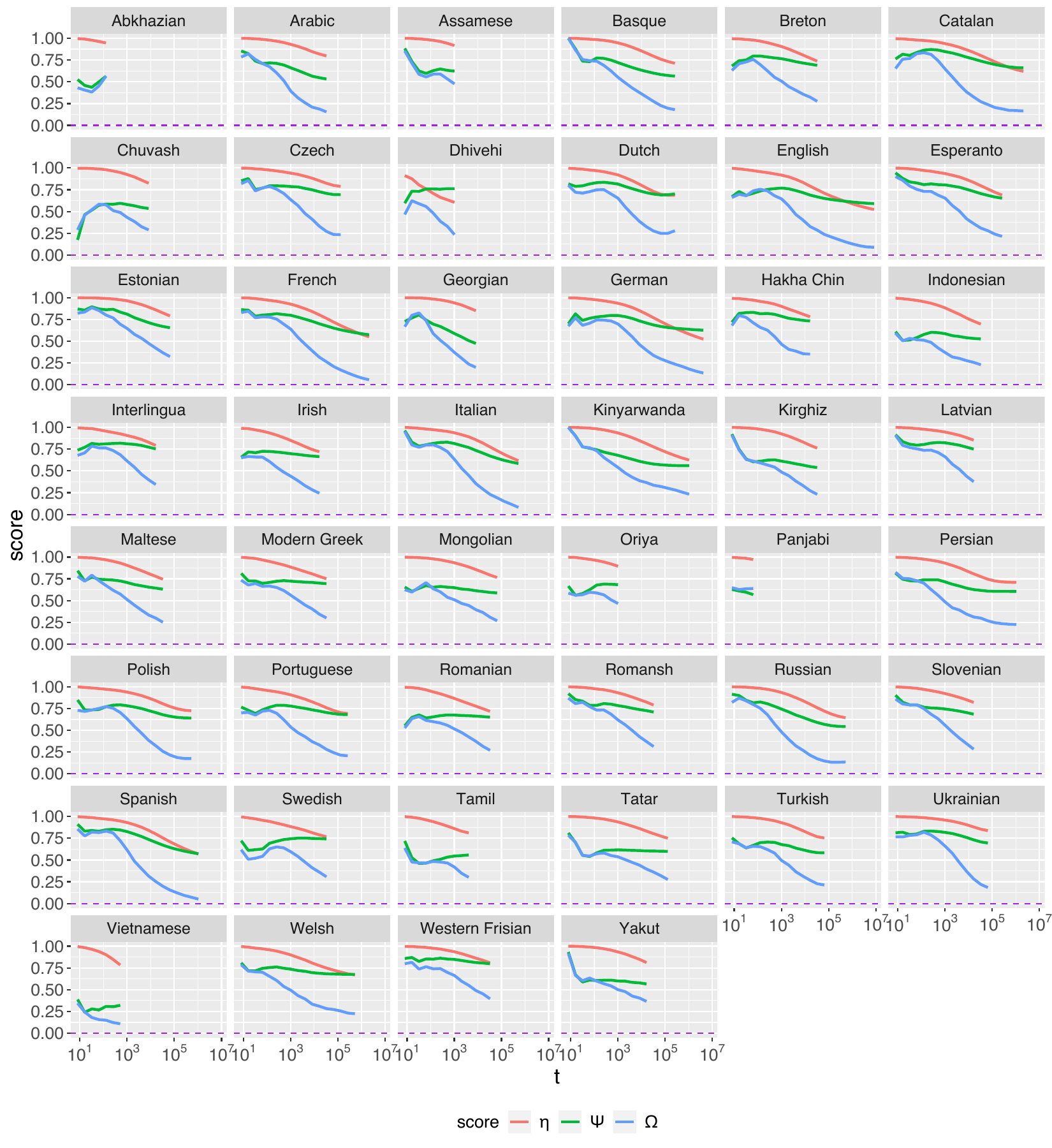}
    \caption{\label{fig:convergence_cv_medianDuration} Convergence of the scores in the CV collection with word length measured in duration. The format is the same as in \autoref{fig:convergence_pud}.
    % Values of the optimality scores ($\eta$, $\Psi$, and $\Omega$) for increasing number of tokens $t$. For each value of $t$, we show the average value of the score over $10^{2}$ random experiments (only those for which a value could be computed). The dashed line marks 0.
    }
\end{figure}

In \autoref{fig:convergence_pud}, we show the results for PUD; in \autoref{fig:convergence_cv_characters} for CV when length is measured in characters; in \autoref{fig:convergence_cv_medianDuration} when length is measured in duration. In all cases $\Omega$ has the widest range of variation, and it keeps rapidly decreasing as sample size grows, making it hard to make inferences about its possible convergence.  
Both $\eta$ and $\Psi$ evolve within a smaller range, however -- while the former mainly shows a slow but decreasing trend -- $\Psi$ seems to approach stability in some languages, especially when the sample is large enough.
In fact, we can mainly observe this phenomenon in CV, which contains larger samples. 

PUD is parallel but texts are rather short. Now we focus on CV because it has the largest samples and then is ideal for investigating convergence.
The languages for which convergence is suggested visually are: English, Kinyarwanda, Persian, Tatar and Welsh when length is measured in characters, and French, Kinyarwanda, Persian, Romanian and Tatar when length is measured in duration. The observed trends also suggest that the scores computed in the under-sampled languages are likely to vary largely in a larger sample of the same language, turning comparisons of these scores  potentially misleading in non-parallel sources.

% \begin{figure}[H]
% \centering
%     \begin{subfigure}{.75\textwidth}
%         \includegraphics[scale=0.45]{figures/convergence_cv_medianDuration_1.pdf}
%     \end{subfigure}
%     \begin{subfigure}{.75\textwidth}
%         \includegraphics[scale=0.45]{figures/convergence_cv_medianDuration_2.pdf}
%     \end{subfigure}
%     \caption{\label{fig:convergence_cv} Convergence of the scores. CV collection with word length measured in duration.
%     Values of the optimality scores ($\eta$, $\Psi$, and $\Omega$) for increasing number of tokens $t$. For each value of $t$, we show the average value of the score over $10^{2}$ random experiments (only those for which a value could be computed). The dashed line marks 0.
%     }
% \end{figure}

\subsection{Can a score be replaced by a simpler one?}

\begin{figure}[H]
    \centering
    \includegraphics[width = 0.9\linewidth]{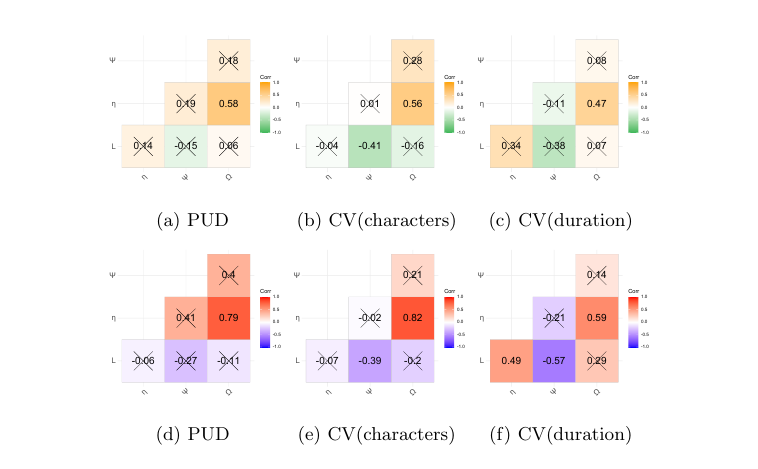}
    \caption{\label{fig:correlograms_scores} 
    The correlation between scores. Correlation tests were two-sided the $p$-values were corrected with a Holm-Bonferroni correction.  A crossed value signals a non significant correlation coefficient at a 95\% confidence level. 
    (a) Kendall $\tau$ correlation in PUD collection with word length measured in characters. (b) Kendall $\tau$ correlation in the CV collection with word length measured in characters. (c) Kendall $\tau$ correlation for CV collection with word length measured in duration. (d) Same as (a) using Pearson correlation. (e) Same as (b) using Pearson correlation. (f) Same as (c) using Pearson correlation.
    }
\end{figure}

To examine the replaceability of a score with a simpler one, we analyze the correlations between scores. In particular, $L$ is considered the simplest score (it does not integrate any baseline), after which comes $\eta$ (as it only integrates the minimum baseline). According to both Kendall and Pearson correlation, the only significant association in PUD is found between $\Omega$ and $\eta$ (\autoref{fig:correlograms_scores} (a-d)). This relation is consistently found also for both definitions of length in CV (\autoref{fig:correlograms_scores} (b, c, e and f)), suggesting its reality even in non-parallel corpora. In the context of CV, $\Psi$ turns out to be significantly associated with $L$, the simplest score, especially when length is measured in characters. Despite the significance, CV is not parallel and the associations are not particularly strong (the absolute value of the correlations does not exceed 0.6), suggesting that the additional complexity introduced by $\Psi$ still adds additional information on top of the information provided by $L$. 

\end{appendices}

\end{document}